\documentclass[12pt]{article}
\usepackage{amsmath,amssymb,amsthm,amsfonts,bm,dsfont,mathrsfs,mathtools,float,algorithm,algpseudocode}
\usepackage{avant,array,geometry,fontenc,inputenc,natbib,hyperref,multirow,enumerate,rotating,booktabs,color,latexsym,times,stmaryrd,xr,cleveref}
\allowdisplaybreaks

\newtheorem{theorem}{Theorem}
\newtheorem{lemma}{Lemma}

\newtheorem{proposition}{Proposition}
\newtheorem{definition}{Definition}

\newtheorem{assumption}{Assumption}

\DeclareMathOperator*{\argmax}{arg\,max}
 \DeclareMathOperator*{\argmin}{arg\,min}

\def\eop{\hfill $\Box$}

\begin{document}
\begin{center}
\Large{{\bf A Statistical Hypothesis Testing Framework for}} \\
\Large{{\bf Data Misappropriation Detection in Large Language Models}} \\
\medskip
\medskip
\medskip

{\large{\sc Yinpeng Cai$^\dag$, Lexin Li$^\ddag$\footnote{\label{note1}co-corresponding authors}, and Linjun Zhang$^*$}$^{\ref{note1}}$} \\
\medskip
{\normalsize {\it $^\dag$Peking University, $^\ddag$University of California at Berkeley, $^*$Rutgers University}}
\medskip
\end{center}

\begin{abstract}
Large Language Models (LLMs) are rapidly gaining enormous popularity in recent years. However, the training of LLMs has raised significant privacy and legal concerns, particularly   regarding the inclusion of copyrighted materials in their training data without proper attribution or licensing, an issue that falls under the broader concern of data misappropriation. In this article, we focus on a specific problem of data misappropriation detection, namely, to determine whether a given LLM has incorporated the data generated by another LLM. We propose embedding watermarks into the copyrighted training data and formulating the detection of data misappropriation as a hypothesis testing problem. We develop a general statistical testing framework, construct test statistics, determine optimal rejection thresholds, and explicitly control type I and type II errors. Furthermore, we establish the asymptotic optimality properties of the proposed tests, and demonstrate the empirical effectiveness through intensive numerical experiments.
\end{abstract}
\smallskip



\section{Introduction}
\label{sec:intro}

Large Language Models (LLMs) are rapidly gaining enormous popularity in recent years, thanks to their transformative ability to process and generate human-like text. The versatility and scalability of LLMs are driving widespread adoption, and are revolutionizing fields such as healthcare, education, and finance, with applications in content generation, data analysis, customer support, and research automation \citep{openai2024gpt4technicalreport}. However, alongside their vast potential, the training of LLMs has raised significant privacy and legal concerns, particularly regarding the distillation or inclusion of copyrighted materials in their training data without proper attribution or licensing \citep{beyond_fair, wu2023unveilingsecurityprivacyethical}. These concerns fall under the broader issue of \emph{data misappropriation}, namely, the unauthorized use, access, or exploitation of data by individuals or entities for unintended or unpermitted purposes, often in violation of governing regulations. Data misappropriation has been central to several high-profile debates, for instance, the lawsuit between the New York Times and OpenAI \citep{nyt2023-debate}, and OpenAI's \href{https://openai.com/policies/terms-of-service}{Terms of Service} that explicitly prohibits distilling ChatGPT's output to develop competing models. Detection of such data misappropriation is challenging though, especially when the probabilistic nature of LLMs generates content that may resemble, but does not directly copy, original works \citep{sag2023-safety, gesmer2024-challenge}. Such challenges underscore the importance of developing methods to identify and trace machine-generated text, and have recently generated considerable research interest in this area \citep{sadasivan2023can, jiang2023evading, mitchell2023detectgptzeroshotmachinegeneratedtext, ren2024copyright, huang2024optimalstatisticalwatermarking, zhao2023protecting, zhao2023provablerobustwatermarkingaigenerated}. In this article, we aim to address a specific and crucial question in data misappropriation detection in LLMs, i.e., \emph{How can one determine whether a given LLM has used the data generated by another LLM as part of its training corpus?} 

Watermarking is a technique used to embed identifiable patterns into the generated text, enabling traceability to distinguish AI-generated content from human-authored text \citep{Boenisch_2021}. This is typically achieved by subtly modifying token generation probabilities to favor specific words or patterns without compromising text coherence. There have been numerous watermarking techniques developed for LLMs, which can be broadly categorized as biased techniques \citep{atallah2001natural, yang2023watermarkingtextgeneratedblackbox, kirchenbauer2024watermarklargelanguagemodels}, and unbiased techniques \citep{Aaronson2023-Gumbel, fernandez2023bricksconsolidatewatermarkslarge, christ2023undetectablewatermarkslanguagemodels, hu2023unbiasedwatermarklargelanguage, zhao2024permuteandflipoptimallyrobustwatermarkable, wu2024dipmark}. The former explicitly alters token probabilities, favoring a predefined set of tokens, and often introducing detectable statistical biases in the text. The latter ensures the watermark signal is embedded without introducing noticeable deviations from the model's natural token distribution, preserving the text's statistical properties. 

In this article, we focus on a specific problem of data misappropriation detection, i.e., to determine whether a given LLM has incorporated the data generated by another LLM. To answer this question, we propose embedding watermarks into the copyrighted training data and formulating the detection of data misappropriation as a hypothesis testing problem. We develop a general statistical testing framework, construct test statistics, determine optimal rejection thresholds, and explicitly control type I and type II errors. Furthermore, we establish the asymptotic optimality properties of the proposed tests, and demonstrate the empirical effectiveness through intensive numerical experiments. Our proposal introduces a number of novel components and makes several unique contributions to the field. 

First, we develop a formal statistical hypothesis testing framework for data misappropriation detection. The core of our method lies in assessing the relationship between the tokens and secret keys. Under the null hypothesis $\mathcal{H}_0$, an LLM is not trained on the watermarked data, and the secret keys derived from its outputs are independent of the tokens. Conversely, under the alternative hypothesis $\mathcal{H}_1$, the tokens and secrete keys exhibit dependency. For each generation step $t$, given the \emph{next-token prediction} (NTP) distribution $\mathbf{P}_t$, the null hypothesis assumes that the token generation adheres to the multinomial distribution $\mathbf{P}_t$. However, under $\mathcal{H}_1$, the distribution of token $\omega_t$ also depends on the secret key $\zeta_t$, and deviates from $\mathbf{P}_t$. To capture this dependency, we use a test statistic comprising $\omega_t$ and $\zeta_t$ satisfying that: under $\mathcal{H}_0$, the statistic follows a fixed distribution independent of the NTP mechanism, whereas under $\mathcal{H}_1$, the statistic follows a distribution depending on the NTP. To find the rejection threshold for the proposed test statistic, we employ the large deviation theory to evaluate the asymptotic rates of type I and type II errors, and employ the class-dependent efficiency to account for variations in the NTP distributions. We then transform the detection problem into a solvable minimax optimization problem, and derive the optimal rejection threshold accordingly. 

Second, we consider two types of dependency  between the LLMs. Specifically, for an unwatermarked LLM, the tokens are sampled from a multinomial distribution conditional on preceding tokens. In contrast, for a watermarked LLM, a secret key is introduced that perturbs this multinomial distribution, generating tokens from a modified distribution. For any text, these secret keys can be computed at each generation step. In models unrelated to the watermarking process, these keys remain independent of the token sequences. However, when an LLM has been trained on a watermarked data, its token generation process depends, at least partially, on these keys. We consider two settings for such dependency: the \emph{complete inheritance} where the new LLM trained on the data from a watermarked LLM fully inherits the same watermarking distribution, and the \emph{partial inheritance} where the newly trained LLM generates outputs with distributions close to, but not identical to, the watermarked distribution. We quantify the difference between the two inheritance settings using the total variation distance, which serves as a bounded and computationally convenient metric for comparing multinomial distributions. We derive a separate rejection threshold for each inheritance setting. We also remark that we are the first to establish the optimality guarantees for data misappropriation detection under partial inheritance, while the existing literature only studies complete inheritance. 

Third, we also consider two types of rejection rule designs, and establish the optimality guarantees of the corresponding tests. Specifically, we consider the traditional setting where we fix the type I error at a predefined level and seek to maximize the asymptotic efficiency in the minimax sense. We also consider the setting where we aim to asymptotically minimize the sum of the type I and type II errors in the minimax sense. For the first setting, we show that our proposed test is asymptotically optimal in the sense that it achieves the largest power compared to \emph{all} other testing methods. For the second setting, we show that our test is minimax optimal in that it achieves the smallest sum of type I and type II errors among \emph{all} possible testing methods. Achieving such an optimality requires two key steps: deriving the optimal testing scheme based on the selected test statistic, and demonstrating that its asymptotic efficiency serves as an upper bound for the asymptotic efficiency of any alternative testing scheme. In cases where the minimax optimization problem derived from the chosen test statistic is convex, the solution is relatively straightforward. However, convexity is not always guaranteed in the partial inheritance setting, where the bounded total variation  distance introduces additional uncertainty. Addressing this involves fully characterizing the equality conditions of the underlying inequalities to ensure their validity throughout the estimation process. In the context of minimax-type inequalities, relying on a single equality condition is often insufficient. Instead, it is necessary to consider combinations of multiple cases. Evaluating the efficiency of these combined cases adds another layer of complexity, as it requires careful analysis of interdependencies and potential deviations. Our approach rigorously handles such challenging scenarios, ensuring robust detection performance even under non-convexity. 

Finally, we develop the tests for two commonly used, highly representative watermarking techniques. Specifically, we consider the Gumbel-max watermark \citep{Aaronson2023-Gumbel}, which adds structured noise from Gumbel distributions into the sampling process. We then consider the red-green-list watermark \citep{kirchenbauer2024watermarklargelanguagemodels}, which randomly divides the tokens into the red list and the green list, and biases token selection toward the green list. The former represents an unbiased watermarking technique, whereas the latter a biased one. For each watermarking, we derive the corresponding optimal detection rule and demonstrate its efficacy through numerical experiments. 

Our proposal is related to but also clearly distinctive of the existing literature. The line of research most closely related to ours is \emph{watermarking detection}, where the goal is to distinguish between human-authored text and machine-generated text \citep{mitchell2023detectgptzeroshotmachinegeneratedtext, Weber_Wulff_2023, li2024statisticalframeworkwatermarkslarge}, and \citep{li2024statisticalframeworkwatermarkslarge} was the first to develop a formal statistical framework and rigorously study the statistical efficiency of the testing methods. Nevertheless, our work, while related, differs from \citep{li2024statisticalframeworkwatermarkslarge} in numerous ways. First,  \cite{li2024statisticalframeworkwatermarkslarge} focused on detecting if a text is written by a human or an LLM, whereas we study an utterly different problem of whether one LLM has used the data generated by another LLM for training. Second, \citep{li2024statisticalframeworkwatermarkslarge} only considered the complete inheritance setting, while we also study the partial inheritance setting, which is closer to real-world situations. Theoretically, complete inheritance corresponds to a simple-vs-simple test \cite{van2000asymptotic}, where both null and alternative hypotheses only involve one distribution. In contrast, partial inheritance leads to a simple-vs-composite test, where the alternative hypothesis involves a class of distributions. Relatedly, the technical analysis in partial inheritance becomes much more challenging than that in complete inheritance. In particular, the objective function in partial inheritance lacks global convexity, and the convexity arguments used in complete inheritance no longer apply directly. To address this, we adopt a two-step strategy: we first fix the largest coordinate and analyze convexity in the remaining $n-1$ variables, then reduce the problem to a univariate optimization to complete the analysis. This careful decomposition is necessary to obtain tight minimax lower bounds. Third, \cite{li2024statisticalframeworkwatermarkslarge} studied the optimality within a restricted class of tests, namely, those based on thresholding sums of test statistics, whereas we establish the minimax optimality over the space of \emph{all} possible tests. That is, we prove our test achieves the lowest type II error among all tests with type I error at most $\alpha$, and also achieves the minimum total error when both types of error are penalized equally.  Another line of relevant research is membership inference attacks, where the goal is to determine whether a specific data point was part of the training dataset used to train a machine learning model \citep{shokri2017membershipinferenceattacksmachine, carlini2022membershipinferenceattacksprinciples, duan2023flocksstochasticparrotsdifferentially}. These attacks exploit overfitting or unintended information leakage, where the model exhibits different behaviors for training versus non-training data. Nevertheless, our work is different in that, instead of detecting the presence of specific data points in the training set, we aim to determine whether an LLM has incorporated the data generated by another LLM into its training corpus. This leads to a distinct analytical framework, as we focus on detecting subtle traces of patterns in data generation, rather than individual-level overfitting or leakage.

In summary, we are among the first to conceptualize data misappropriation detection in LLMs as a statistical hypothesis testing problem. We develop a general testing framework incorporating different inheritance settings, different rejection rule designs, and different watermarking techniques. We establish the minimax optimality guarantees for the proposed tests. Our proposal offers a valuable integration of statistical principles and large language models, and provides meaningful statistical insights into the rapidly evolving field of generative AI. 

The rest of the article is organized as follows. Section \ref{sec:setup} introduces the problem setup. Section \ref{sec:framework} develops the statistical hypothesis testing framework. Sections \ref{sec:gumbel} and \ref{sec:redgreen} study the Gumbel-max watermarking and the red-green-list watermarking. Sections \ref{sec:numerical} and \ref{sec:realdata} present the numerical results. The supplementary appendix collects all proofs and additional results.

\section{Problem Setup}
\label{sec:setup}

Throughout this article, we adopt the following notation. For an event $A$, let  $\mathbf{1}\{A\}$ denote the indicator function on $A$. Let  $g(n)=O(f(n)), g(n)=\Omega(f(n))$ denote that there exists numerical constant $c_1,c_2$, such that $\lim_{n\to\infty}g(n)/f(n)<c_1$ and $\lim_{n\to\infty}g(n)/f(n)=\infty$, respectively. Let $TV(\mu,\nu)$ denote the total variation (TV) distance between two probability measures $\mu,\nu$. In particular, for the case that $\mu=(p_1,p_2,\cdots,p_m),\nu=(q_1,q_2,\cdots,q_m)$, we can calculate their TV distance by $TV(\mu,\nu)=1/2 \sum_{i=1}^{m}|p_i-q_i|$. Let $TV_{\cdot|\zeta}(\mu,\nu)$ denote the TV distance between two probability measures $\mu,\nu$ conditional on a random variable $\zeta$.

Language models are equipped with a vocabulary  $\mathcal{W}$, which consists of words or word fragments called \emph{tokens}. Typically, a vocabulary contains \( |\mathcal{W}| = 50,000 \) tokens or more. In this article, we denote the number of tokens in the vocabulary by $|\mathcal{W}|=m$. To generate text, a language model requires a sequence of tokens that constitutes a \emph{prompt}. In the generated text, the entries with nonpositive indices, \( \omega_{-L_p}, \omega_{-L_p+1}, \dots, \omega_0 \), represent the prompt, with a length of \( L_p + 1 \). The entries with positive indices, \( \omega_1, \dots, \omega_T \), are tokens generated by the language model in response to the prompt. A language model for next-token prediction (NTP) is a function \( f \), often parameterized by a neural network, which takes as input a sequence of known tokens, \( \omega_{-N_p}, \omega_{-N_p+1}, \dots, \omega_{t-1} \), comprising the prompt and the first \( t - 1 \) tokens generated by the model. The output is an \( m \)-dimensional multinomial distribution, which represents the probability distribution for the next token. Formally, let \( \mathbf{P}_t := (P_{t,1}, P_{t,2}, \dots, P_{t,m}) \) denote this multinomial distribution at step \( t \), conditional on the prompt and previously generated tokens. As a valid probability distribution, \( \mathbf{P}_t \) satisfies that $\sum_{i=1}^{m} P_{t,i} = 1$ and $P_{t,i}\ge 0$. 

Watermarking is a technique for adjusting the NTP distribution so that the generated text exhibits specific statistical properties. The key insight is that the text generated by a watermarked LLM adheres to the watermarking generation rules. In this way, watermarks function as markers of an LLM’s generation rules, and enable differentiating a watermarked LLM from an unwatermarked one based on a given text. This principle forms the foundation of our research. In Appendix~\ref{append-sec:impossibility}, we further show theoretically that it is impossible to detect data misappropriation without watermarking.

More specifically, a watermarked LLM generates the next token through a process jointly determined by a secret key and the NTP distribution. Let \( \zeta_t \) denote the secret key at step \( t \), which is known only to the owner of the LLM. Formally, the generation rule of the watermarked LLM is $\omega_t \sim \mathcal{S}(\mathbf{P}_t, \zeta_t)$, where \( \mathcal{S} \) is a decoding function that outputs an \( m \)-dimensional multinomial distribution. In watermarking, a fundamental requirement is that averaging over the randomness in \( \zeta_t \), the difference between the watermarked and unwatermarked generation rules should not be too large. In other words, the NTP distribution conditional on previous tokens should not vary significantly between the watermarked and unwatermarked LLMs. For instance, for an unbiased watermarking technique, the watermarked NTP distribution averaged over the randomness in \( \zeta_t \) matches the unwatermarked NTP distribution conditional on previous tokens. Consequently, from a user's perspective, there is no noticeable difference between the text generated by a watermarked LLM and that produced by an unwatermarked LLM. 

Nevertheless, the secret keys possess statistical properties that enable effective detection. Typically, the secret keys are generated using hash functions. Hash functions map data of arbitrary size to fixed-sized values, with specific forms varying across different watermarking techniques. In many hashing-based watermarks, the distribution of watermarked text is biased towards certain \( k \)-grams by hashing a sliding window of the previous \( k - 1 \) tokens to determine the next token pseudorandomly. In practice, \( \zeta_t \) can be computed using a hash function \( \mathcal{A} \) that depends only on, for instance, the last five tokens \( \omega_{t-5}, \dots, \omega_{t-1} \) \citep{Aaronson2023-Gumbel}.

Formally, we consider the following setup for our data misappropriation detection problem. Suppose there are three roles. A \emph{victim} is the owner of an LLM who has trained an LLM with plentiful resources, has added some type of watermark in their LLM, and has the knowledge of the secret keys that correspond to the hash functions. A \emph{suspect} is the owner of another LLM, and is suspected to have used data generated by the victim's LLM as part of the training data for their own LLM. The suspect has no knowledge of the secret keys. A \emph{detector} is a trusted third party, such as law enforcement, who cooperates with the victim. The detector has the knowledge of both the secret keys and the NTP distribution of the victim's LLM, but does not know the NTP distribution of the suspect's LLM. The detector is to determine whether the suspect has used the data generated by the victim's LLM to train their own LLM.

\section{A Statistical Hypothesis Testing Framework}
\label{sec:framework}

\subsection{Hypotheses and test statistics}
\label{subsec:hypo}

Let $n \in \mathbb{N}$ denote the text length. Let $\omega_{1:n} := \omega_1\cdots\omega_n$ and $\zeta_{1:n} := \zeta_1\cdots\zeta_n$ denote the generated text and the secret keys, respectively. Given a prompt and a text $\omega_{1:n}$ generated by the LLM under investigation, we target the problem of whether this LLM utilizes the data generated by another watermarked LLM, in the form of hypotheses:
\vspace{-0.01in}
\begin{align} \label{eqn:hypo}
\begin{split}
\mathcal{H}_0: \;\; & \omega_{1:n} \text{ is generated by the LLM without data misappropriation}; \\
\mathcal{H}_1: \;\; & \omega_{1:n} \text{ is generated by the LLM with data misappropriation}.
\end{split}
\end{align}
In addition, the secret keys are generated using the hash function $\mathcal{A}$ that depends on the previous tokens. The knowledge of $\mathcal{A}$ is only available to the victim and the detector, but not the suspect. As such, $\zeta_{1:n}$ are regarded as random variables by the suspect, but are viewed deterministic by the victim and the detector, since they can be calculated by previous tokens plus the knowledge of the hash function. Meanwhile, the watermarked LLM generates the next token according to the rule, $\omega_t\sim\mathcal{S}(\mathbf{P}_t,\zeta_t)$, for some decoding function $\mathcal{S}$, and $\mathbf{P}_t$ represents the NTP distribution of the unwatermarked LLM given the prompt and previous tokens.

Next, we mathematically formulate complete inheritance and partial inheritance. 

\begin{definition}[\textbf{Complete inheritance}]
The suspect's LLM follows the same generation rule as the victim's LLM, in that the hash function $\mathcal{A}$ and the decoding function $\mathcal{S}$ of both LLMs are the same if data misappropriation occurs.
\end{definition}

Under the complete inheritance setting, the hypotheses in \eqref{eqn:hypo} become
\vspace{-0.01in}
\begin{align} \label{eqn:hypo-complete}
\mathcal{H}_0: \omega_t|\zeta_t\sim\mathbf{P}_t, \quad\quad
\mathcal{H}_1: \omega_t|\zeta_t\sim\mathcal{S}(\mathbf{P}_t,\zeta_t), \quad \text{ for } \; t=1, 2, \ldots, n.
\end{align} 

In practice, a newly trained LLM that uses data generated by a watermarked LLM produces outputs with a distribution that preserves certain patterns of the watermarked distribution, but \emph{not} identical. This scenario is captured by the following partial inheritance setting.

\begin{definition}[\textbf{Partial inheritance}] There exists an upper bound for the total variation (TV) distance between the probability distribution of data generated by the suspect's LLM under data misappropriation, and the probability distribution of data generated by the victim's watermarked LLM, in that $TV_{\cdot|\zeta_t}\left(\omega_t,\mathcal{S}(\mathbf{P}_t,\zeta_t)\right)\leq 1-\theta$, where $1-\theta$ quantifies the upper bound on the allowable difference between the two distributions.
\end{definition}

We make a few remarks. First, we note that partial inheritance may \emph{not} fully capture the nuanced behaviors of real-world LLMs, and we do \emph{not} intend to present it as a definitive or exclusive model of inheritance. Rather, we introduce this setting to capture the intuition that when a suspect model is trained on outputs from a victim model, it inevitably learns both the semantic signals and any embedded watermark patterns, as there is no principled way for the suspect model to disentangle them. Under this assumption, we model the next-token distribution of the suspect model as being close to that of the victim model, and formalize this proximity via a TV distance bound. Importantly, we do \emph{not} fix the TV distance to be small; instead, our theoretical framework allows the bound to vary over the full range of $(0,1)$, offering flexibility in characterizing different degrees of inheritance. Second, while partial inheritance serves a theoretical role in justifying our test construction, the resulting score function is directly used as a practical detection tool. Empirically, as we show in Section \ref{sec:realdata}, our method remains effective and robust even when the true inheritance behavior deviates from the assumed model.

Under the partial inheritance setting, the hypotheses in \eqref{eqn:hypo} become
\begin{align} \label{eqn:hypo-partial}
\mathcal{H}_0: \omega_t|\zeta_t\sim\mathbf{P}_t; \quad\quad
\mathcal{H}_1: TV_{\cdot|\zeta_t}\left(\omega_t,\mathcal{S}(\mathbf{P}_t,\zeta_t)\right)\leq 1-\theta, \quad \text{ for } \; t=1, 2, \ldots, n.
\end{align}
We also briefly remark that, for the detection task to be feasible, the constant $\theta$ cannot be too small, as an overly small $\theta$ would imply that the two distributions are too similar to differentiate.

Next, we describe the distribution of $\omega | \zeta$ under $\mathcal{H}_1$. In commonly used watermarking techniques, $\mathcal{S}(\mathbf{P}, \zeta)$ can only take finitely many values. We provide more details in Sections \ref{sec:gumbel} and \ref{sec:redgreen}. Let $k$ denote the total possible values of $\mathcal{S}(\mathbf{P}, \zeta)$, and partition the space of $(\mathbf{P}, \zeta)$ into multiple regions, $A_1, A_2, \ldots, A_k$, such that in each region $A_i$, the corresponding multinomial distribution $\mathcal{S}(\mathbf{P}, \zeta)$ is identical, and is represented by a multinomial probability vector $(s_{i1}, s_{i2}, \ldots, s_{im})$, for $i\in[k]$. We further assume that, for any $t\in[n]$, and all $(\mathbf{P}, \zeta) \in A_i$, the distribution of $\omega$ in the partial inheritance setting is the same, which is represented by $(q_{t,{i1}}, q_{t,{i2}}, \ldots, q_{t,{im}})$, regardless of the exact values of $(\mathbf{P}, \zeta)$. This assumption simplifies the theoretical discussion and allows us to represent the extreme points using a feature matrix. It can be relaxed without compromising the validity of our theoretical results; see the remarks after Theorem~\ref{hypo-minimax-sum}. Now, we formally define the feature matrix, whose rows describe different possible distributions of $\omega|\zeta$.

\begin{definition}
\label{definition:feature matrix}
At the t-th token, define the feature matrix under the alternative $\mathcal{H}_1$ of the complete inheritance setting as $\mathbf{S}_t=(s_{t,ij})_{k\times m}$, and the feature matrix under the alternative $\mathcal{H}_1$ of the partial inheritance setting as $\mathbf{Q}_t= (q_{t,ij})_{k \times m}$.
\end{definition}

The feature matrix allows us to directly derive the distribution of test statistic under the null and alternative hypotheses. Additionally, the use of the feature matrix helps the representation of extreme points, and provides a more efficient framework for analysis. Importantly, we note that the order of rows in the feature matrix is arbitrary and does not affect the results.

Let $Y_t := Y(\omega_t,\zeta_t)$ denote the test statistic, we have the following rejection rule,
\begin{align*}
T_h(Y_{1:n}) :=
\begin{cases} 
1 & \text{if } \sum_{t=1}^{n}h(Y_t)\geq \gamma_{n}, \\
0 & \text{if } \sum_{t=1}^{n}h(Y_t)< \gamma_{n}.
\end{cases}
\end{align*}
where $h(\cdot)$ is the score function. We derive the specific forms of $Y_t$ and $h(\cdot)$ for specific watermarking techniques later in Sections \ref{sec:gumbel} and \ref{sec:redgreen}. We reject $\mathcal{H}_0$ if $T_h(Y_{1:n}) = 1$. 

We also remark that, for complete inheritance, our testing problem reduces to the problem of determining whether a text is generated by an unwatermarked LLM or by a watermaked LLM, whose NTP distributions are the same, which has been studied in \cite{li2024statisticalframeworkwatermarkslarge}. However, our problem is more general, in that we also consider partial inheritance, different rejection rule designs, and obtain tighter optimality guarantees.

\subsection{Rejection threshold}
\label{subsec:threshold}

For our testing method, it is crucial to choose an appropriate score function $h(\cdot)$ and the threshold $\gamma_n$ to ensure the optimality of the test. Next, we consider two rejection rule designs separately. We first fix the type I error at a predefined level and seek to maximize the asymptotic efficiency. We then aim to asymptotically minimize the sum of the type I and type II errors. 

We first consider the setting of the fixed type I error, where we choose the threshold $\gamma_{n}$, such that $\mathbb{P}_{\mathcal{H}_0}(T_h(Y_{1:n})=1)=\alpha$ for a pre-specified significance level $\alpha$, while we choose the optimal score function $h(\cdot)$ later to maximize the asymptotic power of the test. Our key idea is to turn hypothesis testing to a minimax optimization problem. Let $\mathbb{E}_0$ be the expectation under $\mathcal H_0$, $\mathbb{E}_{1,\mathbf{P}}, \mathbb{E}_{1,\mathbf{P},\mathbf{Q}}$ be the expectation under $\mathcal H_1$ that depends on $\mathbf{P}$ or on $\mathbf{P}, \mathbf{Q}$, and $\mathcal{P}$, $\mathcal{Q}$ be some distribution classes whose forms are specified later for specific watermarking techniques. 

\begin{theorem}\label{hypo-minimax-fixed} 
(Hypothesis testing as minimax optimization, fixed type I error).
\vspace{-0.1in}
\begin{enumerate}[(a)]
\item (Complete inheritance). Suppose $\mathbf{P}_t \in \mathcal{P}$ are i.i.d.\ random vectors for all $t$. For any $h$ satisfying $\mathbb{E}_0 |h| < \infty$, the type II error of the rejection rule $T_h$ satisfies that
\begin{equation}\label{thm:general:h1}
\limsup_{n \to \infty} \mathbb{P}_{\mathcal{H}_1}\left( T_h(Y_{1:n}) = 0 \right)^{1/n} \leq  e^{-R_{\mathcal{P}}(h)},
\end{equation}
where $R_{\mathcal{P}}(h) = -\inf_{\theta\geq 0} \sup_{\mathbf{P} \in \mathcal{P}} \{ \theta  \mathbb{E}_0 h(Y) +  \log \phi_{\mathbf{P}, h}(\theta) \}$, $\phi_{\mathbf{P},h}(\theta)=\mathbb{E}_{1,\mathbf{P}}e^{-\theta h(Y)}$. In addition, the inequality in \eqref{thm:general:h1} is tight, in that there exist $\mathbf{P}^*$, such that, if $\mathbf{P}_t=\mathbf{P}^*$ for all $t$, then for any positive $\epsilon$ and a sufficiently large $n$,
\begin{equation*}
\mathbb{P}_{\mathcal{H}_1}\left( T_h(Y_{1:n})=0 \right)^{1/n} \geq e^{-\{ R_{\mathcal{P}}(h)+\epsilon \}}.
\end{equation*}

\item (Partial inheritance). Suppose $\mathbf{P}_t \in \mathcal{P},\mathbf{Q}_t \in \mathcal{Q}$ are i.i.d.\ random vectors and i.i.d. random matrices for all $t$.  For any $h$ satisfying $\mathbb{E}_0 |h| < \infty$, the type II error of the rejection rule $T_h$ satisfies that 
\vspace{-0.05in}
\begin{equation}\label{thm:general:h1partial}
\limsup_{n \to \infty} \mathbb{P}_{\mathcal{H}_1}\left( T_h(Y_{1:n}) = 0 \right)^{1/n} \leq  e^{-R_{\mathcal{P},\mathcal{Q}}(h)},
\end{equation}
where $R_{\mathcal{P},\mathcal{Q}}(h) = -\inf_{\theta\geq 0}\sup_{\mathbf{Q}\in\mathcal{Q}} \sup_{\mathbf{P} \in \mathcal{P}} \{ \theta \mathbb{E}_0 h(Y) + \log  \phi_{\mathbf{P},\mathbf{Q}, h}(\theta) \}$,  $\phi_{\mathbf{P},\mathbf{Q},h}(\theta)=\mathbb{E}_{1,\mathbf{P},\mathbf{Q}}e^{-\theta h(Y)}$. In addition, the inequality in \eqref{thm:general:h1partial} is tight, in that there exist $\mathbf{P}^*$ and $\mathbf{Q}^*$, such that, if $\mathbf{P}_t=\mathbf{P}^*,\mathbf{Q}_t=\mathbf{Q}^*$ for all $t$, then for any positive $\epsilon$ and a sufficiently large $n$,
\vspace{-0.01in} 
\begin{equation*}
\mathbb{P}_{\mathcal{H}_1}\left( T_h(Y_{1:n})=0 \right)^{1/n} \geq e^{-(R_{\mathcal{P},\mathcal{Q}}(h)+\epsilon)}.
\vspace{-0.1in} 
\end{equation*}
\end{enumerate}
\end{theorem}

Theorem \ref{hypo-minimax-fixed} essentially transforms the hypothesis testing problem into a minimax optimization problem, by showing that the two problems are equivalent under the worst case of NTP distribution $\mathbf{P}_1,\mathbf{P}_2,\cdots,\mathbf{P}_n$. Consequently, we turn to the following optimization problems for complete and partial inheritance, respectively, 
\vspace{-0.05in}
\begin{eqnarray} 
\sup_{h} R_{\mathcal{P}}(h) & = & -\inf_h\inf_{\theta \geq 0} \sup_{\mathbf{P} \in \mathcal{P}} \left\{ \theta  \mathbb{E}_0 h(Y) + \log  \phi_{\mathbf{P}, h}(\theta) \right\} \nonumber \\
& = & -\inf_{h, \theta \geq 0} \sup_{\mathbf{P} \in \mathcal{P}} \left\{ \mathbb{E}_0 \theta h(Y) +  \log \mathbb{E}_{1, \mathbf{P}} e^{-\theta h(Y)} \right\}; \label{eqn:optim-general-fixed-complete} \\
\sup_{h} R_{\mathcal{P},\mathcal{Q}}(h) & = & -\inf_h\inf_{\theta \geq 0} \sup_{\mathbf{Q} \in \mathcal{Q}} \sup_{\mathbf{P} \in \mathcal{P}} \left\{ \theta  \mathbb{E}_0 h(Y) +  \log  \phi_{\mathbf{P},\mathbf{Q}, h}(\theta) \right\}  \nonumber \\
& = & -\inf_{h, \theta \geq 0} \sup_{\mathbf{Q} \in \mathcal{Q}} \sup_{\mathbf{P} \in \mathcal{P}} \left\{ \mathbb{E}_0 \theta h(Y) +  \log \mathbb{E}_{1, \mathbf{P},\mathbf{Q}} e^{-\theta h(Y)} \right\}. \label{eqn:optim-general-fixed-partial}
\end{eqnarray}
We provide the detailed solutions to these minimax optimization problems later in Sections \ref{sec:gumbel} and \ref{sec:redgreen} for specific watermarking techniques. 

We next consider the setting of minimizing the sum of type I and type II errors, 
\begin{align*}
\inf_{h}\limsup_{n\to\infty}\left[ \inf_{\gamma} \left\{ \mathbb{P}_{\mathcal{H}_0}(T_h(Y_{1:n})=1)+\mathbb{P}_{\mathcal{H}_1}(T_h(Y_{1:n})=0) \right\} \right]^{\frac{1}{n}}.
\end{align*}

\begin{theorem} \label{hypo-minimax-sum}
(Hypothesis testing as minimax optimization, sum of type I and type II errors).
\vspace{-0.1in}
\begin{enumerate}[(a)]
\item (Complete inheritance). Suppose $\mathbf{P}_t \in \mathcal{P}$ are i.i.d. random vectors for all $t$. For any $h$ satisfying $\mathbb{E}_0 |h| < \infty$, the type II error of the rejection rule $T_h$ satisfies that
\begin{equation*}
\limsup_{n \to \infty} \left\{ \mathbb{P}_{\mathcal{H}_0}(T_h(Y_{1:n})=1)+\mathbb{P}_{\mathcal{H}_1}(T_h(Y_{1:n}) = 0) \right\}^{1/n} \leq  e^{-S_{\mathcal{P}}(h)},
\end{equation*}
where $S_{\mathcal{P}}(h) = -\inf_{\theta_1, \theta_2 > 0} \big\{ {\theta_2} {(\theta_1 + \theta_2)^{-1}} \log \mathbb{E}_0\left[ \exp\{ \theta_1 h(Y)\} \right] + {\theta_1}{(\theta_1 + \theta_2)^{-1}} \log \sup_{\mathbf{P} \in \mathcal{P}}$ $\mathbb{E}_{1, \mathbf{P}} \left[ \exp\left\{(-\theta_2 h(Y))\right\} \right] \big\}$. In addition, this inequality is tight, in that there exists $\mathbf{P}^*$, such that if $\mathbf{P}_t=\mathbf{P}^*$ for all $t$, then for any $0<b<1$ and a sufficiently large $n$,
\begin{align*}
\inf_{\gamma}\left\{ \mathbb{P}_{\mathcal{H}_0}(T_h(Y_{1:n})=1)+\mathbb{P}_{\mathcal{H}_1}(T_h(Y_{1:n})=0) \right\}^{1/n} \geq b\cdot e^{-S_{\mathcal{P}}(h)}.
\end{align*}

\item (Partial inheritance). Suppose $\mathbf{P}_t \in \mathcal{P},\mathbf{Q}_t \in \mathcal{Q}$ are i.i.d. random vectors and i.i.d. random matrices for all $t$. For any $h$ satisfying $\mathbb{E}_0 |h| < \infty$, the type II error of the rejection rule $T_h$ satisfies that 
\begin{equation*}
\limsup_{n \to \infty} \left\{ \mathbb{P}_{\mathcal{H}_0}(T_h(Y_{1:n})=1)+\mathbb{P}_{\mathcal{H}_1}(T_h(Y_{1:n}) = 0) \right\}^{1/n} \leq e^{-S_{\mathcal{P},\mathcal{Q}}(h)},
\end{equation*}
where $S_{\mathcal{P},\mathcal{Q}}(h) = -\inf_{\theta_1, \theta_2 > 0} \big\{ {\theta_2}{(\theta_1 + \theta_2)^{-1}} \log \mathbb{E}_0\left[ \exp\left\{ \theta_1 h(Y) \right\} \right] + {\theta_1}{(\theta_1 + \theta_2)^{-1}} \log \sup_{\mathbf{Q} \in \mathcal{Q}}$ $\sup_{\mathbf{P} \in \mathcal{P}} \mathbb{E}_{1, \mathbf{P}, \mathbf{Q}} \left[ \exp(-\theta_2 h(Y)) \right] \big\}$. 
In addition, this inequality is tight, in that there exists $\mathbf{P}^*$ and $\mathbf{Q}^*$, such that if $\mathbf{P}_t=\mathbf{P}^*,\mathbf{Q}_t=\mathbf{Q}^*$ for all $t$, then for any $0<b<1$ and a sufficiently large $n$, 
\vspace{-0.1in}
\begin{align*}
\inf_{\gamma}\left\{ \mathbb{P}_{\mathcal{H}_0}(T_h(Y_{1:n})=1)+\mathbb{P}_{\mathcal{H}_1}(T_h(Y_{1:n})=0) \right\}^{1/n} \geq  b\cdot e^{-S_{\mathcal{P},\mathcal{Q}}(h)}.
\end{align*}
\end{enumerate}
\end{theorem}

Theorem \ref{hypo-minimax-sum} again transforms hypothesis testing to minimax optimization, and we turn to the following optimization problems for complete and partial inheritance, respectively, 
\begin{eqnarray} 
\sup_h S_{\mathcal{P}}(h) & = & -\inf_{h}\inf_{\theta_1,\theta_2>0}\frac{\theta_2}{\theta_1+\theta_2}\log\mathbb{E}_0\exp{\{ \theta_1h(Y) \}} \nonumber \\
& & \quad\quad\quad\quad\quad\quad + \; \frac{\theta_1}{\theta_1+\theta_2}\log\sup_{\mathbf{P}\in\mathcal{P}}\mathbb{E}_{1,\mathbf{P}}\exp{\{ -\theta_2h(Y) \}}; \label{eqn:optim-general-sum-complete} \\
\sup_h S_{\mathcal{P},\mathcal{Q}}(h) & = & -\inf_{h}\inf_{\theta_1,\theta_2>0}\frac{\theta_2}{\theta_1+\theta_2}\log\mathbb{E}_0\exp{\{ \theta_1h(Y) \}} \nonumber \\ 
& & \quad\quad\quad\quad\quad\quad + \; \frac{\theta_1}{\theta_1+\theta_2}\log\sup_{\mathbf{Q}\in\mathcal{Q}}\sup_{\mathbf{P}\in\mathcal{P}}\mathbb{E}_{1,\mathbf{P},\mathbf{Q}}\exp{\{ -\theta_2h(Y) \}}. \label{eqn:optim-general-sum-partial} 
\end{eqnarray}

We make a few remarks regarding Theorems \ref{hypo-minimax-fixed} and \ref{hypo-minimax-sum}. 

First, the proofs of both theorems rely on that $\{Y_t\}$, or $\{ h(Y_t) \}$, are i.i.d.\ under both $\mathcal{H}_0$ and $\mathcal{H}_1$. This is justified under two perspectives: the cryptographic property of the hash function guarantees that $\zeta_t$ behaves like a fresh independent key at each time step from the analyst’s point of view, and the next-token distributions $\mathbf{P}_t$ are assumed to be i.i.d. More specifically, our framework depends on a cryptographic-style hash function $\mathcal{A}$, which deterministically maps sequences, typically composed of the prompt and prior tokens, to secret keys. That is, we assume a secret key generation rule of the form $\zeta_{t+1} = \mathcal{A}(\text{prompt}, \omega_{1:t})$. Importantly, while the function is deterministic, it is designed to be indistinguishable from a truly random mapping to any observer who does not know the internal structure or seed of $\mathcal{A}$. This property aligns with standard constructions of pseudorandom functions that are widely used in cryptography \citep{goldreich2001foundations}. Under this design, and from the perspective of a statistical analyst or adversary without access to $\mathcal{A}$, the secret keys $\zeta_t$ can be treated as statistically independent of the token sequence $\omega_{1:t}$. Therefore, even though $\zeta_t$ is a function of the previous tokens, its randomness remains cryptographically opaque. Given a fixed next-token distribution $\mathbf{P}_t$ and a randomized secret key $\zeta_t$, the test statistic $Y_t$ depends only on the current token $\omega_t$ and $\zeta_t$, and is thus modeled as a draw from some distribution $\mu_{1, \mathbf{P}_t}$ under $\mathcal{H}_1$. Furthermore, we assume that the distributions $\mathbf{P}_t$ are i.i.d. across $t$. While this assumption is a simplification, it is standard in theoretical analysis \citep{huang2024optimalstatisticalwatermarking}, and provides a tractable setting for establishing error bounds and optimality guarantees. Under this assumption, the sequence $\{Y_t\}$ can be treated as i.i.d.\ under both $\mathcal{H}_0$ and $\mathcal{H}_1$. In the case of $\mathcal{H}_0$, since there is no watermarking, the secret keys $\zeta_t$ are not present, and the $Y_t$ values depend solely on unwatermarked generation. In the case of $\mathcal{H}_1$, the randomized behavior induced by $\mathcal{A}$ ensures that $\{Y_t\}$ are conditionally independent and identically distributed when marginalizing over the hash randomness. We also note that, this i.i.d.\ assumption may not hold for \emph{any} watermark technique, but instead for the two that we study in this article.

Next, we clarify why it is reasonable to assume the distribution of $\omega|\zeta$ is identical whenever $\mathcal{S}(\mathbf{P},\zeta)$ is the same in the partial inheritance setting. In fact, in \eqref{eqn:optim-general-sum-partial}, and also similarly in \eqref{eqn:optim-general-fixed-partial}, the only term that depends on the distribution of $Y$ under $\mathcal{H}_1$ is $\mathbb{E}_{1,\mathbf{P},\mathbf{Q}}e^{-\theta_2 h(Y)}$. We focus on  optimizing $\sup_{\mathbf{Q}\in\mathcal{Q}}\sup_{\mathbf{P}\in\mathcal{P}}\mathbb{E}_{1,\mathbf{P},\mathbf{Q}}e^{-\theta_2 h(Y)}$ given $\theta_2$ and $h$. For this optimization, it is sufficient to consider the optimal $\mathbf{Q}$ given $\mathbf{P}$. By definition, $\mathbf{Q}$ imposes an upper bound on the total variation distance between $\omega|\zeta$ and $\mathcal{S}(\mathbf{P},\zeta)$. Consequently, the worst-case scenario can be reduced to the case where the distribution of $\omega|\zeta$ is identical whenever $\mathcal{S}(\mathbf{P},\zeta)$ is the same. This reduction also justifies the introduction of the feature matrix, which provides a well-defined and structured representation of the relation between $\omega|\zeta$ and $\mathcal{S}(\mathbf{P},\zeta)$, and simplifies the analysis by capturing all the relevant information in a systematic manner. Furthermore, we clarify that we treat $m$ and $k$ fixed, and we do not impose any additional regularity conditions on $\gamma_n$ or $\mathcal{P}$.

Next, the choice of the distribution class $\mathcal{P}$ and $\mathcal{Q}$ determines the form of the optimization problems derived from the hypothesis testing problems. Specifically, the distribution class $\mathcal{P}$ needs to exclude the singleton distributions, e.g., $(1,0,0,\cdots,0)$, because in such cases, the joint distribution of $(\omega_t,\zeta_t)$ would be identical under $\mathcal H_0$ and $\mathcal H_1$, making the two hypotheses indistinguishable. In the partial inheritance setting, the distribution class $\mathcal{Q}$ must satisfy the condition $TV_{\cdot|\zeta_t}\left(\omega_t,\mathcal{S}(\mathbf{P}_t,\zeta_t)\right)\leq 1-\theta$ for all $t\in[n]$, ensuring the consistency with the bounded total variation condition. Beyond these requirements, our framework is flexible and can, in principle, analyze any distribution classes. Nevertheless, we follow the literature \cite{li2024statisticalframeworkwatermarkslarge,cai2024betterstatisticalunderstandingwatermarking,huang2024optimalstatisticalwatermarking}, and primarily focus on the following distribution class for $\Delta,\theta\in(0,1)$: 
\begin{align*}
\mathcal{P}_\Delta := \left\{ \mathbf{P}:\max_{i\in[m]}P_i\leq 1-\Delta \right\}, \quad
\mathcal{Q}_\theta := \left\{ \omega|\zeta:TV_{\cdot|\zeta}\left(\omega,\mathcal{S}(\mathbf{P},\zeta)\right)\leq 1-\theta \right\}.
\end{align*}

Finally, we comment on the practical usefulness of our method. In our theoretical analysis for partial inheritance, we construct the optimal score function $h(\cdot)$ assuming knowing the true $\Delta$ that controls the entropy of the probability distribution, and $\theta$ that controls the closeness between the next-token distributions for the original and watermarked data. While this is idealized, it helps reveal the structure of the optimal testing procedures. In practice, the true values of $\Delta$ and $\theta$ are generally unknown. Nevertheless, we argue they can be reasonably estimated or approximated in numerous cases. For instance, if one has knowledge of the NTP distribution of the underlying LLM, e.g., its temperature setting or sampling method, such knowledge can be used to estimate $\Delta$. A higher sampling temperature typically induces greater randomness, corresponding to a larger value of $\Delta$. Similarly, the value of $\theta$ can be informed by the robustness of the watermarking method. A more robust watermarking scheme typically yields a smaller value of $\theta$. Furthermore, our numerical experiments in Sections \ref{sec:numerical} and \ref{sec:realdata}  show that our proposed detection method remains effective and robust even when the estimates of $\Delta$ and $\theta$ are imperfect. We show that, with reasonably chosen but approximate values of $\Delta$ and $\theta$, our method outperforms the baseline detection solutions. Besides, precise tuning is not required, and an approximate estimate of the parameters at the right scale is sufficient.

\section{Analysis of Gumbel-max Watermark}
\label{sec:gumbel}

We now apply our general hypothesis testing framework to a specific watermarking scheme: the Gumbel-max watermark \citep{Aaronson2023-Gumbel}. We first derive the feature matrix. We then establish three key results: the distribution of the test statistic, the optimal score function, and the optimality guarantee of our proposed test. We first report the results for complete inheritance in Theorems \ref{thm:cite} to \ref{optimality-complete} , then parallelly for partial inheritance in Theorems \ref{thm:partial:gumbel} to \ref{thm:gumbel:partial:optimal}.

\subsection{Feature matrix}
\label{sec:feature-matrix-gm}

We first quickly review the Gumbel-max watermark. The secret key $\zeta_t$ consists of $m = |\mathcal{W}|$ i.i.d.\ random variables $(U_{t,\omega})_{\omega \in \mathcal W}$ following a uniform $(0,1)$ distribution. Given the NTP distribution $\mathbf{P} = (P_1, P_2,\ldots, P_m)$, the next token is generated following the rule $\omega_{t}= \argmax_{\omega\in \mathcal{W}} {\log U_{t,\omega}} / {P_\omega}$. The corresponding score vector $\mathcal{S}(\mathbf{P}_t, \zeta_t)$ is defined as the one-hot vector indicating which token $\omega$ is selected under this rule; i.e., it takes the form $(0, \ldots, 1, \ldots, 0)$, where $1$ appears at the position of the selected token. As the detector has the knowledge of the hash function, the generating process becomes a deterministic process. In other words, if the detector knows $\zeta_t$ as well as $\mathbf{P}_t$, one can directly determine what the next token is. We note that, although $\mathbf{P}_t$ can vary continuously over the probability simplex and thus takes infinitely many values, the randomness in $\zeta_t$ and the argmax rule ensure that $\mathcal{S}(\mathbf{P}_t, \zeta_t)$ is always one of $m$ possible one-hot vectors. Therefore, regardless of how $\mathbf{P}_t$ varies, the function $\mathcal{S}(\mathbf{P}_t, \zeta_t)$ maps into a finite set of size $m$, corresponding to the $m$ possible token choices. Consequently, the number of possible values $k$ that $\mathcal{S}(\mathbf{P}_t, \zeta_t)$ can take is exactly $m$; in other words, $k = m$.

We next derive the feature matrix for the Gumbel-max watermark. Recall that, for both complete inheritance and partial inheritance, the feature matrix represents the conditional probabilities of token generation under $\mathcal H_0$ and $\mathcal H_1$. Under $\mathcal H_0$, the generation of $\omega_t$ is independent of $\zeta_t$, and $\omega_t$  follows the unwatermarked NTP distribution $\mathbf{P}_t$. Under $\mathcal{H}_1$, the generation rule differs depending on the inheritance setting. Let $\omega^{(i)}$ denote the $i$th token in the vocabulary $\mathcal W$. In the complete inheritance setting, we have $\omega_t = \arg \max_{i \in [m]} {\log U_{t,i}} / {P_{t,i}}$. The probability that $\omega_t = \omega^{(i)}$ is 1, given that ${\log U_{t,i}} / {P_{t,i}}$ is the largest among $\{{\log U_{t,i'}}/{P_{t,i'}}\}_{i'\in[m]}$. In the partial inheritance setting, the generated tokens satisfy that $TV(\omega_t | \zeta_t,\mathcal{S}(\mathbf{P}_t, \zeta_t)) \leq 1 - \theta$, for all $t$. Since $\mathcal{S}(\mathbf{P}_t, \zeta_t)$ is deterministic, the corresponding probability distribution must be $(1, 0, \ldots, 0)$ or one of its permutations. Therefore, the probability that $\omega_t = \omega^{(i)}$ is no smaller than $\theta$. Let  {$\mathbf{S}_{\mathbf{P}_t}$} be an $m \times m$ matrix with each row being $\mathbf{P}_t$. The next proposition formally summarizes the feature matrix for the complete and partial inheritance settings. 
\begin{proposition}\label{prop1}
(Feature matrix). Following Definition~\ref{definition:feature matrix}, for $t\in[n]$, 
\vspace{-0.1in}
\begin{enumerate}[(a)]
\item (Complete inheritance). The feature matrix is $\mathbf{S}_{\mathbf{P}_t}$ under $\mathcal H_0$, and is the identity matrix $I_{m \times m}$ under $\mathcal H_1$. 
\item (Partial inheritance). The feature matrix is $\mathbf{S}_{\mathbf{P}_t}$ under $\mathcal H_0$, and is any matrix in the class $\big\{ \mathbf{Q}_{t}\in\mathbb{R}^{m \times m} : q_{t,ii} \geq \theta, q_{t,ij}\ge 0, \sum_{j\in[m]} q_{t,ij}=1 \big\}$ under $\mathcal H_1$.  
\end{enumerate}
\end{proposition}

\subsection{Complete inheritance}
\label{subsec:gumbel-complete}

We choose the test statistic $Y_t$ as the random number $U_{t,\omega_t}$, which corresponds to the selected token $\omega_t$ at step $t$. {In this case, it is a statistic whose distribution does not rely on $\mathbf{P}_t$ under the null hypothesis.} First, we obtain the distribution of $Y_t$ under the null $\mathcal{H}_0$ and the alternative $\mathcal{H}_1$, a result already established in the literature \cite{piet2024markwordsanalyzingevaluating}.

\begin{theorem}\label{thm:cite}
(Distribution of $Y_t$, adapted from \cite{piet2024markwordsanalyzingevaluating}).
Under $\mathcal{H}_0$, $Y_t \sim U(0,1)$, and $\mathbb{P}_{\mathcal{H}_0}(Y_t \leq r | \mathbf{P}_t) = r$, for $r \in [0,1]$. Under $\mathcal{H}_1$, $\mathbb{P}_{\mathcal{H}_1}(Y_t \leq r | \mathbf{P}_t) = \sum_{\omega \in \mathcal{W}} P_{t,\omega} r^{{1}/{P_{t,\omega}}}$, for $r \in [0,1]$. 
\end{theorem}

Next, we derive the optimal score function and threshold value. We choose the distribution class as $\mathcal{P}_\Delta:=\{\mathbf{P}:\max_{i\in[m]}P_i\leq 1-\Delta\}$ for $\Delta\in[0,1-m^{-1}]$.  

\begin{theorem}\label{optimscore-complete} 
(Optimal score function and threshold value).
\vspace{-0.1in}
\begin{enumerate}[(a)]
\item For both rejection rule design settings,  the optimal score function is 
\begin{align*}
h_{\Delta}^{*}(r)=\log\left( \left\lfloor \frac{1}{1-\Delta} \right\rfloor r^{\frac{\Delta}{1-\Delta}}+r^{\frac{\Tilde{\Delta}}{1-\Tilde{\Delta}}}\right),
\end{align*}
where $\Tilde{\Delta}=(1-\Delta)\cdot\lfloor\frac{1}{1-\Delta}\rfloor$.
Moreover, the worst case in the minimax problem is attained at the following least-favorable NTP distribution in $\mathcal{P}_\Delta$,
\begin{align*}
\mathbf{P}^{*} = \left( 1-\Delta,\ldots,1-\Delta,1-(1-\Delta)\cdot\left\lfloor\frac{1}{1-\Delta}\right\rfloor,0,\ldots \right).
\end{align*}

\item For the setting of fixed type I error with $\alpha$, the optimal threshold is $\gamma_n=n\cdot\mathbb{E}_0h_{\Delta}^*(Y)+\Phi^{-1}(1-\alpha)\sqrt{n\cdot\mathrm{Var}_0(h_{\Delta}^*(Y))}$. For the setting of minimizing the sum of type I and type II errors, the optimal threshold is $\gamma_n = \log{\alpha^*} / {(1-\alpha^*)}$, where $\alpha^*$ is the solution to $\inf_{\alpha^*\in (0,1)}$ $\int_{0}^{1}\left( \left\lfloor \frac{1}{1-\Delta} \right\rfloor r^{\frac{\Delta}{1-\Delta}}+r^{\frac{\Tilde{\Delta}}{1-\Tilde{\Delta}}}\right)^{\alpha^*} d r$.
\end{enumerate}
\end{theorem}

Finally, we establish the optimality guarantee for our proposed test, in that the asymptotic sum of type I and type II errors for our test matches the lower bound across all possible tests.

\begin{theorem} \label{optimality-complete}
(Optimality). 
For a given test $\psi$, define the asymptotic type II error as $f(\psi) = \limsup_{n\to\infty}e_{\psi,2}^{\frac{1}{n}}$, and the asymptotic sum of type I and type II errors as $g(\psi) = \limsup_{n\to\infty}(e_{\psi,1}+e_{\psi,2})^{\frac{1}{n}}$. Denote the test in Theorem \ref{optimscore-complete}(a),(b) by $\psi_1$ and $\psi_2$, respectively. 
\vspace{-0.1in}
\begin{enumerate}[(a)]
\item (Fixed Type I error). Denote the class of all tests whose type I error are lower than $\alpha$ by $\Psi_{\alpha}$. When ${1}/{(1-\Delta)}$ is an integer, $m=\Omega(n)$, and $m= o(\exp{\sqrt{n}})$, or when $\Delta$ is a function of $n$, and ${1}/{\Delta(n)}=\Omega(m)$, then $\inf_{\psi\in \Psi_{\alpha}}f(\psi)=f(\psi_1)-O({1}/{m})$.

\item (Sum of Type I and Type II errors). When ${1}/{(1-\Delta)}$ is an integer, or when $\Delta$ is a function of $n$, and ${1}/{\Delta(n)} = \Omega(m)$, then $\inf_{\psi}g(\psi)=g(\psi_2)-O\left( {1}/{m} \right)$. 
\end{enumerate}
\end{theorem}

Theorem \ref{optimality-complete}(a) shows that, among all tests with type I error not exceeding  $\alpha$, our proposed test $\psi_1$ is asymptotically optimal in terms of the worst-case efficiency. The gap between the asymptotic type II error of $\psi_1$ and the optimal test is $O\left({1}/{m}\right)$. Given that $m$ is typically very large, our test is effectively optimal in practice. Theorem \ref{optimality-complete}(b) shows that, among all tests, our proposed test $\psi_2$ minimizes the asymptotic sum of type I and type II errors. The gap between $g(\psi_2)$ and the theoretical minimum $\inf_{\psi}g(\psi)$ is also $O\left({1}/{m}\right)$. 

In the proof of Theorem \ref{optimality-complete}, a key challenge lies in characterizing $\inf_{\psi\in\Psi_{\alpha}} f(\psi)$ and $\inf_\psi g(\psi)$. Take bounding $\inf_{\psi\in\Psi_{\alpha}} f(\psi)$ as an example. By definition, $\inf_{\psi\in\Psi_{\alpha}} f(\psi)$ can be expressed in a minimax form, $
\inf_\psi \lim_{n \to \infty} \sup_{\mathbf{P}_{1:n}} e_{\psi,2}^{{1}/{n}}$, where $e_{\psi,2}$ is the type II error. A naive lower bound, obtained by setting all $\mathbf{P}_i = \mathbf{P}^*$, turns out to be too loose to match the upper bound. To obtain a sharp lower bound, we construct a specific distribution over $\mathbf{P}_{1:n}$, by sampling each $\mathbf{P}_i$ independently from a carefully chosen mixture over a finite set of least-favorable NTP distributions $\mathbf{R}_i$. This randomized construction enables us to evaluate the type II error analytically. However, this approach also introduces significant technical challenges. Even under randomized $\mathbf{P}_i$, computing the corresponding type II error is non-trivial. To address this, we leverage symmetry among the sampled distributions to simplify combinatorial counting and identify the optimal rejection region. We then compute the limiting type II error, establish a tight lower bound matching the upper bound up to $O(1/m)$, and finally the minimax optimality.

Moreover, in Theorem \ref{optimality-complete}, we treat $1/(1 - \Delta) \in \mathbb{N}$ and $\Delta = \Delta_n$ separately. This is because the least favorable NTP distribution we construct is the equal-probability mixture of $(\underbrace{1-\Delta, \ldots, 1-\Delta}_{1/(1-\Delta) \text{ times}}, 0, \ldots, 0)$ and its permutations when $1/(1 - \Delta)$ is an integer, and $(\Delta,1-\Delta,0,\ldots,0)$ and its permutations when $\Delta$ is a function of $n$. For the latter, the structure of the least-favorable NTP distribution $\mathbf{P}$ changes, and its support reduces to two atoms, leading to a different form of the optimal rejection region.

\subsection{Partial inheritance}

We continue to use the test statistic $Y_t$ as the random number $U_{t,\omega_t}$, and obtain the distribution of $Y_t$ under the null and alternative hypotheses. 

\begin{theorem}\label{thm:partial:gumbel}
(Distribution of $Y_t$). Under $\mathcal{H}_0$, $\mathbb{P}_{\mathcal{H}_0}(Y_t \leq r | \mathbf{P}_t, \mathbf{Q}_t) = r$, for $r \in [0, 1]$. Under $\mathcal{H}_1$, $\mathbb{P}_{\mathcal{H}_1}(Y_t \leq r | \mathbf{P}_t, \mathbf{Q}_t) = \sum_{i=1}^{n} \sum_{j \neq i} \{p_i/(1 - p_j)\} (r - r^{{1}/{p_j}}  p_j) q_{ij} + \sum_{i=1}^{n} p_i r^{{1}/{p_i}} q_{ii}$, for $r \in [0, 1]$. Moreover, $\mathbb{P}_{\mathcal{H}_1}(Y_t \leq r | \mathbf{P}_t, \mathbf{Q}_t) \leq r$. 
\end{theorem}

Next, we derive the optimal score function and threshold value. We choose the distribution class as $\mathcal{P}_\Delta := \{\mathbf{P} : \max_{\omega \in \mathcal{W}} P_\omega \leq 1 - \Delta \}, \Delta \in (0, 1)$, and $\mathcal{Q}_\theta := \{\mathbf{Q} : q_{ii} \geq \theta \}$, for $\theta \in (1/2, 1)$. 

\begin{theorem}\label{optimscore-partial} 
(Optimal score function and threshold value).
\vspace{-0.1in}
\begin{enumerate}[(a)]
\item For both rejection rule design settings,  the optimal score function is
\begin{align*}
h_{\Delta}^{*}(r) = 
\begin{cases} 
\log\left(\frac{1 - \theta}{\Delta} + \left(\left\lfloor \frac{1}{1 - \Delta} \right\rfloor \theta + \frac{1}{\Delta} \theta - \frac{1}{\Delta} \right) r^{\frac{\Delta}{1 - \Delta}} + \theta r^{\frac{\Tilde{\Delta}}{1 - \Tilde{\Delta}}}\right) & \text{if } \Delta \geq \frac{1}{2}, \\
\log\left( 2(1 - \theta) + (2\theta - 1)\left(r^{\frac{1 - \Delta}{\Delta}} + r^{\frac{\Delta}{1 - \Delta}}\right) \right) & \text{if } \Delta < \frac{1}{2}.
\end{cases}
\end{align*}
Moreover, the worst case in the minimax problem is attained at the following least-favorable NTP distribution in $\mathcal{P}_\Delta$ and the feature matrix, 
\vspace{-0.1in}
\begin{equation*}
\mathbf{P}^{*} = \left( 1 - \Delta, \ldots, 1 - \Delta, 1 - (1 - \Delta) \left\lfloor \frac{1}{1 - \Delta} \right\rfloor, 0, \ldots \right), \;
\mathbf{Q}^{*} = \begin{pmatrix}
        \theta & 1 - \theta & 0 & \cdots & 0 \\
        1 - \theta & \theta & 0 & \cdots & 0 \\
        1 - \theta & 0 & \theta & \cdots & 0 \\
        \vdots & \vdots & \vdots & \ddots & \vdots \\
        1 - \theta & 0 & 0 & \cdots & \theta
\end{pmatrix}. 
\end{equation*}

\item For the setting of fixed type I error with $\alpha$, the optimal threshold is $\gamma_n=n\cdot\mathbb{E}_0h_{\Delta}^*(Y)+\Phi^{-1}(1-\alpha)\sqrt{n\cdot\mathrm{Var}_0(h_{\Delta}^*(Y))}$. For the setting of minimizing the sum of type I and type II errors, the optimal threshold is $\gamma_n = \log\{\alpha^*/(1 - \alpha^*)\} if \Delta \geq 1/2$, and $\log\{\beta^*/(1 - \beta^*)\}$ otherwise, where $\alpha^*, \beta^*$ are the solutions to:
\begin{align*}
\inf_{\alpha^* \in (0, 1)} \int_0^1 \left\{ \frac{1 - \theta}{\Delta} + \left( \left\lfloor \frac{1}{1 - \Delta} \right\rfloor \theta + \frac{1}{\Delta} \theta - \frac{1}{\Delta} \right) r^{\frac{\Delta}{1 - \Delta}} + \theta r^{\frac{\Tilde{\Delta}}{1 - \Tilde{\Delta}}} \right\}^{\alpha^*} dr \\
\inf_{\beta^* \in (0, 1)} \int_0^1 \left\{ 2(1 - \theta) + (2\theta - 1)\left(r^{\frac{1 - \Delta}{\Delta}} + r^{\frac{\Delta}{1 - \Delta}}\right) \right\}^{\beta^*} dr.
\end{align*}
\end{enumerate}    
\end{theorem}

We note that the least favorable $\mathbf{Q}^{*}$ in Theorem \ref{optimscore-partial} is not unique. In fact, all feature matrices $\mathbf{Q}$ that achieve the worst-case efficiency satisfy the following conditions: $q_{ii} = \theta$; for $i \leq \left\lfloor \frac{1}{1 - \Delta} \right\rfloor$, $\sum_{j=1}^{\left\lfloor \frac{1}{1 - \Delta} \right\rfloor} q_{ij} = 1$; for $i > \left\lfloor \frac{1}{1 - \Delta} \right\rfloor$, $\sum_{j=1}^{\left\lfloor \frac{1}{1 - \Delta} \right\rfloor} q_{ij} = 1 - \theta$; and for $j > \left\lfloor \frac{1}{1 - \Delta} \right\rfloor$, $q_{ij} = 0$. We also note that, for complete inheritance, the problem can be reduced to considering only the extreme points thanks to the convexity of the objective function. However, for partial inheritance, convexity is no longer guaranteed. To address this challenge, we adopt a two-step approach. That is, we first fix the largest element of the multivariable function and analyze the convexity of the function with respect to the remaining $n-1$ variables. We then remove the restriction on the largest element, and reduce the problem to a univariate function. 

Finally, we establish the optimality guarantee for our proposed test, mirroring the optimality result for the complete inheritance setting. 
\begin{theorem}\label{thm:gumbel:partial:optimal}
(Optimality).
Denote the test in Theorem \ref{optimscore-partial}(a),(b) by $\tilde\psi_1$ and $\tilde\psi_2$, respectively. 
\vspace{-0.1in}
\begin{enumerate}[(a)]
\item (Fixed Type I error). When ${1}/{(1-\Delta)}$ is an integer, $m=\Omega(n)$, and $m= o(\exp{\sqrt{n}})$, or when $\Delta$ is a function of $n$, and ${1}/{\Delta(n)}=\Omega(m)$, then $\inf_{\psi\in \Psi_{\alpha}}f(\psi)=f(\tilde\psi_1)-O({1}/{m})$.

\item (Sum of Type I and Type II errors). 
When ${1}/{(1-\Delta)}$ is an integer, or when $\Delta$ is a function of $n$, and ${1}/{\Delta(n)} = \Omega(m)$, then $\inf_{\psi}g(\psi)=g(\tilde\psi_2)-O\left( {1}/{m} \right)$. 
\end{enumerate}
\end{theorem}

We emphasize that our tests can also be used for the problem of determining if a given text is written by a human or an LLM \cite{Boenisch_2021,Aaronson2023-Gumbel}. While the existing methods perform well when the text strictly adheres to the watermarking scheme, they may fail when modifications such as text editing and paraphrasing are applied to the generated text \cite{zhao2023provablerobustwatermarkingaigenerated}. In contrast, the detection methods we propose are inherently robust to such modifications. Even when the generated text is altered, our methods can still reliably and optimally detect the presence of the watermark. Such robustness enhances the practical utility of our detection methods in real-world scenarios.

\section{Analysis of Red-green-list Watermark}
\label{sec:redgreen}

We next apply our general hypothesis testing framework to another watermarking scheme: the red-green-list watermark \citep{kirchenbauer2024watermarklargelanguagemodels}. Similarly as in Section \ref{sec:gumbel}, we first derive the feature matrix. We then establish the distribution of the test statistic, the optimal score function, and the optimality guarantee of our proposed test. Again, we first report the results for complete inheritance in Theorems \ref{thm:rg:distribution:complete} to \ref{optimality-complete:rg}, then parallelly for partial inheritance in Theorems \ref{thm:rg:partial:distribution} to \ref{optimality-partial:rg}.

\subsection{Feature matrix}
\label{sec:feature-matrix-rg}

We first quickly review the red-green-list watermark. This technique, at each step, randomly divides the vocabulary into two parts, designated as the red list and the green list, respectively. It then selects the next token  from the green list. The resulting watermarking generation process is nondeterministic, even when conditional on $\mathbf{P}_t$ and $\zeta_t$. This is different from the Gumbel-max watermarking technique. The secret key $\zeta_t$ is a subset $D \subset \{1, 2, \ldots, m\}$ satisfying that $|D| = \gamma m$ for some $\gamma\in (0,1)$. The distribution of \( \omega_t \), given $\mathbf{P}_t$ and $\zeta_t$, can be written as $\mathcal{S}(\mathbf{P}_t, \zeta_t) = \left( p_{t,1} \mathbf{1}\{1 \in D_t\}, \ldots, p_{t,m} \mathbf{1}\{m \in D_t\} \right) / {\sum_{i \in D_t} p_{t,i}}$. That is, given $\mathbf{P}_t$ and $\zeta_t$, $\omega_t$ follows the multinomial distribution, $\mathbb{P}(\omega_t = k | \mathbf{P}_t, \zeta_t) = p_{t,k} {\mathbf{1}\{k \in D_t\}} / {\sum_{i \in D_t} p_{t,i}}$.

We next derive the feature matrix for the red-green-list watermark. Let $k = \binom{m}{\gamma m}$, and $A_1, A_2, \cdots, A_k$ denote all subsets of $\{1, 2, \cdots, m\}$ containing $\gamma m$ elements. Let {$\mathbf{S}_{\mathbf{P}_t}$} be a $k \times m$ matrix with each row being $\mathbf{P}_t$, $\mathbf{Q}_t$ be a $k \times m$ matrix with the $i$th row being $\mathbf{Q}_{t,i}=(q_{t,i1},...q_{t,m})$, and $\mathbf{M}_t$ be a $k \times m$ matrix with the $i$th row being $\mathbf{M}_{t,i} = \big( p_{t,1} \mathbf{1}\{1 \in A_i\}, p_{t,2} \mathbf{1}\{2 \in A_i\}, \ldots, p_{t,m} \mathbf{1}\{m \in A_i\} \big) / {\sum_{i \in A_i} p_{t,i}}$, for $i \in [k]$.  

\begin{proposition}\label{prop2}
(Feature matrix). Following Definition~\ref{definition:feature matrix}, for $t\in[n]$, 
\vspace{-0.1in}
\begin{enumerate}[(a)]
\item (Complete inheritance). The feature matrix is $\mathbf{S}_{\mathbf{P}_t}$ under $\mathcal H_0$, and is $\mathbf{M}_t$ under $\mathcal H_1$. 

\item (Partial inheritance). The feature matrix is $\mathbf{S}_{\mathbf{P}_t}$ under $\mathcal H_0$, and is any matrix in the class $\big\{ \mathbf{Q}_{t}\in\mathbb{R}^{k \times m} : TV(\mathbf{Q}_{t,i},\mathbf{M}_{t,i}) \leq 1-\theta, q_{t,ij}\ge 0, \sum_{j\in[m]} q_{t,ij}=1 \big\}$ under $\mathcal H_1$.
\end{enumerate}
\end{proposition}

\subsection{Complete inheritance}

We choose the test statistic $Y_t = \mathbf{1}\{\omega_t \in \zeta_t\}$, and we obtain the distribution of $Y_t$, which does not depend on $P_t$ under $\mathcal H_0$. 

\begin{theorem}\label{thm:rg:distribution:complete}
(Distribution of $Y_t$).  
Under $\mathcal{H}_0$, $\mathbb{P}(Y_t = 0)=1-\gamma$, and $\mathbb{P}(Y_t = 1) = \gamma$. Under $\mathcal{H}_1$, $\mathbb{P}(Y_t = 1) = 1$.
\end{theorem}

Theorem \ref{thm:rg:distribution:complete} highlights the distinct behavior of $Y_t$ under $\mathcal{H}_0$ versus $\mathcal{H}_1$. Under $\mathcal H_0$, $Y_t$  is a Bernoulli random variable, reflecting the random selection process for subsets in the red-green-list watermarking scheme. Under $\mathcal H_1$, $Y_t$ always equals 1, as complete inheritance ensures that every token $\omega_t$ is fully aligned with the watermarking process.

Next, we derive the optimal threshold value. We also note that, since $Y_t$ can only take the value 0 or 1, any score function $h(\cdot)$ yields a test statistic equivalent to counting the occurrences of $\{\omega_t \in \zeta_t\}$, and as such the specific form of $h(\cdot)$ becomes irrelevant.

\begin{theorem}\label{optimscore-red} 
(Optimal threshold value).
For the setting of fixed type I error with $\alpha$, the optimal threshold is $\gamma_n=n \gamma+\sqrt{n\gamma(1-\gamma)}\Phi^{-1}(1-\alpha)$. For the setting of minimizing the sum of type I and type II errors, the optimal threshold is $\gamma_n=n$.
\end{theorem}

Finally, we establish the optimality guarantee for our proposed test.

\begin{theorem} \label{optimality-complete:rg}
(Optimality). 
Denote the test in Theorem \ref{optimscore-red}(a),(b) by $\psi_1^{\prime}$ and $\psi_2^{\prime}$, respectively.
\vspace{-0.1in}
\begin{enumerate}[(a)]
\item (Fixed Type I error). $\inf_{\psi\in\Psi_{\alpha}} f(\psi) = f(\psi_1^{\prime})$.

\item (Sum of Type I and Type II errors). $\inf_{\psi} g(\psi) = g(\psi_2^{\prime})$.
\end{enumerate}
\end{theorem}

Theorem \ref{optimality-complete:rg} establishes that our proposed test achieves the infimum of the asymptotic type II error and the asymptotic sum of type I and type II errors, respectively, confirming its optimality. Its proof is considerably simpler than that of the Gumbel-max watermark. Specifically, it suffices to focus on the case where each $\mathbf{P}_i$ is $(1/m, 1/m, \dots, 1/m)$. Once the NTP distributions are fixed, deriving the optimal testing scheme reduces to analyzing a Bernoulli process under $\mathcal H_0$ and a deterministic sequence under $\mathcal H_1$. This structure ensures that the test is determined entirely by the number of occurrences of $\{\omega_t\in\zeta_t\}$, namely, $\sum_{t=1}^n Y_t$. It also  eliminates the $O(1/m)$ term in the Gumbel-max watermark case, resulting the value of $f(\psi_1')$ and $g(\psi_2')$ precisely matching the theoretical infimum $\inf_{\psi}f(\psi)$ and $\inf_{\psi}g(\psi)$, respectively.

\subsection{Partial inheritance}

We continue to use the test statistic $Y_t = \mathbf{1}\{\omega_t \in \zeta_t\}$, where the token $\omega_t$ conditional on $\zeta_i$ follows a multinomial distribution with parameter ${\mathbf{Q}_{t,i}}=(q_{t,i1}, \ldots, q_{t,im})$ , and we obtain the distribution of $Y_t$. 
 
\begin{theorem}\label{thm:rg:partial:distribution}
(Distribution of $Y_t$).  
Under $\mathcal{H}_0$, $\mathbb{P}(Y_t = 1) = \gamma$ and $\mathbb{P}(Y_t = 0) = 1-\gamma$.  Under $\mathcal{H}_1$, $\mathbb{P}(Y_t = 1) = \sum_{i=1}^{C_{m,\gamma}} C_{m,\gamma}^{-1} \sum_{j \in \xi_i} q_{ij} \geq \theta$, where $C_{m,\gamma} = \binom{m}{\gamma  m}$, and $\mathbb{P}(Y_t = 0) \leq 1 - \theta$. 
\end{theorem}

Theorem \ref{thm:rg:partial:distribution} reflects the additional complexity of partial inheritance. Under $\mathcal H_0$, $Y_t$ remains a Bernoulli random variable. However, under $\mathcal H_1$,  the probability $\mathbb{P}(Y_t = 1)$ is influenced by the distributions {$\mathbf{Q}_{t,i}$} and the constraint $TV(\omega_t|\zeta_t, \mathcal{S}(\mathbf{P}_t, \zeta_t)) \leq 1 - \theta$.

Next, we derive the optimal threshold value. 

\begin{theorem}\label{rg:partial}
(Optimal threshold value).
For the setting of fixed type I error with $\alpha$, the optimal threshold is $\gamma_n=n \gamma+\sqrt{n\gamma(1-\gamma)}\Phi^{-1}(1-\alpha)$. For the setting of minimizing the sum of type I and type II errors, the optimal threshold is $\gamma_n = \big\lceil n \{ \log(1-\gamma)-\log(1-\theta) \} / \{ \log\theta+\log(1-\gamma)$ $-\log\gamma-\log(1-\theta) \} \big\rceil$.
\end{theorem}

Finally, we establish the optimality guarantee for our proposed test.

\begin{theorem} \label{optimality-partial:rg}
(Optimality).
Denote the test in Theorem \ref{rg:partial}(a),(b) by $\tilde\psi_1^{\prime}$ and $\tilde\psi_2^{\prime}$, respectively. 
\vspace{-0.1in}
\begin{enumerate}[(a)]
\item (Fixed Type I error). $\inf_{\psi\in\Psi_{\alpha}} f(\psi) = f(\tilde\psi_1^{\prime})$.

\item (Sum of Type I and Type II errors). $\inf_{\psi} g(\psi) = g(\tilde\psi_2^{\prime})$.  
\end{enumerate}
\end{theorem}

The key to prove Theorem \ref{optimality-partial:rg} lies in analyzing the worst-case scenario for $\mathcal H_1$. After fixing $\mathbf{P}_i$ and $\mathbf{Q}_i$, the optimal test can be derived by considering the constraint on the TV distance $TV(\omega_t|\zeta_t, \mathcal{S}(\mathbf{P}_t, \zeta_t)) \leq 1 - \theta$. This leads to a feasible determination of the test statistic $\sum_{t=1}^n Y_t$ and its corresponding rejection region.

\section{Simulation Studies}
\label{sec:numerical}

\subsection{Gumbel-max watermark}
\label{subsec:gm-sim}

We first investigate the empirical performance of our proposed tests through simulations, and begin with the Gumbel-max watermark. We consider a vocabulary $\mathcal{W}$ of size $m=1000$, generate token sequences of length $n$ with $n_{\text{max}} = 3000$, and repeat the generation for 5000 times. For text generation without watermark, we sample each token from $\mathcal{W}$ uniformly. For text generation with watermark, we first randomly select a prompt, then generate the subsequent tokens based on the random seed that depends on the values of the previous five tokens. The resulting NTP distribution $\mathbf{P}_t$ satisfies multinomial distribution with the largest entry of the probability vector being $1 - \Delta^*$, and the remaining terms uniformly distributed with their sum equal to $\Delta^*$. For the complete inheritance setting, at each step $t$, we generate 1000 i.i.d.\ random variables $U_{t,\omega}$ from the uniform $(0,1)$ distribution, then generate the watermarked text following the rule $\omega_t = \arg\max_{\omega \in \mathcal{W}} {\log U_{t,\omega}} / {P_{t,\omega}}$. For the partial inheritance setting, we need to further ensure that the total variation distance between the adjusted probability distribution and the original one remains within a specified bound. Towards that end, we calculate $\mathcal{S}(\mathbf{P}_t, \zeta_t)$ at each step $t$. Since it takes the form $(0, \ldots, 1, \ldots, 0)$, where 1 appears at the position of the selected token, we only need to control the value at this selected position in $\mathcal{S}(\mathbf{P}_t, \zeta_t)$. We set the value as $1 - \theta^*$, corresponding to the upper bound of the TV distance. Then, at each step, we randomly select $\theta'$ from the interval $[\theta^*, 1]$, and the corresponding distribution of $\omega | \zeta$ satisfies that the term corresponding to 1 is $\theta'$, and the sum of the remaining terms equals $1 - \theta'$. We randomly sample those remaining terms from the interval $[0,1]$ then normalize them to sum to $1-\theta^{'}$. We set the true value of $\Delta^*$ randomly sampled from $[0.001, 0.5]$ and $\theta^* = 0.8$, whereas we set the working value of $\Delta = \{0.005, 0.01\}$ and $\theta = \{0.7, 0.8, 0.9, 0.95\}$. 

We apply our proposed tests to the generated sequences, and consider both the test of fixed type I error with $\alpha = 0.05$ and the test of minimizing the sum of type I and type II errors . For each setting, we use the theoretically optimal score function $h_\Delta^*(r)$ and corresponding threshold $\gamma_n$, as derived in Theorems~\ref{optimscore-complete} and~\ref{optimscore-partial}. We also compare with two baseline score functions, $h_{\text{ars}}(r) = -\log(1 - r)$ as proposed by \cite{Aaronson2023-Gumbel}, and $h_{\text{log}}(r) = \log r$. To ensure a fair comparison, we compute their corresponding optimal threshold using the procedure detailed in Appendix~\ref{append-sec:threshold}. 

\begin{figure}[t!]
\centering
\begin{tabular}{cc}
\includegraphics[width=0.4\textwidth,height=1.7in]{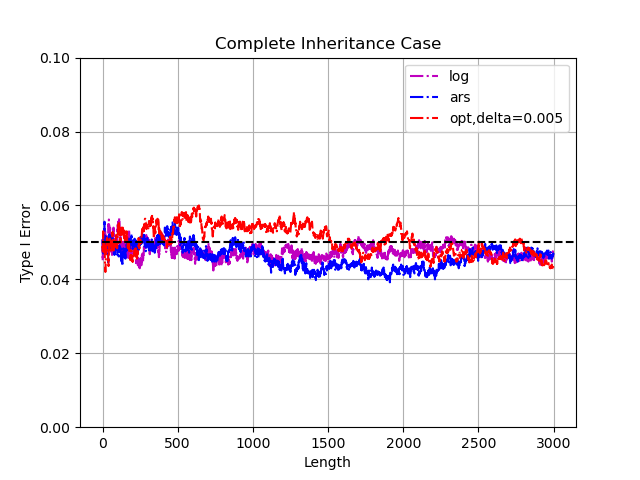} & 
\includegraphics[width=0.4\textwidth,height=1.7in]{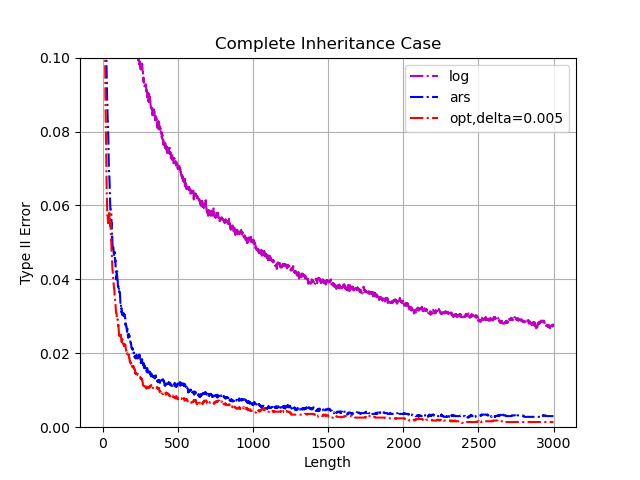} \\
\includegraphics[width=0.4\textwidth,height=1.7in]{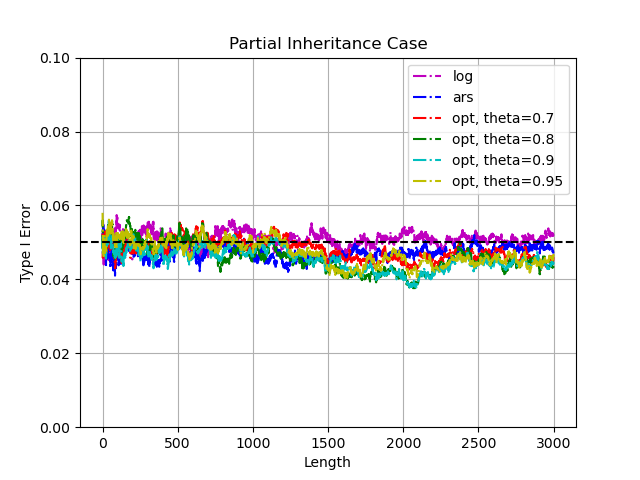} &
\includegraphics[width=0.4\textwidth,height=1.7in]{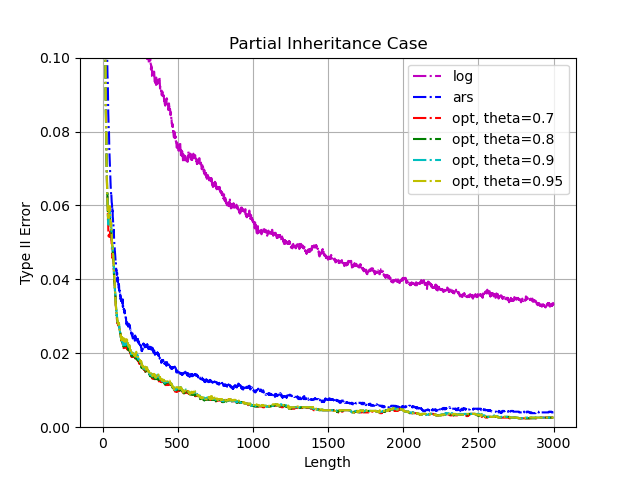} \\
\end{tabular}
\caption{Average type I and type II errors versus text length for the Gumbel-max watermark under the setting of fixed type I error, where $\Delta=0.005$ and $\theta = \{0.7, 0.8, 0.9, 0.95\}$.}
\label{fig:gm-fix}
\end{figure}

Figures \ref{fig:gm-fix} reports the average type I and type II errors  versus the text length, based on 5000 replications, under the setting of fixed type I error, where $\Delta = 0.005$ and $\theta = \{0.7, 0.8, 0.9, 0.95\}$. In this case, the rejection thresholds for the baseline score functions $h_{\text{log}}$ and $h_{\text{ars}}$ depend solely on $\alpha$, but not on $\theta$. We see that, empirically, the type I errors generally align with the nominal level of 0.05 and become increasingly closer to 0.05 as the text length increases, agreeing with our theory regarding the asymptotic control of type I error. We also see that, in both the complete and partial inheritance settings, our proposed test achieves consistently lower type II error compared to the baseline methods, highlighting the effectiveness of our approach in minimizing type II errors while maintaining type I error control. We report the results for $\Delta = 0.01$ in Appendix~\ref{append-sec:gm-sim}, and observe similar qualitative patterns. 

\begin{figure}[t!]
\centering
\begin{tabular}{cc}
\includegraphics[width=0.4\textwidth,height=1.7in]{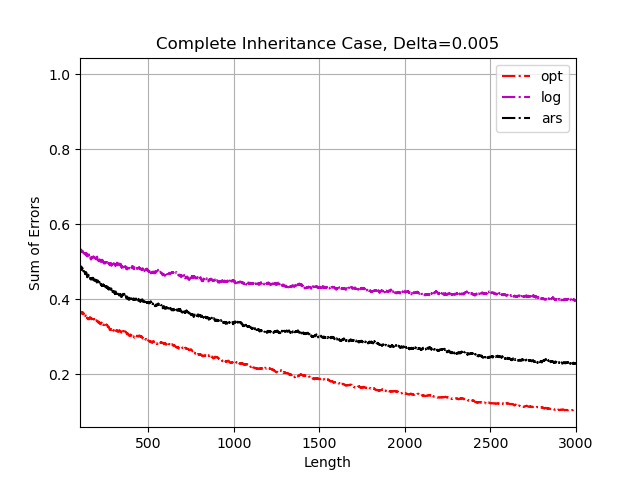} & 
\includegraphics[width=0.4\textwidth,height=1.7in]{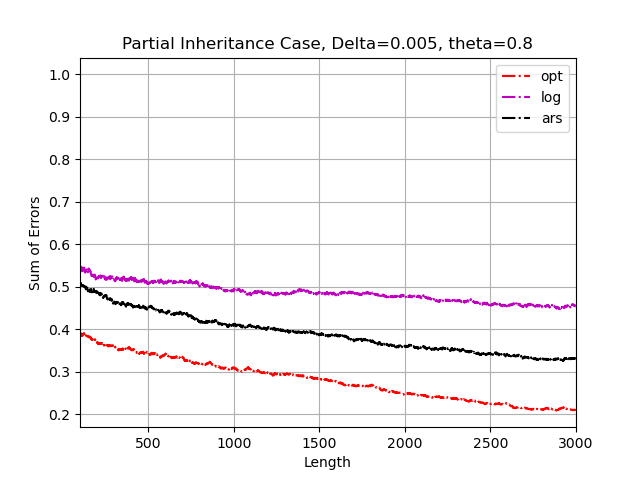} \\
\end{tabular}
\caption{Average sum of type I and type II errors versus text length for the Gumbel-max watermark under the setting of minimizing the sum of type I and type II errors, where $\Delta=0.005$ and $\theta=0.8$.}
\label{fig:gm-sum}
\end{figure}

Figures~\ref{fig:gm-sum} report the average sum of type I and type II errors versus text length, based on 5,000 repetitions, under the setting of minimizing the sum of two errors, where $\Delta=0.005$ and $\theta=0.8$. In this case, the rejection thresholds for the baseline score functions $h_{\text{log}}$ and $h_{\text{ars}}$ are numerically solved and do depend on $\theta$. We see again that our proposed test consistently achieves a smaller sum of errors than the baseline methods. We report the results for $\Delta = 0.01$ and $\theta = \{0.7, 0.8, 0.9, 0.95\}$ in Appendix~\ref{append-sec:gm-sim}, and again observe similar patterns. In general, our proposed tests remain effective and robust as long as $\Delta$ and $\theta$ are within a reasonable range.

\subsection{Red-green-list watermark}
\label{subsec:rg-sim}

We next study the red-green-list watermark. We adopt a similar simulation setup as before, and consider a vocabulary $\mathcal{W}$ of size $m=1,000$, token sequences of  length $n$ with $n_{\text{max}} = 200$. For text generation with watermark, the NTP distribution is a 1000-dimensional multinomial distribution, and the parameter $\gamma$ is set to $0.5$, meaning the red list and the green list are of equal length. For the complete inheritance setting, the secret key $\zeta_t$ is a subset of $\{1, 2, \ldots, 1000\}$ containing 500 elements. We randomly select the secret keys based on a random seed determined by the previous five tokens. For the partial inheritance setting, we set the TV distance between $\omega|\zeta$ and $\mathcal{S}(\mathbf{P}_t, \zeta_t)$ to be $1 - \theta^*$. To satisfy this constraint, we adjust the probabilities for the green and red tokens, such that the sum of probabilities for the green list equals $1 - \theta^*$, whereas the sum for the red list equals $\theta^*$. We set the true value of $\theta^* = 0.8$, and set the working value of $\theta = \{0.7, 0.8, 0.9, 0.95\}$. We report the results for $\theta^* = \{0.6, 0.7\}$ in Appendix~\ref{append-sec:rg-sim}, and observe similar qualitative patterns.

We apply our proposed tests to the generated sequences. Since for the red-green-list watermark, the specific form of the score function  $h(\cdot)$ does not affect the hypothesis testing outcome, and thus there is no need to compare different score functions. We choose the optimal threshold $\gamma_n$ following Theorems \ref{optimscore-red} and \ref{rg:partial} for complete inheritance and partial inheritance, respectively. 
 
\begin{figure}[t!]
\centering
\begin{tabular}{cc}
\includegraphics[width=0.4\textwidth,height=1.7in]{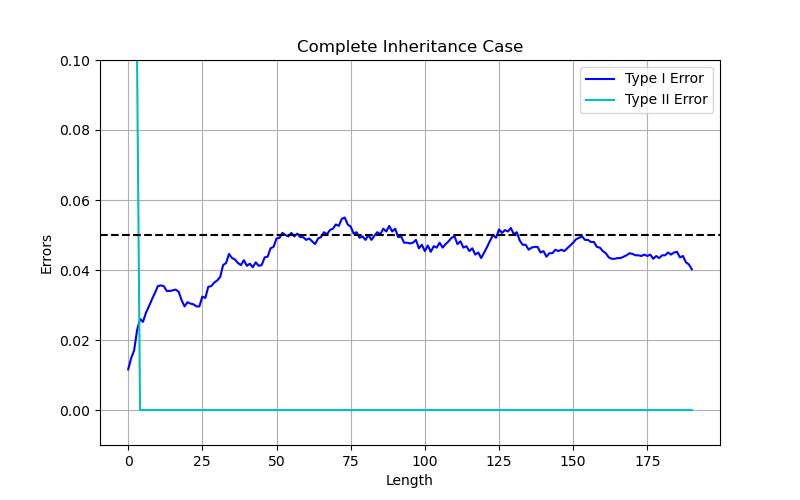} & 
\includegraphics[width=0.4\textwidth,height=1.7in]{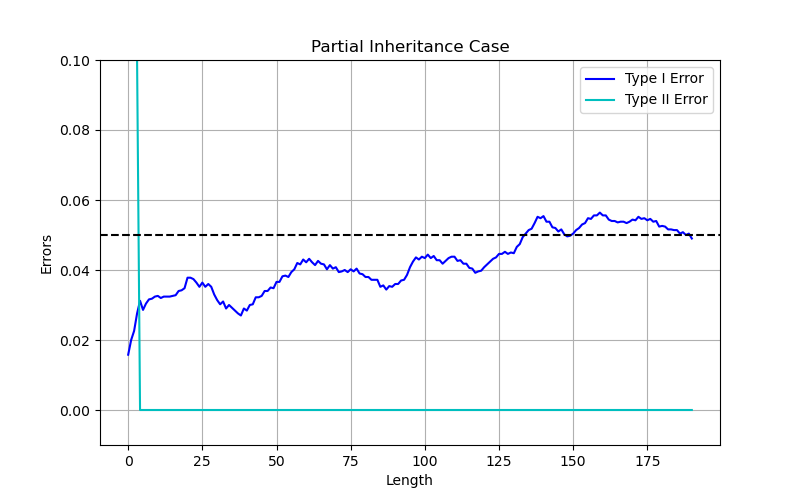} \\
\end{tabular}
\caption{Average type I and type II errors versus text length for the red-green-list watermark under the setting of fixed type I error, where $\theta^*=0.8$ and $\theta = \{0.7, 0.8, 0.9, 0.95\}$.}
\label{fig:rg-fix}
\end{figure}

Figure \ref{fig:rg-fix} reports the average type I and type II errors versus the text length, based on 5000 replications, under the setting of fixed type I error. In this case, the rejection threshold of our test depends only on the nominal level $\alpha$, but not on the value of $\theta$. We again see that the type I errors empirically align closely with the nominal level of 0.05 and become increasingly closer to 0.05 as the text length increases, agreeing with our theory. Additionally,  in both the complete and partial inheritance settings, the Type II errors decrease rapidly to 0. In fact, we can compute the type II error of our optimal test, which is given by 
\[
\mathbb{P}\left( Z < \frac{\sqrt{n}(\gamma-\theta)}{\sqrt{\theta(1-\theta)}}+\sqrt{\frac{\gamma(1-\gamma)}{\theta(1-\theta)}}\Phi^{-1}(1-\alpha) \right), 
\]
where $Z$ is a standard normal variable and $\Phi$ is its cdf. When $\theta$ is close to 1, the term on the right-hand-side of becomes large, leading to a rapid decay of the type II error. This highlights the efficiency of our test, particularly for the cases when the TV distance constraint is tight.

\begin{figure}[t!]
\centering
\begin{tabular}{cc}
\includegraphics[width=0.4\textwidth,height=1.7in]{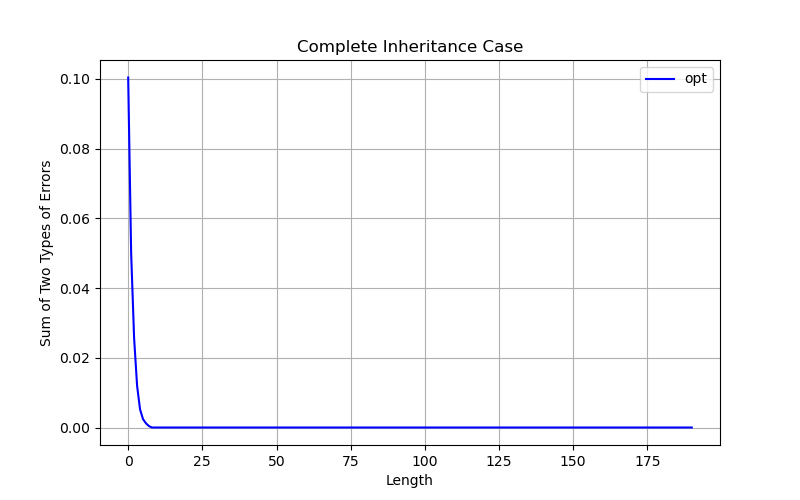} & 
\includegraphics[width=0.4\textwidth,height=1.7in]{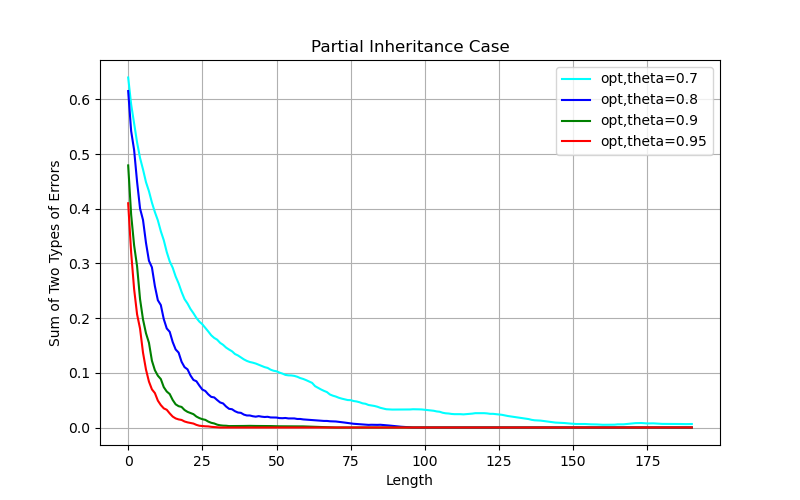} \\
\end{tabular}
\caption{Average sum of type I and type II errors versus text length for the red-green-list watermark under the setting of minimizing the sum of type I and type II errors, where $\theta^*=0.8$ and $\theta = \{0.7, 0.8, 0.9, 0.95\}$.}
\label{fig:rg-sum}
\end{figure}

Figures \ref{fig:rg-sum} reports the average sum of type I and type II errors versus the text length, based on 5,000 replications, under the setting of minimizing the sum of type I and type II errors. In this case, the rejection threshold of our test depends on $\theta$. We see that the sum of type I and type II errors drops quickly to 0, reflecting the effectiveness of our test. 

{Finally, we clarify the difference in how $\theta$ appears in Figures~\ref{fig:rg-fix} and~\ref{fig:rg-sum}. In the fixed type I error setting (Figure~\ref{fig:rg-fix}), the rejection threshold depends only on the nominal level $\alpha$, and is independent of $\theta$ (as shown in Theorem~\ref{rg:partial}). Therefore, only one curve appears in each panel. In contrast, in the setting where the goal is to minimize the total error (Figure~\ref{fig:rg-sum}), the rejection threshold depends on the chosen value of $\theta$, and hence multiple curves are plotted. Additional results for other values of $\theta$ {that are used for data generation} are reported in Appendix~\ref{append-sec:gm-sim}.}

\section{A Study with Large Language Models}
\label{sec:realdata}

We conduct a numerical study using real-world LLMs. More specifically, we use OPT-1.3B \citep{zhang2022optopenpretrainedtransformer} to generate 1500 watermarked sentences with the Gumbel-max watermark. We use these sentences to fine-tune a publicly available GPT-2 model \citep{radford2019language}. We then generate 1000 tokens, first from the original GPT-2 model, then separately from the fine-tuned GPT-2 model. The former corresponds to the scenario without data misappropriation, and the latter with data misappropriation. We repeat this experiment for 100 times.

A technical challenge in this experiment arises from the fact that OPT-1.3B and GPT-2 use different tokenization schemes. The sentences are formatted in the Alpaca instruction-response style \citep{alpaca} and used to fine-tune the GPT-2 model \citep{radford2019language}, mimicking an unauthorized reuse of LLM-generated data.  To align them, we tokenize the generated sentences with the encoder of OPT-1.3B, embed the watermarks at the token level, decode the watermarked tokens back into text using the decoder of OPT-1.3B, and use this text to fine-tune GPT-2. We apply our proposed test with the optimal score function, and continue to compare with the two baseline solutions, $h_{\text{ars}}(r) = -\log(1 - r)$ and $h_{\text{log}} = \log r$. Moreover, we set the working value of $\Delta = \{0.005, 0.01\}$ and $\theta = {0.8}$ for the partial inheritance setting.
 
\begin{figure}[t!]
\centering
\begin{tabular}{ccc}
\includegraphics[width=0.31\textwidth,height=1.8in]{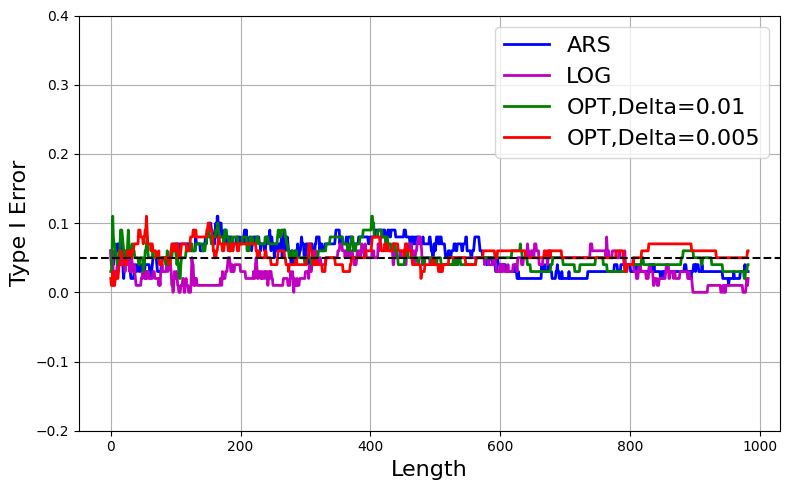} &
\includegraphics[width=0.31\textwidth,height=1.8in]{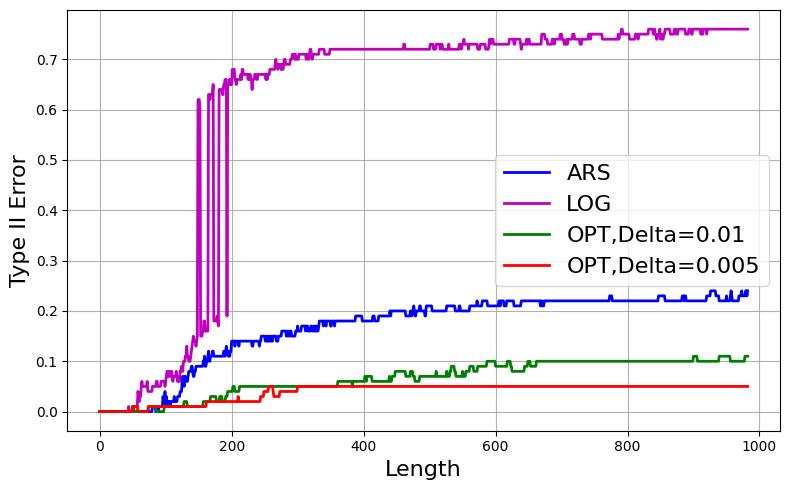} &
\includegraphics[width=0.31\textwidth,height=1.8in]{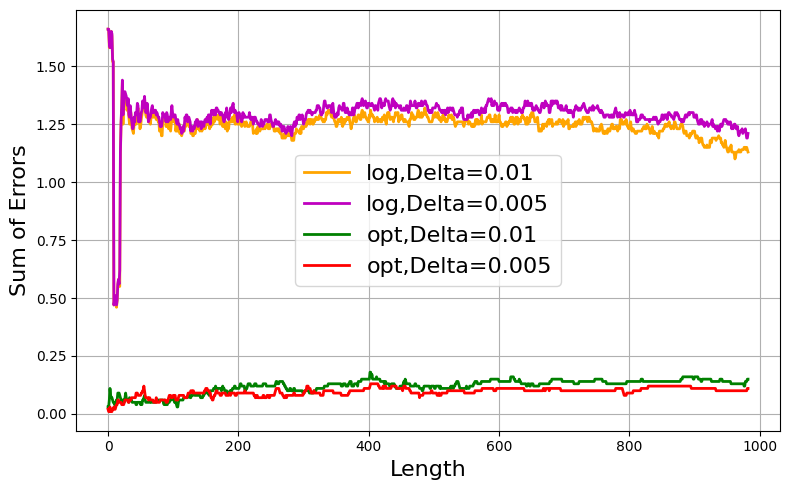} \\
\end{tabular}
\caption{Average type I error, type II error, and sum of type I and type II errors for the real LLM study.}
\label{fig:realdata}
\end{figure}

Figure \ref{fig:realdata}(a),(b) report the average type I and type II errors, based on 100 repetitions, under the setting of fixed type I error, and Figure \ref{fig:realdata}(c) reports the average sum of type I and type II errors under the setting of minimizing the sum of two errors. We see that all methods maintain the type I error around the nominal level of 0.05, but our proposed method achieves a substantially lower type II error as well as the sum of two errors, indicating a clearly superior performance over the baseline solutions by a large margin. 

This experiment thus provides empirical evidence that our detection method can reliably identify data misappropriation in practical scenarios involving real LLMs. Also importantly, our  method remains effective even when the victim and suspect models differ in architecture and tokenization, and when only limited quantities of watermarked data are used for fine-tuning. These results underscore the practical relevance and robustness of our proposed method.

\bibliographystyle{plain}
\bibliography{references}

\newpage
\appendix

\section{An Impossibility Result}
\label{append-sec:impossibility}

We show that, based solely on the generated text and without watermarking, detecting data misappropriation is impossible. Consider $N$ sequences $x_1, \ldots, x_N \stackrel{\text{\tiny i.i.d.}}{\sim} \mathbf{P}$ that are used for training, where $N$ denotes the number of training sequences. We first introduce an assumption.

\begin{assumption}
\label{impossibility}
If the training data $x_1, \ldots, x_N$ follow an identical probability distribution $\mathbf{P}$, then as $N \to \infty$, the distribution of the output of the LLM converges to $\mathbf{P}$.
\end{assumption}

It is easy to see that this assumption is natural and reasonable. Suppose we have two LLMs, $f_1$ and $f_2$, and the data used to train $f_1$ is i.i.d.\ and follows the distribution $\mathbf{P}$. We aim to detect whether the LLM $f_2$ has used the data generated by the first LLM $f_1$ for training. That is, the null hypothesis states that $f_2$ is not trained using the data generated from $f_1$, and the alternative hypothesis states the otherwise. We have the following result. 

\begin{proposition}
The summation of type-I and type-II errors converge to 1 as $N \to \infty$.
\end{proposition}

\begin{proof}
By Assumption \ref{impossibility}, the distribution of the output generated by $f_1$ converges to $\mathbf{P}$ as $N \to \infty$. Therefore, under both hypotheses, the distribution of the training data converges to $\mathbf{P}$. Then the TV distance between the output distributions under the two hypotheses converge to 0 as $N \to \infty$, making the summation of type-I and type-II errors converge to 1.  
\end{proof}

This result highlights the inherent limitations of detecting data misappropriation solely based on the generated text. Consequently, watermarking is essential to assist in reliable detection. In practice, one may infer that a model has distilled ChatGPT if it responds with ``ChatGPT" when asked "What language model are you?". However, we argue that this alone is insufficient evidence to support such a claim. Since ``ChatGPT" frequently co-occurs with ``language model" in many publicly available texts, models trained on these datasets may naturally learn to generate ``ChatGPT" as a likely response, even without direct exposure to the ChatGPT-generated data.

\section{Proofs for General Framework in Section \ref{sec:framework}}
\label{append-sec:framework}

\subsection{Proof of Theorem \ref{hypo-minimax-fixed}}
\label{pf:fix}

Since $(a)$ is a special case of $(b)$, in the following, we directly prove $(b)$. Consider the moment-generating function $\phi_{\mathbf{P},\mathbf{Q},h}(\theta):=\mathbb{E}_{1,\mathbf{P},\mathbf{Q}}e^{-\theta h(Y)}$. By Markov's inequality, for any $\theta>0$,
\begin{align*}
& 1-\mathbb{E}_1T_h(Y_{1:n}) = \mathbb{P}_1(\sum_{t=1}^{n}-h(Y_t)\geq -\gamma_{n,\alpha})\\
\leq \; & \exp\{\gamma_{n,\alpha}\theta\}\mathbb{E}_1\exp\{\sum_{t=1}^{n}-\theta h(Y_t)\} 
\leq \exp\{\gamma_{n,\alpha}\theta\}\cdot\exp\{n\log\phi_{\mathcal{P},\mathcal{Q},h}(\theta)\},
\end{align*}
where $\phi_{\mathcal{P},\mathcal{Q},h}(\theta)=\max_{\mathbf{Q}\in\mathcal{Q}}\max_{\mathbf{P}\in\mathcal{P}}\mathbb{E}_{1,\mathbf{P},\mathbf{Q}}e^{-\theta h(Y)}$.

Recall the definition of $\gamma_{n,\alpha}$ that it is the threshold chosen to ensure the significance level at $\alpha$, i.e., $\mathbb{P}_{\mathcal{H}_0}(\sum_{t=1}^{n}h(Y_t)\geq \gamma_{n,\alpha})=\alpha$. Following the strong law of large numbers, $\lim_{n\to\infty}\gamma_{n,\alpha}/n\leq\mathbb{E}_0h(Y)$. 
Therefore,
\begin{align*}
& \left\{ 1-\mathbb{E}_1T_h(Y_{1:n}) \right\}^{\frac{1}{n}} \leq \exp\left\{ \frac{\gamma_{n,\alpha}\theta}{n}+\log\phi_{\mathcal{P},\mathcal{Q},h}(\theta) \right\}, \;\; \forall\theta>0.\\
& \limsup_{n\to\infty}\left(1-\mathbb{E}_1T_h(Y_{1:n})\right)^{\frac{1}{n}} \leq \limsup_{n\to\infty}\inf_{\theta\geq 0}\exp\{\theta\gamma_{n,\alpha}/n\}\cdot\exp\{\log\phi_{\mathcal{P},\mathcal{Q},h}(\theta)\}\\
\leq \; & \inf_{\theta\geq 0}\limsup_{n\to\infty}\exp\{\theta\gamma_{n,\alpha}/n\}\cdot\exp\{\log\phi_{\mathcal{P},\mathcal{Q},h}(\theta)\}
\leq \inf_{\theta\geq 0}\exp\{\theta\mathbb{E}_0h(Y)+\log\phi_{\mathcal{P},\mathcal{Q},h}(\theta)\}.
\end{align*}

Next, we prove the tightness. By standard results from the large deviation theory, if we set all the next-token-prediction (NTP) distribution as $\mathbf{P}_k$, all the feature matrix as $\mathbf{Q}_k$, then
\begin{align} \label{30}
\lim_{n\to\infty}\mathbb{P}(T_h(Y_{1:n})=0|\mathcal{H}_1)^{\frac{1}{n}}=e^{-R_{\mathbf{P}_k,\mathbf{Q}_k}(h)}\geq e^{-\left( R_{\mathcal{P},\mathcal{Q}}(h)+\epsilon_k^{(1)} \right)},
\end{align}
where $\epsilon_k^{(1)}\to 0$ as $k\to\infty$. It then follows that
\begin{align*}
\mathbb{P}(T_h(Y_{1:n})=0|\mathcal{H}_1)\geq e^{-n\left( R_{\mathcal{P},\mathcal{Q}}(h)+\epsilon_k^{(1)}+\epsilon_{n}^{(2)} \right)},
\end{align*}
where $\epsilon_{n}^{(2)}$ denotes a sequence of positive numbers approaching to zero as $n\to\infty$ to ensure \eqref{30} to hold. Then the lower bound holds by choosing a sufficiently large $k$.

This completes the proof of Theorem \ref{hypo-minimax-fixed}.
\eop

\subsection{Proof of Theorem \ref{hypo-minimax-sum}}
\label{pf:sum}

We first present two supporting lemmas. 

\begin{lemma} \label{key}
(\cite{10.1214/aoms/1177729330}) Let $S_n=X_1+\cdots+X_n$, where $X_1,\cdots,X_n$ are i.i.d. Define $m(a)=\inf_{t}e^{-at}\mathbb{E}(e^{tX})$. If $\mathbb{E}(X)>-\infty$ and $a\leq \mathbb{E}(X)$, then $\mathbb{P}(S_n\leq na)\leq [m(a)]^n$. For $0<\epsilon<m(a)$, we have $\lim_{n\to\infty}{\{ m(a)-\epsilon \}^n} / {\mathbb{P}(S_n\leq na)}=0$. If $\mathbb{E}(X)<+\infty$ and $a\geq \mathbb{E}(X)$, then $\mathbb{P}(S_n\geq na)\leq [m(a)]^n$. For $0<\epsilon<m(a)$, we have $\lim_{n\to\infty} {\{ m(a)-\epsilon \}^n} / {\mathbb{P}(S_n\geq na)}=0$. 
\end{lemma}

\begin{lemma} \label{exchange}
(Exchangeability of infimum) For any function $f(\theta,t)$, we have $\inf_\theta\inf_{t}f(\theta,t)=\inf_{t}\inf_{\theta}f(\theta,t)$. 
\end{lemma}
\begin{proof}
We have
\begin{align*}
f(\theta,t)\geq\inf_{\theta}f(\theta,t) 
\Rightarrow \inf_t f(\theta,t)\geq \inf_t\inf_{\theta}f(\theta,t) 
\Rightarrow \inf_{\theta}\inf_t f(\theta,t)\geq \inf_t\inf_{\theta}f(\theta,t).
\end{align*}
Similarly, we have $\inf_\theta\inf_{t}f(\theta,t)\leq\inf_{t}\inf_{\theta}f(\theta,t)$.
    
This completes the proof of Lemma \ref{exchange}.     
\end{proof}

We now proceed with the proof of Theorem \ref{hypo-minimax-sum}. Again, we prove (b), as (a) is a special case. 

We first obtain the asymptotic result of the sum of two types of errors. By the Markov inequality, there exist $\theta_1>0$ and $\theta_2>0$, such that
\begin{align*}
      & \mathbb{P}_{\mathcal{H}_0}\left( \sum_{t=1}^{n}h(Y_t)\geq\gamma \right) + \mathbb{P}_{\mathcal{H}_1}\left( \sum_{t=1}^{n}h(Y_t)<\gamma \right)\\
= \; & \mathbb{P}_{\mathcal{H}_0}\left( \sum_{t=1}^{n}\theta_1h(Y_t)\geq\theta_1\gamma \right) + \mathbb{P}_{\mathcal{H}_1}\left( -\theta_2\sum_{t=1}^{n}h(Y_t)>\theta_2\gamma \right)\\
\leq \; & \frac{\mathbb{E}_{\mathcal{H}_0}\exp\{\sum_{t=1}^{n}\theta_1h(Y_t)\}}{\exp\{\theta_1\gamma\}}+\exp\{\theta_2\gamma\}\cdot\mathbb{E}_{\mathcal{H}_1}\exp\{-\sum_{t=1}^{n}\theta_2h(Y_t)\}\\
\leq \; & \frac{\exp\{n\cdot\log\mathbb{E}_0\exp{(\theta_1h(Y))}\}}{\exp\{\theta_1\gamma\}}+\exp\{\theta_2\gamma\}\cdot\exp\{n\log\mathbb{E}_{1,\mathbf{P},\mathbf{Q}}\exp{(-\theta_2h(Y))}\}.
\end{align*}
Therefore,
\begin{align*}
& \inf_{\gamma} \left\{ \mathbb{P}_{\mathcal{H}_0}\left( \sum_{t=1}^{n}h(Y_t)\geq\gamma \right) + \mathbb{P}_{\mathcal{H}_1}\left( \sum_{t=1}^{n}h(Y_t)<\gamma \right) \right\} \\
\leq \; & \inf_{\gamma} \inf_{\theta_1,\theta_2>0} \left[ \frac{\exp\{n\cdot\log\mathbb{E}_0\exp{(\theta_1h(Y))}\}}{\exp\{\theta_1\gamma\}}+\exp\{\theta_2\gamma\}\cdot\exp\{n\log\mathbb{E}_{1,\mathbf{P},\mathbf{Q}}\exp{(-\theta_2h(Y))}\} \right] \\
= \; & \inf_{\theta_1,\theta_2>0} \inf_{\gamma} \left[ \frac{\exp\{n\cdot\log\mathbb{E}_0\exp{(\theta_1h(Y))}\}}{\exp\{\theta_1\gamma\}}+\exp\{\theta_2\gamma\}\cdot\exp\{n\log\mathbb{E}_{1,\mathbf{P},\mathbf{Q}}\exp{(-\theta_2h(Y))}\} \right].
\end{align*}

By choosing 
\[
\gamma=\frac{1}{\theta_1+\theta_2}\ln{\frac{\theta_1\exp\{n\cdot\log\mathbb{E}_0\exp{(\theta_1h(Y))}\}}{\theta_2c\exp\{n\cdot\log\mathbb{E}_{1,\mathbf{P},\mathbf{Q}}\exp{(-\theta_2h(Y))}\}}}, 
\]
the above sum attains its minimum.

We then prove the tightness. By Lemma \ref{key}, we have that, for any $b\in(0,1)$, there exists $N>0$, when $n>N$,
\begin{gather*}
\mathbb{P}(S_n\geq na)\leq [m(a)]^n, \quad 
\mathbb{P}(S_n\geq na)\geq [bm(a)]^n
\end{gather*}
Then we have, 
\begin{align*}
&\inf_{\gamma} \left\{ \mathbb{P}_{\mathcal{H}_0}\left( \sum_{t=1}^{n}h(Y_t)\geq\gamma \right) +\mathbb{P}_{\mathcal{H}_1}\left( \sum_{t=1}^{n}h(Y_t)<\gamma \right) \right\} \\
\geq \; &\inf_{\gamma} \inf_{\theta_1,\theta_2>0} \left( \left[ b\cdot\frac{\exp{n\cdot\log\mathbb{E}_0\exp{\{ \theta_1h(Y) \}}}}{\exp{\theta_1\gamma}} \right]^n + \left[ b\cdot\exp{\theta_2\gamma}\cdot\exp{n\log\mathbb{E}_{1,\mathbf{P},\mathbf{Q}}\exp{\{ -\theta_2h(Y) \}}} \right]^n \right) \\
= \; &\inf_{\theta_1,\theta_2>0} \inf_{\gamma} \left( \left[ b\cdot\frac{\exp{n\cdot\log\mathbb{E}_0\exp{\{ \theta_1h(Y) \}}}}{\exp{\theta_1\gamma}} \right]^n + \left[ b\cdot\exp{\theta_2\gamma}\cdot\exp{n\log\mathbb{E}_{1,\mathbf{P},\mathbf{Q}}\exp{\{-\theta_2h(Y)\}}} \right]^n \right) \\
= \; &\left[ b\cdot\inf_{\theta_1,\theta_2>0}\exp{\frac{\theta_2}{\theta_1+\theta_2}\cdot\log\mathbb{E}_0\exp{\{  \theta_1h(Y) \}}}\cdot\exp{\frac{\theta_1}{\theta_1+\theta_2}\log\mathbb{E}_{1,\mathbf{P},\mathbf{Q}}\exp{\{ -\theta_2h(Y) \}}} \right]^n.
\end{align*}
When we set $\mathbf{P}_t, \mathbf{Q}_t$ at $\mathbf{P}^*, \mathbf{Q}^*$, such that 
\[
\sup_h S_{\mathcal{P}, \mathcal{Q}}(h)=-\inf_{h}\inf_{\theta_1,\theta_2>0}\frac{\theta_2}{\theta_1+\theta_2}\log\mathbb{E}_0\exp{\{ \theta_1h(Y) \}} \nonumber + \; \frac{\theta_1}{\theta_1+\theta_2}\log\sup_{\mathbf{P},\mathbf{Q}\in\mathcal{P}, \mathcal{Q}}\mathbb{E}_{1,\mathbf{P},\mathbf{Q}}\exp{\{ -\theta_2h(Y) \}},
\]
we get the desired result, which completes the proof of Theorem \ref{hypo-minimax-sum}. 
\eop

\section{Proofs for Gumbel-max Watermark in Section \ref{sec:gumbel}}
\label{append-sec:gumbel}

\subsection{Proof of Proposition~\ref{prop1}}

For the complete inheritance setting, under $\mathcal{H}_1$, $\mathcal{S}(\mathbf{P},\zeta)$ can only take values $e_1,e_2,\ldots,e_m$, where $e_k$ is the $m$-dimensional multinomial distribution satisfying the $k$-th term being 1 and the remaining terms being 0.

For the partial inheritance setting, under $\mathcal{H}_1$, the TV distance between $\mathbf{Q}_i$ and $\mathcal{S}(\mathbf{P},\zeta)$ has an upper bound $1-\theta$, so the term corresponding to 1 in $\mathcal{S}(\mathbf{P},\zeta)$ is larger than $\theta$.

This completes the proof of Proposition~\ref{prop1}.
\eop

\subsection{Proof of Theorem~\ref{thm:cite}}

This result is adapted from \cite{piet2024markwordsanalyzingevaluating}. For completeness, we outline the proof here.

Suppose $\zeta = (U_\omega)_{\omega \in \mathcal{W}}$ with each $U_{\omega}$ is i.i.d.\ sampled from $U(0, 1)$. We split the probability $\mathbb{P}(U_{\omega_t} \leq r)$ into the sum of probabilities of $|\mathcal{W}|$ disjoint events as follows:
\begin{align*}
\mathbb{P}(U_{\omega_t} \leq r) &= \sum_{\omega \in \mathcal{W}} \mathbb{P}(U_{\omega_t} \leq r, \omega_t = \omega) 
= \sum_{\omega \in \mathcal{W}} \mathbb{P}(U_{\omega} \leq r, \omega_t = \omega).
\end{align*}
By definition, we have that
\begin{align*}
    \mathbb{P}(U_{\omega} \leq r, \omega_t = \omega)
    &= \mathbb{P}\left(U_{\omega} \leq r, \omega = \argmax_{j \in \mathcal{W}} U_{j}^{1/P_{t,j}}\right) \\
    &= \mathbb{P}\left(U_{\omega} \leq r, U_{j} \leq U_{\omega}^{P_{t,j}/P_{t,\omega}} \text{for all } j \neq \omega \right) \\
    &= \int_0^r \int_{[0, 1]^{|\mathcal{W}|-1}} 
        \mathbf{1}_{\{ u_{j} \leq u_{\omega}^{P_{t,j}/P_{t,\omega}} \, \text{for all } j \neq \omega \}} 
        \, d u_1 \ldots d u_{\omega-1} d u_{\omega+1} \ldots d u_{|\mathcal{W}|} \, d u_\omega \\
    &= \int_0^r \prod_{j \neq \omega} u_{\omega}^{P_{t,j}/P_{t,\omega}} \, d u_\omega 
    = \int_0^r u_{\omega}^{1/P_{t,\omega}-1} \, d u_\omega 
    = P_{t,\omega} \cdot r^{1/P_{t,\omega}},
\end{align*}
where the third equality follows from the independence of $U_{1}, \ldots, U_{|\mathcal{W}|}$.

This completes the proof of Theorem~\ref{thm:cite}.
\eop

\subsection{Proof of Theorem \ref{optimscore-complete}}

The proof of (a) follows \cite{li2024statisticalframeworkwatermarkslarge}.

We next prove (b). It suffices to consider the following optimization problem,
\begin{align}
\inf_{h}\inf_{\theta_1,\theta_2>0}\frac{\theta_2}{\theta_1+\theta_2}\log\mathbb{E}_0\exp{(\theta_1h(Y))}+\frac{\theta_1}{\theta_1+\theta_2}\log\mathbb{E}_{1,\mathbf{P}}\exp{(-\theta_2h(Y))}. \label{c.3}
\end{align}

Suppose $(\alpha^{*}, p^{*})$ is the solution to the minimax problem
\begin{align*}
\inf_{\alpha>0}\sup_{p\in\mathcal{P}}\int p_0^{\frac{k}{1+k}}(y)p^{\frac{1}{1+k}}(y)dy.
\end{align*}
By the Holder inequality, we have that, for a distribution class $\mathcal{P}$, the optimal score function of \eqref{c.3} is
\begin{align*}
        h(y)=\frac{1}{1+\alpha^{*}}\log\frac{p^{*}(y)}{p_0(y)}.
\end{align*}
For this score function, the optimal threshold is 
\[
\gamma=\frac{1}{\theta_1+\theta_2}\ln{\frac{\theta_1\exp{n\cdot\log\mathbb{E}_0\exp{(\theta_1h(Y))}}}{\theta_2c\exp{n\cdot\log\mathbb{E}_{1,\mathbf{P}}\exp{(-\theta_2h(Y))}}}}.
\] 
Note that this score function and the threshold value are equivalent to the ones in (b), as they only differ by a constant ${1}/{(1+\alpha^{*})}$. This completes the proof of Theorem \ref{optimscore-complete}.
\eop

\subsection{Proof of Theorem~\ref{optimality-complete}}

We first prove (a) under two scenarios, one is when ${1}/{(1-\Delta)}$ is an integer, and the other is when $\Delta$ is a function of $n$.

\medskip
\noindent 
\textbf{Case 1.1}: Fixed type I error, and ${1}/{(1-\Delta)}$ is an integer. 

Our focus is to derive
\[
\sup_{\mathbf{P}_{1:n}}\inf_{\psi}\lim_{n\to\infty}(e_2)^{\frac{1}{n}}, 
\]
where $\psi$ is chosen from all tests satisfying that the type I error is smaller than or equal to the pre-specified significance level $\alpha$. We first note that
\begin{align*}
    \sup_{\mathbf{P}_{1:n}}\inf_{\psi}\lim_{n\to\infty}(e_2)^{\frac{1}{n}}\geq \inf_{\psi}\sup_{\mathbf{P}_{1:n}}\lim_{n\to\infty}(e_2)^{\frac{1}{n}}\geq \lim_{n\to\infty}\inf_{\psi}\sup_{\mathbf{P}_{1:n}}(e_2)^{\frac{1}{n}}.
\end{align*}
As such, we turn to the problem,
\begin{align*}
A:=\lim_{n\to\infty}\inf_{\psi}\sup_{\mathbf{P}_{1:n}}(e_2)^{\frac{1}{n}}.
\end{align*}
Recall that each $\mathbf{P}_i$ is a multinomial distribution. To make the detection method more powerful, we require that each term $p_{ij}$ of $\mathbf{P}_i$ is no larger than $1-\Delta$. 

There are $\binom{m}{\frac{1}{1-\Delta}}$ permutations of $(\underbrace{1-\Delta,1-\Delta,\cdots,1-\Delta}_{\frac{1}{1-\Delta}times},0,\ldots,0)$. Denote these multinomial distributions as $ {\mathbf{R}_i}, i=1,2,\cdots,\binom{m}{\frac{1}{1-\Delta}}$. Let each $\mathbf{P}_j^0$ be {a mixture of} ${\mathbf{R}_i}$, i.e., each $\mathbf{P}_j^0$ has the probability $1/\binom{m}{\frac{1}{1-\Delta}}$ to be the distribution ${\mathbf{R}_i}$, for $i=1, \ldots, \binom{m}{\frac{1}{1-\Delta}}$. Then we have that,
\begin{align}\label{eq:B}
A \geq B:=\lim_{n\to \infty}\inf_{\psi}\mathbb{P}_{\mathcal{H}_1,\mathbf{P}_{1:n}^{0}}(\psi(\omega_{1:n},\zeta_{1:n})=0)^{\frac{1}{n}}.
\end{align}

We next calculate type I and type II errors under the distribution $\mathbf{P}_{1:n}^{0}$, where each $\mathbf{P}_j^{0}$  has the equal probability $1/\binom{m}{\frac{1}{1-\Delta}}$ to be the distribution $\mathbf{R}_i, i=1, \ldots, \binom{m}{\frac{1}{1-\Delta}}$. 

We note that, under $\mathcal{H}_0$, $\omega_t$ is independent of $\zeta_t$ for all $t$. Meanwhile, under $\mathcal{H}_1$, the distribution of $\omega_t$ follows the Gumbel-max watermark rule. Formally, the hypotheses are: 
\begin{align*}
\mathcal{H}_0: \;\; \omega_t\perp\zeta_t,  \quad\quad  \mathcal{H}_1: \;\; \omega_t=\mathcal{S}(\mathbf{P}_t^0,\zeta_t), \quad\quad t=1,\ldots,n,
\end{align*}
where the decoder function $\mathcal{S}(\mathbf{P}_t^0,\zeta_t)=\argmax_{i}U_{ti}^{\frac{1}{p_{ti}^{0}}}$, $U_{ti}$'s are the $U(0,1)$ random variables generated in the step $t$ in the Gumbel-max watermarking scheme. 

We now derive the optimal rejection region. By the Neyman-Pearson lemma, we only need to calculate the likelihood ratio function. Under $\mathcal{H}_0$, $\omega_t\sim\mathbf{P}_t^0$, $\zeta_t=(U_{t1},U_{t2},\cdots,U_{tm})$, where $U_{ti} \sim U(0,1)$ and is independent from each other. We view $\omega_t$ as a random variable that takes values uniformly from $1, \ldots, m$. Therefore, the likelihood of $(\omega_t,\zeta_t)$ is ${1}/{m} \cdot 1 = {1}/{m}$, and the likelihood of $(\omega_{1:n},\zeta_{1:n})$ is $({1} / {m})^n$. Under $\mathcal{H}_1$, the relationship between $\omega_t$ and $\zeta_t$ is $\omega_t=\argmax_{i}U_{ti}^{\frac{1}{p_{ti}^{0}}}$. We consider each case of $\mathbf{R}_i$ separately, then combine all cases to compute the likelihood. 

The likelihood of $(\omega_t,\zeta_t)$ is $\frac{1}{\binom{m}{\frac{1}{1-\Delta}}} \cdot \sum_{S\subset \{1,2,\cdots,m\},|S|=\frac{1}{1-\Delta}}\mathbf{1}\{\text{$U_{t,\omega_i}$ is the largest one among $U_{t,i|i\in S}$}\}$.  Here, $U_{t,i|i\in S}$ denotes $U_{ti}$ when $i\in S$. As a result, the likelihood of $(\omega_{1:n},\zeta_{1:n})$ is 
\[
\prod_{t=1}^{n}\left(\frac{1}{\binom{m}{\frac{1}{1-\Delta}}} \cdot \sum_{S\subset \{1,2,\cdots,m\},|S|=\frac{1}{1-\Delta}}\mathbf{1}\{\text{$U_{t,\omega_t}$ is the largest one among $U_{t,i|i\in S}$}\}\right).
\]

Therefore, by the Neyman-Pearson lemma, the optimal rejection region can be written as
\begin{align}
\label{rej_1}
R = \Bigg\{ (\omega_{1:n}, & \zeta_{1:n}) : \left( \eta(1-\Delta)\cdot\binom{m-1}{\frac{1}{1-\Delta}-1} \right)^n \nonumber \\ 
& \leq\prod_{t=1}^{n} \Bigg( \sum_{S\subset \{1,2,\cdots,m\},|S|=\frac{1}{1-\Delta}}\mathbf{1}\{\text{$U_{t,\omega_t}$ is the largest one among $U_{t,i|i\in S}$}\} \Bigg) \Bigg\}.
\end{align}
where $\eta$ is a constant determined by $\alpha$.

Denote the right-hand-side of \eqref{rej_1} as $\prod_{t=1}^{n}X_t$, .i.e.,
\[
X_t := \sum_{S\subset \{1,2,\cdots,m\},|S|=\frac{1}{1-\Delta}}\mathbf{1}\{\text{$U_{t,\omega_t}$ is the largest one among $U_{t,i|i\in S}$}\}.
\]
We have $X_1,X_2,\cdots,X_n$ are i.i.d. We next derive the distribution of $X_t$ under $\mathcal{H}_0$ and $\mathcal{H}_1$. As $U_{ti}$'s are $i.i.d.$, to simplify the notation, we omit $t$ in $U_{ti}$ when we compute the likelihood below.

Under $\mathcal{H}_0$, the rank of $U_{\omega_t}$ distributes uniformly on $\{1,2,\cdots,m\}$. If its rank is smaller than $\frac{1}{1-\Delta}$, then $X_t=0$. If its rank is $k\geq\frac{1}{1-\Delta}$, then $X_t=\binom{k-1}{\frac{1}{1-\Delta}-1}$. So under $\mathcal{H}_0$, the distribution of $X_t$ is
\begin{gather*}
    \mathbb{P}(X_t=0)=\frac{\frac{1}{1-\Delta}-1}{m}\\
    \mathbb{P}(X_t=\binom{k-1}{\frac{1}{1-\Delta}-1})=\frac{1}{m},k=\frac{1}{1-\Delta},\cdots,m.
\end{gather*}

Under $\mathcal{H}_1$, the prior condition is equivalent to that there exists $\frac{1}{1-\Delta}-1$ times of $U_t$ that must be smaller than $U_{\omega_t}$. Therefore, the distribution of $X_t$ is related to the distribution of $r(U_1)|U_1\geq U_2,\cdots,U_{\frac{1}{1-\Delta}}$, where $r(U_1)$ is the rank of $U_t$ among $U_1,\cdots,U_m$. For $k\geq \frac{1}{1-\Delta}$,
\begin{align*}
    \mathbb{P}(r(U_1)=k|U_1\geq U_2,\cdots,U_{\frac{1}{1-\Delta}})&=\mathbb{P}(r(U_1)=k,U_1\geq U_2,\cdots,U_{\frac{1}{1-\Delta}})/\mathbb{P}(U_1\geq U_2,\cdots,U_{\frac{1}{1-\Delta}})
\end{align*}
We have that,
\begin{align*}
\mathbb{P}(r(U_1)&=k,U_1\geq U_2,\cdots,U_{\frac{1}{1-\Delta}})=\binom{k-1}{\frac{1}{1-\Delta}-1}/\binom{m}{\frac{1}{1-\Delta}};\\
\mathbb{P}(U_1\geq U_2,\cdots,U_{\frac{1}{1-\Delta}})&=\sum_{k=\frac{1}{1-\Delta}}^{m}\mathbb{P}(r(U_1)=k,U_1\geq U_2,\cdots,U_{\frac{1}{1-\Delta}-1})\\
&=\sum_{k=\frac{1}{1-\Delta}}^{m}\binom{k-1}{\frac{1}{1-\Delta}-1}/\binom{m}{\frac{1}{1-\Delta}}.
\end{align*}
As a result, 
\begin{align*}
    \mathbb{P}(X_t=\binom{k-1}{\frac{1}{1-\Delta}-1})=\mathbb{P}(r(U_1)=k|U_1\geq U_2,\cdots,U_{\frac{1}{1-\Delta}})=\frac{\binom{k-1}{\frac{1}{1-\Delta}-1}}{\sum_{i=\frac{1}{1-\Delta}}^{m}\binom{i-1}{\frac{1}{1-\Delta}-1}},k=\frac{1}{1-\Delta},\cdots,m.
\end{align*}

So under $\mathcal{H}_1$, the distribution of $X_t$ is
\begin{gather*}
    \mathbb{P}(X_t=\binom{k-1}{\frac{1}{1-\Delta}-1})=\frac{\binom{k-1}{\frac{1}{1-\Delta}-1}}{\sum_{i=\frac{1}{1-\Delta}}^{m}\binom{i-1}{\frac{1}{1-\Delta}-1}},k=\frac{1}{1-\Delta},\cdots,m.
\end{gather*}

Consequently, we can represent the two types of error as
\begin{align*}
    \text{type I error}&=\mathbb{P}_{\mathcal{H}_0}\left( \prod_{t=1}^{n}X_t \geq \left\{ \eta(1-\Delta)\cdot\binom{m-1}{\frac{1}{1-\Delta}-1} \right\}^n \right);\\
    \text{type II error}&=\mathbb{P}_{\mathcal{H}_1}\left( \prod_{t=1}^{n}X_t < \left\{ \eta(1-\Delta)\cdot\binom{m-1}{\frac{1}{1-\Delta}-1} \right\}^n \right).
\end{align*}

Note that 
\begin{align*}
    \alpha&\geq\mathbb{P}_{\mathcal{H}_0}\left( \sum_{i=1}^{n}\ln X_t \geq n\ln\left\{ \eta(1-\Delta) \right\} + n\ln\binom{m-1}{\frac{1}{1-\Delta}-1} \right) \\
    &=\left( 1-\frac{\frac{1}{1-\Delta}-1}{m} \right)^n\mathbb{P}\left( \sum_{i=1}^{n}\ln Y_t\geq n\ln\{ \eta(1-\Delta) \} + n\ln\binom{m-1}{\frac{1}{1-\Delta}-1} \right),
\end{align*}
where $Y_t$ obeys the following probability distribution, 
\begin{align*}
\mathbb{P}\left( Y_t=\binom{k-1}{\frac{1}{1-\Delta}-1} \right) = \frac{1}{m-\frac{1}{1-\Delta}+1}, k=\frac{1}{1-\Delta},\ldots,m.
\end{align*}
When $m=\Omega(n)$, $\left( 1-\frac{\frac{1}{1-\Delta}-1}{m} \right)^n\to 1$. Then according to the large number rule, we can choose $\hat{\eta}$ satisfying that
\begin{align*}
\ln\left( \hat{\eta}(1-\Delta)\binom{m-1}{\frac{1}{1-\Delta}-1} \right) = \mathbb{E}\ln Y+\Phi^{-1}(1-\alpha)\sqrt{\mathrm{Var}(\ln Y)/n}.
\end{align*}

Next, we derive the approximations for the expectation and the variance of $\ln Y$. 
\begin{align*}
    \mathbb{E}\ln Y&=\sum_{k=\frac{1}{1-\Delta}}^{m}\ln\binom{k-1}{\frac{1}{1-\Delta}-1}\cdot\frac{1}{m-\frac{1}{1-\Delta}+1}\\
    &=\frac{1}{m-\frac{1}{1-\Delta}+1}\sum_{k=\frac{1}{1-\Delta}}^{m}\ln\frac{(k-1)(k-2)\cdots(k-\frac{1}{1-\Delta}+1)}{(\frac{1}{1-\Delta}-1)!}\\
    &=\frac{1}{m-\frac{1}{1-\Delta}+1}\sum_{k=\frac{1}{1-\Delta}}^{m}\left\{\sum_{i=k-\frac{1}{1-\Delta}+1}^{k-1}\ln i-\ln(\frac{1}{1-\Delta}-1)! \right\}\\
    &\leq \frac{1}{m-\frac{1}{1-\Delta}+1}\cdot (\frac{1}{1-\Delta}-1)\cdot \sum_{i=1}^{m-1}\ln i-\ln(\frac{1}{1-\Delta}-1)!\\
    &\leq \frac{1}{m-\frac{1}{1-\Delta}+1}\cdot (\frac{1}{1-\Delta}-1)\cdot \int_{1}^{m}\ln xdx-\ln(\frac{1}{1-\Delta}-1)!\\
    &=(\frac{1}{1-\Delta}-1)\frac{m\ln m-m+1}{m-\frac{1}{1-\Delta}+1}-\ln(\frac{1}{1-\Delta}-1)!\\
    &=(\frac{1}{1-\Delta}-1)\ln m+o(\ln m)\\
    &=O(\ln m)
\end{align*}
\begin{align*}
    \mathrm{Var}(\ln Y)&=\mathbb{E}(\ln Y)^2-(\mathbb{E}\ln Y)^2\\
    &=\sum_{k=\frac{1}{1-\Delta}}^{m}\left[ \ln\binom{k-1}{\frac{1}{1-\Delta}-1} \right]^2\cdot\frac{1}{m-\frac{1}{1-\Delta}+1} - \left[\sum_{k=\frac{1}{1-\Delta}}^{m}\ln\binom{k-1}{\frac{1}{1-\Delta}-1}\cdot\frac{1}{m-\frac{1}{1-\Delta}+1} \right]^2
\end{align*}
For the second term, using the same method to estimate the lower bound of $\mathbb{E}\ln Y$, we can prove that $\mathbb{E}\ln Y\geq O(\ln m)$. We next derive the approximation to the first term.
\begin{align*}
    \mathbb{E}(\ln Y)^2&=\sum_{k=\frac{1}{1-\Delta}}^{m}\left[ \ln\binom{k-1}{\frac{1}{1-\Delta}-1} \right]^2\cdot\frac{1}{m-\frac{1}{1-\Delta}+1}\\
    &=\sum_{k=\frac{1}{1-\Delta}}^{m}\left[ \ln(k-1)+\ln(k-2)+\cdots+\ln(k-\frac{1}{1-\Delta}+1)-\ln(\frac{1}{1-\Delta}-1)! \right]^2\cdot\frac{1}{m-\frac{1}{1-\Delta}+1}\\
    &=\sum_{k=\frac{1}{1-\Delta}}^{m}\frac{[\ln(k-1)+\ln(k-2)+\cdots+\ln(k-\frac{1}{1-\Delta}+1)]^2}{m-\frac{1}{1-\Delta}+1}\\
    &-2\ln(\frac{1}{1-\Delta}-1)!\cdot\frac{\sum_{k=\frac{1}{1-\Delta}}^m[\ln(k-1)+\ln(k-2)+\cdots+\ln(k-\frac{1}{1-\Delta}+1)]}{m-\frac{1}{1-\Delta}+1} + \left[ \ln(\frac{1}{1-\Delta}-1)! \right]^2\\
    &\leq \frac{\sum_{k=\frac{1}{1-\Delta}}^m[(\frac{1}{1-\Delta}-1)\ln(k-1)]^2}{m-\frac{1}{1-\Delta}+1}\\
    &\leq \frac{(\frac{1}{1-\Delta}-1)^2\int_{0}^{m}\ln^2 xdx}{m-\frac{1}{1-\Delta}+1}\\
    &=O(\ln^2 m)
\end{align*}
Henceforth, we have derived the result that 
\begin{align*}
\mathbb{E}\ln Y=O(\ln m), \mathrm{Var}(\ln Y)=O(\ln^2 m).
\end{align*}
When $m= o(\exp(\sqrt{n}))$, we have $\Phi^{-1}(1-\alpha)\sqrt{n\cdot \mathrm{Var}(\ln Y)}=o(n\cdot\mathbb{E}\ln Y)$.

Combining this result and \cref{key}, we have
\begin{align*}
    B&=\inf_{s\geq 0}\exp\{-s\mathbb{E}\ln Y\}\mathbb{E}\exp\{s\ln X_t\}\\
    &=\inf_{s\geq 0}\frac{\sum_{k=\frac{1}{1-\Delta}}^{m}\binom{k-1}{\frac{1}{1-\Delta}-1}^s\cdot\binom{k-1}{\frac{1}{1-\Delta}-1}/\sum_{i=\frac{1}{1-\Delta}}^{m}\binom{i-1}{\frac{1}{1-\Delta}-1}}{e^{s\mathbb{E}\ln Y}}\\
    &\geq\inf_{s\geq 0}\frac{\sum_{k=\frac{1}{1-\Delta}}^{m}\binom{k-1}{\frac{1}{1-\Delta}-1}^{s+1}}{\sum_{i=\frac{1}{1-\Delta}}^{m}\binom{i-1}{\frac{1}{1-\Delta}-1}\cdot e^{s\cdot\left(\frac{(\frac{1}{1-\Delta}-1)\cdot(m\ln m-m+1)}{m-\frac{1}{1-\Delta}+1}-\ln(\frac{1}{1-\Delta}-1)!\right)}}\\
    &=\inf_{s\geq 0}\frac{\sum_{k=\frac{1}{1-\Delta}}^m[(k-1)(k-2)\cdots(k-\frac{1}{1-\Delta}+1)]^{s+1}}{\sum_{k=\frac{1}{1-\Delta}}^m(k-1)(k-2)\cdots(k-\frac{1}{1-\Delta}+1)\cdot\exp\{\frac{m\ln m-m+1}{m-\frac{1}{1-\Delta}+1}\cdot(\frac{1}{1-\Delta}-1)\cdot s\}}\\
    &\geq \inf_{s\geq 0}\frac{\sum_{k=\frac{1}{1-\Delta}}^m((k-\frac{1}{1-\Delta}+1)/m)^{(s+1)(\frac{1}{1-\Delta}-1)}\cdot m^{-s(\frac{1}{1-\Delta}-1)}}{\sum_{k=\frac{1}{1-\Delta}}^m((k-1)/m)^{\frac{1}{1-\Delta}-1}\cdot\exp\{\frac{m\ln m-m+1}{m-\frac{1}{1-\Delta}+1}\cdot(\frac{1}{1-\Delta}-1)\cdot s\}}\\
    &\geq \inf_{s\geq 0}\frac{\int_{0}^{\frac{m-\frac{1}{1-\Delta}}{m}}x^{(s+1)(\frac{1}{1-\Delta}-1)}dx\cdot m^{s-\frac{s}{1-\Delta}}}{\int_{\frac{1}{m(1-\Delta)}}^{1}x^{\frac{1}{1-\Delta}-1}dx\cdot\exp\{\frac{m\ln m-m+1}{m-\frac{1}{1-\Delta}+1}\cdot(\frac{1}{1-\Delta}-1)\cdot s\}}\\
    &=\inf_{s \geq 0}\frac{1}{1-\Delta s}(\frac{m-\frac{1}{1-\Delta}}{m})^{s\cdot(\frac{1}{1-\Delta}-1)}\frac{m^{s\cdot(1-\frac{1}{1-\Delta})}}{\exp\{\frac{m\ln m-m+1}{m-\frac{1}{1-\Delta}+1}\cdot(\frac{1}{1-\Delta}-1)\cdot s\}}\\
    &=\inf_{s \geq 0}\frac{1}{1-\Delta s}(\frac{m-\frac{1}{1-\Delta}}{m})^{s\cdot(\frac{1}{1-\Delta}-1)}\exp\left\{(\frac{1}{1-\Delta}-1)\cdot s\cdot (\ln m-\frac{m\ln m-m+1}{m-\frac{1}{1-\Delta}+1}) \right\}
\end{align*}
When $m\to\infty$, the above expression tends to $\inf_{s\geq 0}\frac{1}{e^{\frac{s\Delta}{1-\Delta}}\cdot(1-\Delta s)}$.

Next we compute the asymptotic type II error of our test. By Theorem \ref{hypo-minimax-fixed}, we have
\begin{align*}
    & \lim_{n\to\infty}\sup_{\mathbf{P}_{1:n}}\left( \mathbb{P}_{\mathcal{H}_1}(T_h(Y_{1:n})=0) \right)^{\frac{1}{n}} = e^{-R_{\mathcal{P}}(h)}\\
= \; & \exp{\inf_{s\geq 0}\sup_{\mathbf{P}\in\mathcal{P}}\{t\mathbb{E}_0h(Y)+\log\phi_{\mathbf{P},h}(s)\}}\\
= \; & \exp{\inf_{s\geq 0}\left\{ s\cdot(\ln\frac{1}{1-\Delta}-\frac{\Delta}{1-\Delta})+\ln\left[ (\frac{1}{1-\Delta})^{1-s}\cdot\frac{1}{\frac{\Delta}{1-\Delta}\cdot(1+s)+1} \right] \right\}}\\
= \; & \inf_{s\geq 0}\exp\left\{\ln\frac{1}{1-\Delta}-s\cdot\frac{\Delta}{1-\Delta}-\ln\left(\frac{1}{1-\Delta}+\frac{\Delta}{1-\Delta}\cdot s \right) \right\}\\
= \; & \inf_{s\geq 0}\frac{1}{e^{\frac{s\Delta}{1-\Delta}}\cdot(1-\Delta s)}
\end{align*}
which is exactly the lower bound of $B$ as defined in \eqref{eq:B}).

\medskip
\noindent 
\textbf{Case 1.2}: Fixed type I error, and $\Delta$ is a function of $n$.

As $n$ increases, $1-\Delta$ tends to 1. We first calculate the asymptotic type II error. 
\begin{align*}
    \limsup_{n\to\infty}\mathbb{P}_{\phi_0,\mathcal{H}_1}(R^c)&=\exp\{-R_{\mathcal{P}}(h)\}\\
    R_{\mathcal{P}}(h)&=-\inf_{h}\sup_{\mathbf{P}\in\mathcal{P}}\mathbb{E}_0h(Y)+\log \mathbb{E}_{1,\mathbf{P}}e^{-h(Y)} =-\int_{0}^{1}\log (y^a+y^{\frac{1}{a}})dy,
\end{align*}
where $a=\frac{1-\Delta}{\Delta}$.

As a direct consequence of Neyman-Pearson lemma, under the condition that each $\mathbf{P}_j$ has the equal probability $1/m(m-1)$ to be the distribution $(\Delta,1-\Delta,0,0,\cdots,0)$ and its permutation, the best rejection region can be represented as
\begin{align}
\label{fix_function}
R = \left\{(\omega_{1:n},\zeta_{1:n}):(\eta\cdot(m-1))^n\leq\prod_{t=1}^{n}\left(\sum_{j=1}^{m-1}(\mathbf{1}\{U_{i,m}\geq U_{i,j}^{a}\}+\mathbf{1}\{U_{i,m}\geq U_{i,j}^{\frac{1}{a}}\})\right)\right\},
\end{align}
where $\eta$ is determined by the pre-specified constant $\alpha$, and $U_{i,j}$'s are i.i.d.\ random variables, $U_{i,j}\sim U(0,1),i=1,2,\cdots,n,j=1,2,\ldots,m$.

Denote the right-hand-side of \eqref{fix_function} as $\prod_{t=1}^{n}X_t$. As we can see, $X_1,X_2,\cdots,X_n$ are i.i.d. We compute the following probabilities,
\begin{align*}
    &\mathbb{P}\left( \sum_{i=1}^{m-1}\mathbf{1}\{U_m\geq U_{i}^{a}\}=y,\sum_{i=1}^{m-1}\mathbf{1}\{U_m\geq U_{i}^{\frac{1}{a}}\}=x|U_m\geq U_{1}^{\frac{1}{a}} \right) \\
    = \; &(a+1)\int_{0}^{1}\binom{m-2}{x-1}\cdot\binom{m-x-1}{y-x}(r^a)^x\cdot(r^{\frac{1}{a}}-r^a)^{y-x}\cdot(1-r^{\frac{1}{a}})^{m-y-1}dr,y\geq x\geq 1\\
    &\mathbb{P}\left( \sum_{i=1}^{m-1}\mathbf{1}\{U_m\geq U_{i}^{a}\}=y,\sum_{i=1}^{m-1}\mathbf{1}\{U_m\geq U_{i}^{\frac{1}{a}}\}=x|U_m\geq U_{1}^{a} \right)\\
    = \; &\frac{a+1}{a}\int_{0}^{1}\binom{m-2}{x-1}(r^a)^x\cdot(1-r^{\frac{1}{a}})^{m-y-1}dr,y=x\geq 1\\
    &\mathbb{P}\left( \sum_{i=1}^{m-1}\mathbf{1}\{U_m\geq U_{i}^{a}\}=y,\sum_{i=1}^{m-1}\mathbf{1}\{U_m\geq U_{i}^{\frac{1}{a}}\}=x|U_m\geq U_{1}^{a} \right)\\
    = \; &\frac{a+1}{a}\int_{0}^{1}\binom{m-2}{x}\cdot\binom{m-x-2}{y-x-1}(r^a)^x\cdot(r^{\frac{1}{a}}-r^a)^{y-x}\cdot(1-r^{\frac{1}{a}})^{m-y-1}dr,y>x.
\end{align*}

Henceforth, under $\mathcal{H}_0$, the distribution of $X_t$ is
\begin{align*}
    &\mathbb{P}_{\mathcal{H}_0}\left( \sum_{i=1}^{m-1}\mathbf{1}\{U_m\geq U_{i}^{a}\}=y,\sum_{i=1}^{m-1}\mathbf{1}\{U_m\geq U_{i}^{\frac{1}{a}}\}=x \right)\\
= \; &\int_{0}^{1}\binom{m-1}{x}\binom{m-1-x}{y-x}(r^a)^x\cdot(r^{\frac{1}{a}}-r^a)^{y-x}\cdot(1-r^{\frac{1}{a}})^{m-1-y}dr,y\geq x.
\end{align*}
Under $\mathcal{H}_1$, the distribution of $X_t$ is
\begin{align*}
    &\mathbb{P}_{\mathcal{H}_1}\left( \sum_{i=1}^{m-1}\mathbf{1}\{U_m\geq U_{i}^{a}\}=y,\sum_{i=1}^{m-1}\mathbf{1}\{U_m\geq U_{i}^{\frac{1}{a}}\}=x \right)\\
= \; &\frac{1}{2}\mathbb{P}(\sum_{i=1}^{m-1}\mathbf{1}\{U_m\geq U_{i}^{a}\}=y,\sum_{i=1}^{m-1}\mathbf{1}\{U_m\geq U_{i}^{\frac{1}{a}}\}=x|U_m\geq U_{1}^{a})\\
& + \frac{1}{2}\mathbb{P}(\sum_{i=1}^{m-1}\mathbf{1}\{U_m\geq U_{i}^{a}\}=y,\sum_{i=1}^{m-1}\mathbf{1}\{U_m\geq U_{i}^{\frac{1}{a}}\}=x|U_m\geq U_{1}^{\frac{1}{a}}).
\end{align*}

We then calculate the type II error of this case as, 
\begin{align*}
    \lim_{n\to\infty}\mathbb{P}_{\mathcal{H}_1}^{\frac{1}{n}}(R^c)&=\lim_{n\to\infty}\mathbb{P}_{\mathcal{H}_1}^{\frac{1}{n}}\left( \prod_{t=1}^{n}X_t<(\eta\cdot(m-1))^n \right)\\
    &=\lim_{n\to\infty}\mathbb{P}_{\mathcal{H}_1}^{\frac{1}{n}}\left( \sum_{i=1}^{n}\ln X_t<n\cdot \ln(\eta(m-1)) \right)\\
    &=\inf_t\frac{\mathbb{E}_{\mathcal{H}_1}X_{i}^{t}}{[\eta(m-1)]^t}.
\end{align*}

Similar to Case 1.1, we have $\ln(\eta(m-1))=\mathbb{E}\ln Y+o(\mathbb{E}\ln Y)$, where $Y$ is the random variable that has the identical distribution as $X_t$ under $\mathcal{H}_0$. Since $a(n)=\frac{1}{\Delta(n)}-1$ and $\frac{1}{\Delta(n)}=\Omega(m)$, we have $a(n)\gg m\gg0$. Now we derive the asymptotic type II error under the condition that $a(n)\gg m\gg0$.
\begin{align*}
& \mathbb{E}_{\mathcal{H}_1}X_t^s = \Gamma_1 + \Gamma_2 + \Gamma_3 \\
:= \; & \frac{1}{2}\sum_{x=1}^{m-1}\sum_{y=x}^{m-1}(a+1)\int_{0}^{1}\binom{m-2}{x-1}\binom{m-x-1}{y-x}\cdot(r^a)^x(r^{\frac{1}{a}}-r^a)^{y-x}\cdot(1-r^{\frac{1}{a}})^{m-y-1}dr \cdot(x+y)^s\\
+ \; & \frac{1}{2}\sum_{x=1}^{m-1}\frac{a+1}{a}\int_{0}^{1}\binom{m-2}{x-1}(r^a)^x\cdot(1-r^{\frac{1}{a}})^{m-x-1}dr\cdot(2x)^s\\
+ \; & \frac{1}{2}\sum_{x=0}^{m-1}\sum_{y=x+1}^{m-1}\frac{a+1}{a}\int_{0}^{1}\binom{m-2}{x}\binom{m-x-2}{y-x-1}\cdot(r^a)^x(r^{\frac{1}{a}}-r^a)^{y-x}\cdot(1-r^{\frac{1}{a}})^{m-y-1}dr \cdot(x+y)^s.
\end{align*}
\begin{align*}
\mathbb{E}\ln Y&=\frac{1}{1-\frac{1}{\binom{m+a-1}{a}}}(\sum_{y=1}^{m-1}\int_{0}^{1}\binom{m-1}{y}(r^{\frac{1}{a}}-r^a)^y(1-r^{\frac{1}{a}})^{m-1-y}dr\cdot \ln y\\
    &+\sum_{x=1}^{m-1}\sum_{y=x}^{m-1}\int_{0}^{1}\binom{m-1}{x}\binom{m-1-x}{y-x}(r^a)^x(r^{\frac{1}{a}}-r^a)^{y-x}(1-r^{\frac{1}{a}})^{m-1-y}dr\cdot \ln(x+y))
\end{align*}
Under the condition that $a\gg m\gg0$, we have $r^a\to 0,r\in (0,1)$ and $r^{\frac{1}{a}}\to 1,r\in (0,1)$, $\mathbb{E}\ln Y\to ln(m-1)$, $\Gamma_1\to\frac{1}{2}m^s$, $\Gamma_2\to 0$ and $\Gamma_3 \to \frac{1}{2}(m-1)^s$. Henceforth, we have $\mathbb{E}_{\mathcal{H}_1}X_t^s/(\eta(m-1))^s\to 1$. We also have $R_\mathcal{P}(h)\to 0, e^{-R_\mathcal{P}(h)}\to 1$. This shows that the asymptotic type II error of our proposed test matches the lower bound of the asymptotic type II error of all tests.

\medskip
We next prove (b) under two scenarios.

\medskip
\noindent 
\textbf{Case 2.1}: Minimizing the sum of type I and type II errors, and ${1}/{1-\Delta}$ is an integer. 

We first present a supporting lemma.
\begin{lemma} \label{tv}
For any two distributions $\mathbf{P}_0,\mathbf{P}_1$, we have that,
\begin{align*}
        \inf_{\Psi}\frac{1}{2}[\mathbb{P}_0(\Psi(X)\neq 0)+\mathbb{P}_1(\Psi(X)\neq 1)]=\frac{1}{2}(1-TV(\mathbf{P}_0,\mathbf{P}_1)).
\end{align*}
In particular, the likelihood ratio function $\Psi=\mathbf{1}_{\{f_0(X)\leq f_1(X)\}}$ can achieve the infimum.
\end{lemma}
\begin{proof}
For any test $\Psi$, let $A$ denote its acceptance region, i.e., if $X\in A$ then $\Psi(X)=0$. We then have that, 
\begin{align*}
        \frac{1}{2}[\mathbb{P}_0(\Psi(X)\neq 0)+\mathbb{P}_1(\Psi(X)\neq 1)]&=\frac{1}{2}[\mathbb{P}_0(X\notin A)+\mathbb{P}_1(X\in A)]\\
        &=\frac{1}{2}[1-(\mathbb{P}_0(X\in A)-\mathbb{P}_1(X\in A))].
\end{align*}
As a result, we have
\begin{align*}
        \inf_{\Psi}\mathbb{P}(\Psi(X)\geq T)&=\frac{1}{2}[1-\sup_A(\mathbb{P}_0(X\in A)-\mathbb{P}_1(X\in A))]\\
        &=\frac{1}{2}(1-TV(\mathbf{P}_0,\mathbf{P}_1)).
\end{align*}
We can verify that the infimum can be achieved when choosing $\Psi=\mathbf{1}_{f_0(X)\leq f_1(X)}$. This completes the proof of Lemma \ref{tv}. 
\end{proof}

We next proceed with the proof. Our focus is to derive
\begin{align*}
\inf_{\psi}\lim_{n\to\infty}\sup_{\mathbf{P}_{1:n}}(e_1+e_2)^{\frac{1}{n}}.
\end{align*}

We first note that
\begin{align*}
\inf_{\psi}\lim_{n\to\infty}\sup_{\mathbf{P}_{1:n}}(e_1+e_2)^{\frac{1}{n}}\geq\lim_{n\to\infty}\inf_{\psi}\sup_{\mathbf{P}_{1:n}}(e_1+e_2)^{\frac{1}{n}}.
\end{align*}
We can then turn to the following problem,
\begin{align}
    A:=\lim_{n\to \infty}[\inf_{\psi}\sup_{\mathbf{P}_{1:n}}(\text{type I error}+\text{type II error})]^{\frac{1}{n}}
\end{align}
Here we also add some restrictions on the distributions $\mathbf{P}_{1:n}$. Recall that each distribution $\mathbf{P}_i$ is a multinomial distribution. To make the detection method more powerful, we impose that each term $p_{ij}$ of $\mathbf{P}_i$ cannot be larger than $1-\Delta$.

There are $\binom{m}{\frac{1}{1-\Delta}}$ permutations of distribution $(\underbrace{1-\Delta,1-\Delta,\cdots,1-\Delta}_{\frac{1}{1-\Delta}times},0,0,\cdots,0)$. Denote these multinomial distributions as $\mathbf{R}_i, i=1,2,\cdots,\binom{m}{\frac{1}{1-\Delta}}$. Let each $\mathbf{P}_j^0$ be a combination of $\mathbf{R}_i, i=1,2,\ldots,\binom{m}{\frac{1}{1-\Delta}}$, i.e., each $\mathbf{P}_j^0$ has the probability $1/\binom{m}{\frac{1}{1-\Delta}}$ to be the distribution $\mathbf{R}_i, i=1,2,\cdots,\binom{m}{\frac{1}{1-\Delta}}$. Then we have, 
\begin{align}
    A\geq B:=\lim_{n\to \infty}[\inf_{\psi}(\mathbb{P}_{\mathcal{H}_0,\mathbf{P}_{1:n}^{0}}(\psi(\omega_{1:n},\zeta_{1:n})=1)+\mathbb{P}_{\mathcal{H}_1,\mathbf{P}_{1:n}^{0}}(\psi(\omega_{1:n},\zeta_{1:n})=0)]^{\frac{1}{n}}
\end{align}
Here, both of the two types of error are calculated under the distribution $\mathbf{P}_{1:n}^{0}$, each $\mathbf{P}_j^0$ has the equal probability $1/\binom{m}{\frac{1}{1-\Delta}}$ to be the distribution $\mathbf{R}_i, i=1,2,\cdots,\binom{m}{\frac{1}{1-\Delta}}$. We then use the distribution to calculate the two types of error.

Similar to Case 1.1, the likelihood of $(\omega_{1:n},\zeta_{1:n})$ under $\mathcal{H}_0$ is $(\frac{1}{m})^n$, and the likelihood of $(\omega_{1:n},\zeta_{1:n})$ under $\mathcal{H}_1$ is 
\[
\prod_{t=1}^{n}\left(1/\binom{m}{\frac{1}{1-\Delta}}\cdot \sum_{S\subset \{1,2,\cdots,m\},|S|=\frac{1}{1-\Delta}}\mathbf{1}\{\text{$U_{t,\omega_t}$ is the largest one among $U_{t,i|i\in S}$}\}\right).
\]
By Lemma \ref{tv}, the best rejection region can be represented as
\begin{align}
\label{rej}
R&=\Bigg\{ (\omega_{1:n},\zeta_{1:n}) : \left( (1-\Delta)\cdot\binom{m-1}{\frac{1}{1-\Delta}-1} \right)^n \nonumber \\ 
&\leq\prod_{t=1}^{n}\Bigg(\sum_{S\subset \{1,2,\cdots,m\},|S|=\frac{1}{1-\Delta}}\mathbf{1}\{\text{$U_{t,\omega_t}$ is the largest one among $U_{t,i|i\in S}$}\}\Bigg) \Bigg\}.
\end{align}

Denote the right-hand-side of \ref{rej} as $\prod_{t=1}^{n}X_t$. We have $X_1,X_2,\cdots,X_n$ are i.i.d. We next derive the distribution of $X_t$ under $\mathcal{H}_0$ and $\mathcal{H}_1$. Following Case 1.1, the distribution of $X_t$ under $\mathcal{H}_0$ and $\mathcal{H}_1$ can be represented as, 
\begin{align*}
& \mathbb{P}_{\mathcal{H}_0}(X_t=0) =\frac{\frac{1}{1-\Delta}-1}{m};\\
& \mathbb{P}_{\mathcal{H}_0}\left( X_t = \binom{k-1}{\frac{1}{1-\Delta}-1} \right) = \frac{1}{m}, \; k=\frac{1}{1-\Delta},\ldots,m.\\
& \mathbb{P}_{\mathcal{H}_1}\left( X_t = \binom{k-1}{\frac{1}{1-\Delta}-1} \right) = \frac{\binom{k-1}{\frac{1}{1-\Delta}-1}}{\sum_{i=\frac{1}{1-\Delta}}^{m}\binom{i-1}{\frac{1}{1-\Delta}-1}}, \; k=\frac{1}{1-\Delta},\ldots,m.
\end{align*}

Consequently, we can represent the two types of error as,
\begin{align*}
\text{type I error}&=\mathbb{P}_{\mathcal{H}_0}\left( \prod_{t=1}^{n}X_t \geq \left\{ (1-\Delta)\cdot\binom{m-1}{\frac{1}{1-\Delta}-1} \right\}^n \right);\\
\text{type II error}&=\mathbb{P}_{\mathcal{H}_1}\left( \prod_{t=1}^{n}X_t < \left\{ (1-\Delta)\cdot\binom{m-1}{\frac{1}{1-\Delta}-1} \right\}^n \right).
\end{align*}

By Lemma \ref{key}, we have that, 
\begin{align*}
B \geq \lim_{n\to \infty}(\text{type I error})^{\frac{1}{n}} 
= \inf_s \exp\{-s\ln((1-\Delta)\cdot\binom{m-1}{\frac{1}{1-\Delta}-1})\}\cdot\mathbb{E}\exp\{s\ln{X_t}\}.
\end{align*}
Furthermore, we have that,
\begin{align*}
\frac{\mathbb{E}(X_t^s)}{\left((1-\Delta)\cdot\binom{m-1}{\frac{1}{1-\Delta}-1}\right)^s}&=\frac{\sum_{k=\frac{1}{1-\Delta}}^{m}\frac{1}{m}\cdot(\binom{k-1}{\frac{1}{1-\Delta}-1})^s}{\left((1-\Delta)\cdot\binom{m-1}{\frac{1}{1-\Delta}-1}\right)^s}\\
    &=\frac{\sum_{k=\frac{1}{1-\Delta}}^{m}\frac{1}{m}\cdot[\frac{(k-1)(k-2)\cdots(k-\frac{1}{1-\Delta}+1)}{(\frac{1}{1-\Delta}-1)!}]^s}{((1-\Delta)\cdot\binom{m-1}{\frac{1}{1-\Delta}-1})^s}\\
    &=\frac{\sum_{k=\frac{1}{1-\Delta}}^{m}\frac{1}{m}\cdot m^{s\cdot(\frac{1}{1-\Delta}-1)}\cdot\frac{1}{(\frac{1}{1-\Delta}-1)!^s}\cdot(\frac{k-1}{m}\cdot\frac{k-2}{m}\cdots\frac{k-\frac{1}{1-\Delta}+1}{m})^t}{((1-\Delta)\cdot\binom{m-1}{\frac{1}{1-\Delta}-1})^s}\\
    &\geq\frac{\frac{1}{m}\sum_{k=\frac{1}{1-\Delta}}^{m}(\frac{k-1}{m}\cdot\frac{k-2}{m}\cdots\frac{k-\frac{1}{1-\Delta}+1}{m})^s}{(1-\Delta)^s}\\
    &\geq \frac{\frac{1}{m}\sum_{k=1}^{m-\frac{1}{1-\Delta}+1}(\frac{k}{m})^s}{(1-\Delta)^s}\\
    &=\frac{\frac{1}{m}\sum_{k=1}^{m}(\frac{k}{m})^t}{(1-\Delta)^s}-\frac{\frac{1}{m}\sum_{k=m-\frac{1}{1-\Delta}+2}^{m}(\frac{k}{m})^s}{(1-\Delta)^s}\\
    &\geq \frac{\frac{1}{m}\sum_{k=1}^{m}(\frac{k}{m})^s}{(1-\Delta)^s}-\frac{\frac{1}{m}(\frac{1}{1-\Delta}-1)}{(1-\Delta)^s}\\
    &\geq\frac{\int_{0}^{1}(r^{\frac{1}{1-\Delta}-1})^s dr-\frac{1}{m}(\frac{1}{1-\Delta}-1)}{(1-\Delta)^s}\\
    &=\frac{\int_{0}^{1}(r^{\frac{1}{1-\Delta}-1})^s dr}{(1-\Delta)^s}-O(\frac{1}{m}),
\end{align*}
which is exactly the result of our testing method. By Theorem \ref{optimscore-complete}, the asymptotic sum of type I and II errors of our proposed test is $\inf_{s\in(0,1)}\frac{\int_{0}^{1}(r^{\frac{1}{1-\Delta}-1})^s dr}{(1-\Delta)^s}$. This shows that our test attains the minimal asymptotic sum of the two types of errors over all tests.

\medskip
\noindent 
\textbf{Case 2.2}: Minimizing the sum of type I and type II errors, and $\Delta$ is a function of $n$. 

Our focus is to derive 
\begin{align*}
\inf_{\psi}\lim_{n\to\infty}\sup_{\mathbf{P}_{1:n}}(e_1+e_2)^{\frac{1}{n}}.
\end{align*}
We first note that
\begin{align*}
\inf_{\psi}\lim_{n\to\infty}\sup_{\mathbf{P}_{1:n}}(e_1+e_2)^{\frac{1}{n}}\geq\lim_{n\to\infty}\inf_{\psi}\sup_{\mathbf{P}_{1:n}}(e_1+e_2)^{\frac{1}{n}}.
\end{align*}
We can then turn to the following problem:
\begin{align}
    A:=\lim_{n\to \infty}[\inf_{\psi}\sup_{\mathbf{P}_{1:n}}(\text{type I error}+\text{type II error})]^{\frac{1}{n}}
\end{align}

We first calculate the asymptotic sum of two types of errors of our test. 
\begin{align*}
    \limsup_{n\to\infty}\mathbb{P}_{\phi_0,\mathcal{H}_1}(R^c)&=\exp\{-S_{\mathcal{P}}(h)\}\\
    S_{\mathcal{P}}(h)&=-\inf_{h}\sup_{\mathbf{P}\in\mathcal{P}}\mathbb{E}_0h(Y)-\log \mathbb{E}_{1,\mathbf{P}}e^{-h(Y)}\\
    &=-\inf_{h}\inf_{\theta_1,\theta_2>0}\frac{\theta_2}{\theta_1+\theta_2}\log\mathbb{E}_0\exp{\{ \theta_1h(Y) \}} \nonumber-\frac{\theta_1}{\theta_1+\theta_2}\log\sup_{\mathbf{P}\in\mathcal{P}}\mathbb{E}_{1,\mathbf{P}}\exp{\{ -\theta_2h(Y) \}}\\
    &=-\inf_{k\in(0,1)}\log\int_{0}^{1}p_{0}^{k}(y)p_{1}^{1-k}(y)dy\\
    &=-\inf_{k\in(0,1)}\log\int_{0}^{1}(y^{\frac{\Delta}{1-\Delta}}+y^{\frac{1-\Delta}{\Delta}})^kdy
\end{align*}

By Lemma \ref{tv}, the best rejection region can be represented as,
\begin{align}
\label{best_rejection_region}
R=\left\{ (\omega_{1:n},\zeta_{1:n}):((m-1))^n\leq\prod_{t=1}^{n}\left(\sum_{j=1}^{m-1} \left( \mathbf{1}\{U_{i,m}\geq U_{i,j}^{a}\}+\mathbf{1}\{U_{i,m}\geq U_{i,j}^{\frac{1}{a}}\} \right) \right) \right\}.
\end{align}

Denote the right-hand-side of \ref{best_rejection_region} as $\prod_{t=1}^{n}X_t$. We see that $X_1,X_2,\cdots,X_n$ are i.i.d. Similar as in Case 1.2, we first obtain that the distribution of $X_t$ under $H_0$ is, 
\begin{align*}
&\mathbb{P}_{\mathcal{H}_0}\left( \sum_{i=1}^{m-1}\mathbf{1}\{U_m\geq U_{i}^{a}\}=y, \; \sum_{i=1}^{m-1}\mathbf{1}\{U_m\geq U_{i}^{\frac{1}{a}}\}=x \right)\\
= \; &\int_{0}^{1}\binom{m-1}{x}\binom{m-1-x}{y-x}(r^a)^x\cdot(r^{\frac{1}{a}}-r^a)^{y-x}\cdot(1-r^{\frac{1}{a}})^{m-1-y}dr, \; y\geq x.
\end{align*}
The distribution of $X_t$ under $\mathcal{H}_1$ is,
\begin{align*}
&\mathbb{P}_{\mathcal{H}_1}\left( \sum_{i=1}^{m-1}\mathbf{1}\{U_m\geq U_{i}^{a}\}=y,\sum_{i=1}^{m-1}\mathbf{1}\{U_m\geq U_{i}^{\frac{1}{a}}\}=x \right)\\
= \; & \frac{1}{2}\mathbb{P}\left( \sum_{i=1}^{m-1}\mathbf{1}\{U_m\geq U_{i}^{a}\}=y,\sum_{i=1}^{m-1}\mathbf{1}\{U_m\geq U_{i}^{\frac{1}{a}}\}=x|U_m\geq U_{1}^{a} \right)\\
& + \frac{1}{2}\mathbb{P}\left( \sum_{i=1}^{m-1}\mathbf{1}\{U_m\geq U_{i}^{a}\}=y,\sum_{i=1}^{m-1}\mathbf{1}\{U_m\geq U_{i}^{\frac{1}{a}}\}=x|U_m\geq U_{1}^{\frac{1}{a}} \right),
\end{align*}
where the first probability equals 
\[
\frac{a+1}{a}\int_{0}^{1}\binom{m-2}{x-1}(r^a)^x\cdot(1-r^{\frac{1}{a}})^{m-y-1}dr, \; \text{ when } y=x\geq 1, 
\]
and equals 
\[
\frac{a+1}{a}\int_{0}^{1}\binom{m-2}{x}\cdot\binom{m-x-2}{y-x-1}(r^a)^x\cdot(r^{\frac{1}{a}}-r^a)^{y-x}\cdot(1-r^{\frac{1}{a}})^{m-y-1}dr, \; \text{ when } y>x,
\]
and the second probability equals 
\[
(a+1)\int_{0}^{1}\binom{m-2}{x-1}\cdot\binom{m-x-1}{y-x}(r^a)^x\cdot(r^{\frac{1}{a}}-r^a)^{y-x}\cdot(1-r^{\frac{1}{a}})^{m-y-1}dr, \; \text{ when } y\geq x\geq 1.
\]

Next, we derive the lower bound of the optimal asymptotic sum of two types of errors,
\begin{align*}
    A&\geq \lim_{n\to\infty}\inf_{\psi}(\text{type II error under $\mathbf{P}_{1:n}^{*}$})^{\frac{1}{n}}
\end{align*}

We obtain the type II error as
\begin{align*}
& \lim_{n\to\infty}\mathbb{P}_{\mathcal{H}_1}^{\frac{1}{n}}(R^c) =\lim_{n\to\infty}\mathbb{P}_{\mathcal{H}_1}^{\frac{1}{n}}\left( \prod_{t=1}^{n}X_t<(m-1)^n \right) \\
= \; & \lim_{n\to\infty}\mathbb{P}_{\mathcal{H}_1}^{\frac{1}{n}}\left( \sum_{i=1}^{n}\ln X_t<n\cdot \ln(m-1) \right) 
= \inf_t\frac{\mathbb{E}_{\mathcal{H}_1}X_{i}^{t}}{(m-1)^t}.
\end{align*}
Under the condition that $a(n)\gg m\gg0$, we have, 
\begin{align*}
& \mathbb{E}_{\mathcal{H}_1}X_t^s = \Gamma_1 + \Gamma_2 + \Gamma_3 \\
:= \; & \frac{1}{2}\sum_{x=1}^{m-1}\sum_{y=x}^{m-1}(a+1)\int_{0}^{1}\binom{m-2}{x-1}\binom{m-x-1}{y-x}\cdot(r^a)^x(r^{\frac{1}{a}}-r^a)^{y-x}\cdot(1-r^{\frac{1}{a}})^{m-y-1}dr \cdot(x+y)^s \\
+ \; & \frac{1}{2}\sum_{x=1}^{m-1}\frac{a+1}{a}\int_{0}^{1}\binom{m-2}{x-1}(r^a)^x\cdot(1-r^{\frac{1}{a}})^{m-x-1}dr\cdot(2x)^s\\
+ \; & \frac{1}{2}\sum_{x=0}^{m-1}\sum_{y=x+1}^{m-1}\frac{a+1}{a}\int_{0}^{1}\binom{m-2}{x}\binom{m-x-2}{y-x-1}\cdot(r^a)^x(r^{\frac{1}{a}}-r^a)^{y-x}\cdot(1-r^{\frac{1}{a}})^{m-y-1}dr \cdot(x+y)^s.
\end{align*}
Then, we have $r^a\to 0,r\in (0,1)$, $r^{\frac{1}{a}}\to 1,r\in (0,1)$, $\Gamma_1 \to \frac{1}{2}m^s$, $\Gamma \to 0$, and $\Gamma_3 \to \frac{1}{2}(m-1)^s$. 

Henceforth, we have $\mathbb{E}_{\mathcal{H}_1}X_t^s/(m-1)^s\to 1$. We also have $R_\mathcal{P}(h)\to 0, e^{-R_\mathcal{P}(h)}\to 1$. This result shows that the asymptotic sum of the two types of errors of our proposed test matches the lower bound of the asymptotic sum of the two types of errors of all tests.

This completes the proof of Theorem~\ref{optimality-complete}. 
\eop

\subsection{Proof of Theorem~\ref{thm:partial:gumbel}}

By the definition that $Y=U_{\omega_t}$, we have.
\begin{align*}
        \mathbb{P}(Y\leq r)&=\sum_{i=1}^{N}\sum_{j=1}^{N}\mathbb{P}\left( U_{i}^{\frac{1}{p_i}}\text{is the largest one},U_j\leq r \right) \cdot q_{ij}\\
        &=\sum_{i=1}^{N}\sum_{j\neq i}\frac{p_i}{1-p_j}(r-r^{\frac{1}{p_j}}\cdot p_j)\cdot q_{ij}+\sum_{i=1}^{N}p_ir^{\frac{1}{p_i}}\cdot q_{ii} := f(r).
\end{align*}
Furthermore, 
\begin{align*}
        f'(r)&=\sum_{i=1}^{N}\sum_{j\neq i}\frac{p_i}{1-p_j}p_j\cdot(-1)\cdot q_{ij}(\frac{1}{p_j}-1)r^{\frac{1}{p_j}-2}+\sum_{i=1}^{N}p_i(\frac{1}{p_i}-1)r^{\frac{1}{p_i}-2}q_{ii}\\
        &=\sum_{i=1}^{N}\sum_{j\neq i}(-1)p_iq_{ij}r^{\frac{1}{p_j}-2}+\sum_{i=1}^{N}(1-p_i)r^{\frac{1}{p_i}-2}q_{ii}\\
        &=\sum_{i=1}^{N}\sum_{j\neq i}(-1)p_jq_{ji}r^{\frac{1}{p_i}-2}+\sum_{i=1}^{N}(1-p_i)r^{\frac{1}{p_i}-2}q_{ii}.
\end{align*}
By the condition that $q_{ii}\geq \frac{\sum_{j\neq i}p_jq_{ji}}{1-p_i}$, the coefficients of $r^{\frac{1}{p_i}-2}$ are all positive. This implies that $f'(r) \geq 0$. Then, as $f(1)=1$, we have
\begin{align*}
        \mathbb{P}(Y\leq r)=rf(r)\leq r.
\end{align*}
This completes the proof of Theorem~\ref{thm:partial:gumbel}.
\eop

\subsection{Proof of Theorem~\ref{optimscore-partial}}

We first present a number of supporting lemmas.

\begin{lemma} \label{compare}
For any constant $r\in[0,1]$, $f(x) := {(r-r^{\frac{1}{x}}\cdot x)}/{(1-x)}$ is a monotonically increasing function of $x\in[0,1]$.
\end{lemma} \label{lemma b.2}
\begin{proof}
We have
\begin{align*}
        f'(x)=\left(r-r^{\frac{1}{x}}+\frac{1}{x}r^{\frac{1}{x}}\ln{r}-r^{\frac{1}{x}}\ln{r}\right)/(1-x)^2.
\end{align*}
Using the inequality $\ln{r^{1-\frac{1}{x}}}\leq r^{1-\frac{1}{x}}-1$, we have
\begin{align*}
\ln{r} \geq \frac{r^{1-\frac{1}{x}}-1}{1-\frac{1}{x}}, \text{ and } 
f'(x) =r-r^{\frac{1}{x}}+(\frac{1}{x}-1)r^{\frac{1}{x}}\ln{r} 
\geq r-r^{\frac{1}{x}}+(\frac{1}{x}-1)r^{\frac{1}{x}}\cdot\frac{r^{1-\frac{1}{x}}-1}{1-\frac{1}{x}} =0.
\end{align*}
This completes the proof of Lemma \ref{compare}.    
\end{proof}

\begin{lemma} \label{lemma b.3}
For $\mathbf{P}$, suppose $p_1\geq p_2\geq\cdots\geq p_N$, $F_{1,\mathbf{P},\mathbf{Q}}(r)=\sum_{i=1}^{N}\sum_{j\neq i}\frac{p_i}{1-p_j}(r-r^{\frac{1}{p_j}}\cdot p_j)\cdot q_{ij}+\sum_{i=1}^{N}p_ir^{\frac{1}{p_i}}\cdot q_{ii}$. Then we have $\sup_{\mathbf{Q}\in\mathcal{Q}}F_{1,\mathbf{P},\mathbf{Q}}(r)=F_{1,\mathbf{P},\mathbf{Q}^{*}}(r)$, where
\begin{gather*}
\mathbf{Q}^{*}=\begin{pmatrix}
\theta & 1-\theta & 0 & \cdots & 0 \\
1-\theta & \theta & 0 & \cdots & 0 \\
1-\theta & 0 & \theta & \cdots & 0 \\
\vdots & \vdots & \vdots & \ddots & \vdots \\
1-\theta & 0 & 0 & \cdots & \theta
\end{pmatrix}.
\end{gather*}
\end{lemma}
\begin{proof}
By the inequality that 
\[
\frac{r-r^{\frac{1}{p_j}}\cdot p_j}{1-p_j}\geq r \geq r^{\frac{1}{p_i}}, 
\]
as well as the monotonicity, we obtain the desired result, which completes the proof of Lemma \ref{lemma b.3}. 
\end{proof}

\begin{lemma}[\cite{li2024statisticalframeworkwatermarkslarge}] \label{extreme}
The set of extreme points of $\mathcal{P}_\Delta$, denoted by $Ext(\mathcal{P}_\Delta)$, is given by 
\begin{equation*}
    Ext(\mathcal{P}_\Delta) = \big\{\pi(\mathbf{P}_{\Delta}^{\star}): \pi~\text{is a permutation on $\{1, 2, \ldots, |\mathcal{W}|\}$} \big\},
\end{equation*}
where $\pi(\mathbf{P})$ denotes the permuted NTP distribution whose $i$th coordinate is $P_{\pi(i)}$.
\end{lemma}

\begin{lemma}[Donsker-Varadhan representation]
\begin{align*}
D_{KL}(\mathbf{P}||\mathbf{Q})=\sup_{T:\mathcal{X}\to \mathbb{R}}\mathbb{E}_{\mathbf{P}}[T(x)]-\log(\mathbb{E}_{\mathbf{Q}}[e^{T(x)}]).
\end{align*}
\end{lemma}

\begin{lemma} \label{lemma b.5}
The Hessian matrix of $f(\mathbf{P},r)=f(p_2,p_3,\cdots,p_N)$ is positive semi-definite.
\end{lemma}
\begin{proof}
For $i\geq 3$,
\begin{align*}
\frac{\partial^2f(\mathbf{P},r)}{(\partial p_i)^2}=r^{\frac{1}{p_i}}\frac{\ln^2(r)}{p_i^3}\geq 0.
\end{align*}
For $i=2$, we have
\begin{align*}
\frac{\partial^2f(\mathbf{P},r)}{(\partial p_2)^2}&= (1-\Delta)\cdot(1-\theta)\bigg\{ \left(r^{\frac{1}{p_2}}\ln^2r\frac{-1}{p_2^3}+r^{\frac{1}{p_2}}\ln^2r\frac{1}{p_2^2}\right)\cdot(1-p_2)^2\\
&+\left(\frac{1}{p_2}r^{\frac{1}{p_2}}\ln{r}-r^{\frac{1}{p_2}}\ln{r}+r-r^{\frac{1}{p_2}}\right)\cdot2(1-p_2) \bigg\} /(1-p_2)^4 +\theta r^{\frac{1}{p_2}}\ln^2r\frac{1}{p_2^3}.
\end{align*}
To show it is non-negative, it suffices to show that 
\begin{align*}
2(r^{1-\frac{1}{x}}-1)\geq (\ln{r^{1-\frac{1}{x}}})^2+2\ln{r^{1-\frac{1}{x}}}, \; \text{ for } r^{1-\frac{1}{x}} > 1,
\end{align*}
which can be verified directly. This completes the proof of Lemma \ref{lemma b.5}. 
\end{proof}

We now proceed with the proof of Theorem~\ref{optimscore-partial}. 

First, we simplify $\mathbb{E}_{1,\mathbf{P},\mathbf{Q}}e^{-h(Y)}$ as follows. 
\begin{align*}
& \mathbb{E}_{1,\mathbf{P},\mathbf{Q}}e^{-h(Y)} = \int_{0}^{1}e^{-h(r)}\cdot F_{1,\mathbf{P},\mathbf{Q}}(dr)\\
= \; & F_{1,\mathbf{P},\mathbf{Q}}(r)e^{-h(r)}|_{0}^{1}+\int_{0}^{1}F_{1,\mathbf{P},\mathbf{Q}}(r)e^{-h(r)}h(dr)
= e^{-h(1)}+\int_{0}^{1}F_{1,\mathbf{P},\mathbf{Q}}(r)e^{-h(r)}h(dr).
\end{align*}
This representation shows that, for a non-decreasing function $h$, 
\begin{align*}
\argmax_{\mathbf{P},\mathbf{Q}}\mathbb{E}_{1,\mathbf{P},\mathbf{Q}}e^{-h(Y)}=\argmax_{\mathbf{P},\mathbf{Q}}F_{1,\mathbf{P},\mathbf{Q}}(r).
\end{align*}
Recall that
\begin{align*}
    F_{1,\mathbf{P},\mathbf{Q}}(r)=\sum_{i=1}^{N}\sum_{j\neq i}\frac{p_i}{1-p_j}\left( r-r^{\frac{1}{p_j}}\cdot p_j \right)\cdot q_{ij}+\sum_{i=1}^{N}p_ir^{\frac{1}{p_i}}\cdot q_{ii}.
\end{align*}

Without loss of generality, suppose $p_1\geq p_2\geq\cdots\geq p_N$. By Lemma \ref{lemma b.3}, 
\begin{align*}
F_{1,\mathbf{P},\mathbf{Q}}(r)&=\frac{p_1}{1-p_2}(r-r^{\frac{1}{p_2}}p_2)(1-\theta)+\sum_{i=2}^{N}\frac{p_i}{1-p_1}(r-r^{\frac{1}{p_1}}p_1)(1-\theta)+\sum_{i=1}^{N}p_ir^{\frac{1}{p_i}}\theta\\
    &=(1-\theta)\frac{p_1}{1-p_2}(r-r^{\frac{1}{p_2}}p_2)+(1-\theta)(r-r^{\frac{1}{p_1}}p_1)+\theta\sum_{i=1}^{N}p_ir^{\frac{1}{p_i}}\\
    &=(1-\theta)\frac{p_1}{1-p_2}(r-r^{\frac{1}{p_2}}p_2)+(1-\theta)r+(2\theta-1)r^{\frac{1}{p_1}}p_1+\theta\sum_{i=2}^{N}p_ir^{\frac{1}{p_i}}\\
    &: = f(\mathbf{P},r).
\end{align*}
We note that for $i\geq 3$, 
\[
\frac{\partial^2f(\mathbf{P},r)}{(\partial p_i)^2}=\theta r^{\frac{1}{p_i}}\frac{\ln^2(r)}{p_i^3}>0.
\]

Combining Lemma \ref{extreme} and the convexity of $p_3,p_4,\cdots,p_N$, we have
\begin{align*}
\argmax_{p_3,\cdots,p_N}f(\mathbf{P},r)=(p_2,p_2,\cdots,p_2,\epsilon,0,0,\cdots,0),
\end{align*}
where $\epsilon<p_2$. Assuming there are $m$ times of $p_2$, then $p_1=1-mp_2-\epsilon$. Moreover, 
\begin{align*}
    f(\mathbf{P},r) = \; & (1-\theta)\frac{p_1}{1-p_2}(r-r^{\frac{1}{p_2}}p_2)+(1-\theta)r+(2\theta-1)r^{\frac{1}{p_1}}p_1+\theta\sum_{i=2}^{N}p_ir^{\frac{1}{p_i}}\\
    = \; & (1-\theta)\frac{1-mp_2-\epsilon}{1-p_2}r+p_2r^{\frac{1}{p_2}}\frac{m(1-2\theta)p_2+\epsilon+\theta-1-\theta\epsilon+\theta m}{1-p_2}+(1-\theta)r\\
    &+\theta\epsilon r^{\frac{1}{\epsilon}}+(2\theta-1)r^{\frac{1}{1-mp_2-\epsilon}}(1-mp_2-\epsilon).
\end{align*}
Fixing $\epsilon,m$, we have that, 
\begin{align*}
\frac{df(\mathbf{P},r)}{dp_2} = \; & (1-\Theta)r\cdot\frac{-m+1-\epsilon}{(1-p_2)^2}+r^{\frac{1}{p_2}}\ln{r}\frac{-1}{p_2}\frac{m(1-2\Theta)p_2+\epsilon+\theta-1-\theta\epsilon+\theta m}{1-p_2}\\
&+r^{\frac{1}{p_2}}\frac{m(1-2\theta)p_2+\epsilon+\theta-1-\theta\epsilon+\theta m}{1-p_2} 
+r^{\frac{1}{p_2}}p_2\frac{m-m\theta+\epsilon+\theta-1-\theta\epsilon}{(1-p_2)^2}+A(r),
\end{align*}
where $A(r)$ is a component smaller than 0. By Lemma \ref{tv}, we can show the above term is smaller than zero. As such, the smaller $p_2$ is, the larger $f(\mathbf{P},r)$ is, whereas the larger $p_1$ is, the larger $f(\mathbf{P},r)$ is. By fixing $p_1$ at $1-\Delta$, we can then calculate the Hessian matrix of $f(\mathbf{P},r)=f(p_2,p_3,\ldots,p_N)$.

Combining the results above and Lemma \ref{lemma b.5}, we obtain the following, for any non-decreasing function $h$, 
\begin{align*}
\mathbf{Q}^{*},\mathbf{P}^{*}=\argmax_{\mathbf{Q},\mathbf{P}\in\mathcal{P}_\Delta} \mathbb{E}_{1,\mathbf{P},\mathbf{Q}}e^{-h(Y)},
\end{align*}
where
\begin{align*}
\mathbf{Q}^{*} & = \begin{pmatrix}
\theta & 1-\theta & 0 & \cdots & 0 \\
1-\theta & \theta & 0 & \cdots & 0 \\
1-\theta & 0 & \theta & \cdots & 0 \\
\vdots & \vdots & \vdots & \ddots & \vdots \\
1-\theta & 0 & 0 & \cdots & \theta
\end{pmatrix};\\
\mathbf{P}^{*} & = \left( 1-\Delta,\cdots,1-\Delta,1-(1-\Delta)\cdot\left\lfloor\frac{1}{1-\Delta}\right\rfloor,0,\ldots \right).
\end{align*}

Replacing $\theta h(\cdot)$ with $h(\cdot)$, we have that,
\begin{align*}
        &\min_{h}\max_{\mathbf{Q}\in\mathcal{Q}}\max_{\mathbf{P} \in \mathcal{P}}
    \left(\mathbb{E}_0 \theta h(Y) +  \log \mathbb{E}_{1, \mathbf{P},\mathbf{Q}} e^{-\theta h(Y)} \right)\\
    &=\min_{h}\max_{\mathbf{Q}\in\mathcal{Q}}\max_{\mathbf{P} \in \mathcal{P}}
    \left(\mathbb{E}_0  h(Y) +  \log \mathbb{E}_{1, \mathbf{P},\mathbf{Q}} e^{- h(Y)} \right)\\
    &\geq \min_{h}\left(\mathbb{E}_0  h(Y) +  \log \mathbb{E}_{1, \mathbf{P}^{*},\mathbf{Q}^{*}} e^{- h(Y)} \right)\\
    &=-D_{KL}(\mu_0,\mu_{1,\mathbf{P}^{*},\mathbf{Q}^{*}}).
\end{align*}
Thanks to the uniqueness of the Donsker-Varadhan representation, $\mathbb{E}_0  h(Y) +  \log \mathbb{E}_{1, \mathbf{P},\mathbf{Q}} e^{- h(Y)}$ is strictly larger than $-D_{KL}(\mu_0,\mu_{1,\mathbf{P}^{*},\mathbf{Q}^{*}})$, unless we take the log-likelihood ratio
\begin{align*}
        h_{\Delta}^{*}(r)=\log\frac{d\mu_{1,\mathbf{P}^{*},\mathbf{Q}^{*}}}{d\mu_0}(r)=
        \begin{cases} 
        \log\left(\frac{1-\theta}{\Delta}+\left(\left\lfloor\frac{1}{1-\Delta}\right\rfloor\theta+\frac{1}{\Delta}\theta-\frac{1}{\Delta}\right)r^{\frac{\Delta}{1-\Delta}}+\theta r^{\frac{\Tilde{\Delta}}{1-\Tilde{\Delta}}}\right) & \text{if } \Delta \geq \frac{1}{2} \\
        \log\left(2(1-\theta)+(2\theta-1)(r^{\frac{1-\Delta}{\Delta}}+r^{\frac{\Delta}{1-\Delta}})\right) & \text{if } \Delta<\frac{1}{2}
        \end{cases}
\end{align*}
which is non-decreasing in $r$. In this case, we can use Lemmas \ref{compare} to \ref{lemma b.5} to derive $\mathbf{P}^{*}$ and $\mathbf{Q}^{*}$.

This completes the proof of Theorem~\ref{optimscore-partial}. 
\eop

\subsection{Proof of Theorem~\ref{thm:gumbel:partial:optimal}}

This theorem can be proved in a similar fashion as that of Theorem~\ref{optimality-complete}. Here we provide the proof for the case when $\frac{1}{1-\Delta}$ is an integer.

There are $\binom{m}{\frac{1}{1-\Delta}}$ permutations of distribution $(\underbrace{1-\Delta,1-\Delta,\cdots,1-\Delta}_{\frac{1}{1-\Delta}times},0,0,\cdots,0)$. Denote these multinomial distributions as $\mathbf{R}_i, i=1, \ldots, \binom{m}{\frac{1}{1-\Delta}}$. For each $\mathbf{R}_i$, suppose the positions of $1-\Delta$ are $x_1,x_2,\cdots,x_{\frac{1}{1-\Delta}}$. Then we define its adjoint feature matrix as $T_i$, which satisfies that, for any $k\in [1,n]$, the $(k,k)$th-element is $\theta$, and each element in the column $x_1,x_2,\cdots,x_{\frac{1}{1-\Delta}}$ and the line $x_1,x_2,\ldots,x_{\frac{1}{1-\Delta}}$ but not on the diagonal is $(1-\theta)/(\frac{1}{1-\Delta}-1)$, each element in the column $x_1,x_2,\cdots,x_{\frac{1}{1-\Delta}}$ but not on the line $x_1,x_2,\cdots,x_{\frac{1}{1-\Delta}}$ and not on the diagonal is $(1-\Delta)(1-\theta)$. That is,
\begin{align*}
& \mathbf R_1=(\underbrace{1-\Delta,1-\Delta,\cdots,1-\Delta}_{\frac{1}{1-\Delta}times},0,0,\cdots,0), \quad T_1=\begin{pmatrix}
  A & B \\
  C & D
\end{pmatrix}\\
& A=\begin{pmatrix}
      \theta & \frac{1-\theta}{\frac{1}{1-\Delta}-1} & \cdots & \frac{1-\theta}{\frac{1}{1-\Delta}-1}\\
      \frac{1-\theta}{\frac{1}{1-\Delta}-1} & \theta & \cdots & \frac{1-\theta}{\frac{1}{1-\Delta}-1}\\
      \vdots & \vdots & \ddots & \vdots\\
      \frac{1-\theta}{\frac{1}{1-\Delta}-1} & \frac{1-\theta}{\frac{1}{1-\Delta}-1} & \cdots & \theta
    \end{pmatrix},
\quad B=\mathbf{0}_{\frac{1}{1-\Delta}\times\frac{1}{1-\Delta}},\\
& C=\begin{pmatrix}
      (1-\Delta)(1-\theta) & (1-\Delta)(1-\theta) & \cdots & (1-\Delta)(1-\theta)\\
      (1-\Delta)(1-\theta) & (1-\Delta)(1-\theta) & \cdots & (1-\Delta)(1-\theta)\\
      \vdots & \vdots & \ddots & \vdots\\
      (1-\Delta)(1-\theta) & (1-\Delta)(1-\theta) & \cdots & (1-\Delta)(1-\theta)
\end{pmatrix},
\quad D=\theta\cdot\mathbf{1}_{(m-\frac{1}{1-\Delta})\times(m-\frac{1}{1-\Delta})}.
\end{align*}
In this setting, $\mathbf{P}_t^0$ is the probability distribution that is the mixture of those $\mathbf{Q}_i$ with $R_i$.

The hypotheses are of the form,  
\begin{align*}
\mathcal{H}_0:\omega_t\perp\zeta_t,t=1,2,\cdots.N\quad\mathcal{H}_1:\omega_t=\mathcal{S}(\mathbf{P}_t^0,\zeta_t),t=1,2,\cdots,n.
\end{align*}
Then the best rejection region is,
\begin{align*}
R = \; & \left\{ (\omega_{1:n},\zeta_{1:n}):((1-\Delta)\cdot\binom{m-1}{\frac{1}{1-\Delta}-1})^n\leq \prod_{t=1}^{n}Y_t \right\}\\
    Y_t = \; & \sum_{S\subset \{1,2,\cdots,m\},|S|=\frac{1}{1-\Delta}}\theta\cdot\mathbf{1}\{\text{$U_{\omega_t}$ is the largest one among $U_{i|i\in S}$}\}\\
    & + \frac{1-\theta}{\frac{1}{1-\Delta}-1}\cdot\mathbf{1}\left\{ \text{$U_{\omega_t}$ has rank between 2 and $\frac{1}{1-\Delta}$, inclusively} \right\}.
\end{align*}

Under $\mathcal{H}_0$, we have
\begin{align*}
& \mathbb{P}\left( Y_t=\frac{1-\theta}{\frac{1}{1-\Delta}-1}\cdot\binom{m-1}{\frac{1}{1-\Delta}-1} \right)=\frac{\frac{1}{1-\Delta}-1}{m};\\
& \mathbb{P}\left( Y_t=\frac{1-\theta}{\frac{1}{1-\Delta}-1}\cdot\binom{m-1}{\frac{1}{1-\Delta}-1}+\left(\theta-\frac{1-\theta}{\frac{1}{1-\Delta}-1}\right)\cdot\binom{k-1}{\frac{1}{1-\Delta}-1} \right)=\frac{1}{m},k\geq\frac{1}{1-\Delta}.
\end{align*}

By the large deviation theory, the Type 1 error of the optimal test is,
\begin{align*}
    &\mathbb{P}_{\mathcal{H}_0}\left(\prod_{t=1}^{n}Y_t \geq \left(\Delta \cdot \binom{m-1}{\frac{1}{1-\Delta}-1}\right)^N \right) \\
    &= \inf_t \frac{\mathbb{E}[Y_t^t]}{\left((1-\Delta) \cdot \binom{m-1}{\frac{1}{1-\Delta}-1}\right)^t} \\
    &= \frac{\frac{1}{1-\Delta}-1}{m} \cdot \left(\frac{1-\theta}{\frac{1}{1-\Delta}-1} \cdot \binom{m-1}{\frac{1}{1-\Delta}-1}\right)^t \\
    &+ \sum_{k=\frac{1}{1-\Delta}}^{m} \frac{1}{m} \left[\frac{1-\theta}{\frac{1}{1-\Delta}-1} \cdot \binom{m-1}{\frac{1}{1-\Delta}-1} + \left(\theta - \frac{1-\theta}{\frac{1}{1-\Delta}-1}\right) \cdot \binom{k-1}{\frac{1}{1-\Delta}-1} \right]^t  \\
    &= \frac{\frac{1}{1-\Delta}-1}{m} \cdot \left( \frac{1-\theta}{\frac{1}{1-\Delta}-1} \cdot m \right)^t \\
    &+ \sum_{k=\frac{1}{1-\Delta}}^{m} \frac{1}{m} \cdot \left[ \frac{\frac{1}{1-\Delta} \cdot \theta - 1}{\frac{1}{1-\Delta}-1} \cdot \frac{1}{1-\Delta} \cdot \frac{(k-1) \cdots (k-\frac{1}{1-\Delta}+1)}{(m-1) \cdots (m-\frac{1}{1-\Delta}+1)} + \frac{1-\theta}{\frac{1}{1-\Delta}-1} \cdot m \right]^t \\
    &\geq \inf_{\alpha} \int p^\alpha(y) \, dy - O\left(\frac{1}{m}\right).
\end{align*}

This completes the proof of Theorem~\ref{thm:gumbel:partial:optimal}. 
\eop

\section{Proofs for Red-green-list Watermark in Section \ref{sec:redgreen}}
\label{append-sec:redgreen}

\subsection{Proof of Proposition~\ref{prop2}}
For the feature matrix under $\mathcal{H}_0$, the NTP distribution is $\mathbf{P}_t$ no matter what the secret key $\zeta_t$ is. 

For the feature matrix under $\mathcal{H}_1$, and for the complete inheritance case, when $\zeta_t=A_i$, 
\begin{align*}
        \mathbb{P}(\omega_t=k|\mathbf{P}_t,\zeta_t=A_i)=p_{t,k}\cdot\frac{\mathbf{1}\{k\in A_i\}}{\sum_{i\in A_i}p_{t,i}}
\end{align*}
As such, the feature matrix is 
$\begin{pmatrix}
        \mathbf{M}_{t,1} \\
        \mathbf{M}_{t,2} \\
        \vdots \\
        \mathbf{M}_{t,k}
\end{pmatrix}$, 
where $\mathbf{S}_{t,i}=\left( p_{t,1} \frac{\mathbf{1}\{1 \in A_i\}}{\sum_{i \in A_i} p_{t,i}}, p_{t,2} \frac{\mathbf{1}\{2 \in A_i\}}{\sum_{i \in A_i} p_{t,i}}, \ldots, p_{t,m} \frac{\mathbf{1}\{m \in A_i\}}{\sum_{i \in A_i} p_{t,i}} \right)$. 
    
For the partial inheritance case, by definition, the $t$th feature matrix belongs to the class $\left\{\begin{pmatrix}
        \mathbf{Q}_{t,1} \\
        \mathbf{Q}_{t,2} \\
        \vdots \\
        \mathbf{Q}_{t,k}
\end{pmatrix}: TV(\mathbf{Q}_{t,i},\mathbf{M}_{t,i})\leq 1-\theta\right\}$.

This completes the proof of Proposition~\ref{prop2}.
\eop

\subsection{Proof of Theorem~\ref{thm:rg:distribution:complete}}

Under $\mathcal{H}_0$, $\omega_t$ is independent of $\zeta_t$. So $Y_t=1$ is equivalent to being painted as red. The painting style is random and the probability for each element to be pained as green is $\gamma$.

Under $\mathcal{H}_1$, $\omega_t$ is sampled from the green list, so the probability of $\omega_t\in\zeta_t$ is 1. That is, $\mathbb{P}_{\mathcal{H}_1}(Y_t=1)=1$.

This completes the proof of Theorem~\ref{thm:rg:distribution:complete}. 
\eop

\subsection{Proof of Theorem~\ref{optimscore-red}}

For (a), the result holds by directly applying the law of large numbers.
        
For (b), under $\mathcal{H}_1$, the statistic $Y_t$ is a constant 1. Thus we have $\sum_{t=1}^{n}Y_t=n$. Choosing the threshold $\gamma_n$ at $n$ can minimize the type I error and keep the type II error at 0.

This completes the proof of Theorem \ref{optimscore-red}. 
\eop

\subsection{Proof of Theorem~\ref{optimality-complete:rg}}

For (b), consider the case that for each $t$, the NTP distribution is $P:=(1/m, \ldots, 1/m)$. Denote the asymptotic sum of type I error and type II error under $P$ as $g_P(\psi)$. Then we have
\begin{align*}
        \inf_{\psi}g(\psi)\geq \inf_{\psi}g_P(\psi).
\end{align*}

We next compute the likelihood of $(\omega_t,\zeta_t)_{1:n}$ under $\mathcal{H}_0$ and $\mathcal{H}_1$.
\begin{align*}
\mathbb{P}_{\mathcal{H}_0}(\omega_{1:n},\zeta_{1:n}) & =\left( \frac{1}{\binom{m}{\gamma\cdot m}}\cdot\frac{1}{m} \right)^n, \\
\mathbb{P}_{\mathcal{H}_1}(\omega_{1:n},\zeta_{1:n}) & =\prod_{t=1}^{n}\mathbf{1}(\omega_t\in\zeta_t)\cdot\frac{1}{\gamma\cdot m}\cdot\frac{1}{\binom{m}{\gamma\cdot m}}.
\end{align*}
By Lemma \ref{tv}, the optimal test is to reject the null hypothesis when $\mathbb{P}_{\mathcal{H}_0}(\omega_{1:n},\zeta_{1:n})\leq \mathbb{P}_{\mathcal{H}_1}(\omega_{1:n},\zeta_{1:n})$. So the rejection region is,
\begin{align*}
R=\big\{ (\omega_{1:n},\zeta_{1:n})|\omega_t\in\zeta_t,t=1,2,\cdots,n \big\}.
\end{align*}
Henceforth, we have proved that $\inf_\psi g(\psi)\geq g(\psi_{2}^{'})$. By definition, we have $g(\psi_2^{'})\geq \inf_\psi g(\psi)$. This proves (b).

For (a), similarly, consider the case that for each $t$, the NTP distribution is $P:=(1/m,\ldots,1/m)$. Denote the asymptotic type II error under $P$ as $f_P(\psi)$. Then we have
\begin{align*}
        \inf_{\psi}f(\psi)\geq \inf_{\psi}f_P(\psi).
\end{align*}
By the Neyman-Pearson lemma, the optimal test is the likelihood ratio test. Henceforth, the optimal rejection region is, 
\begin{align*}
R=\big\{ (\omega_{1:n},\zeta_{1:n})|\mathbb{P}_{\mathcal{H}_1}(\omega_{1:n},\zeta_{1:n})\geq \eta\cdot\mathbb{P}_{\mathcal{H}_0}(\omega_{1:n},\zeta_{1:n}) \big\}.
\end{align*}
The rejection rule is equivalent to,
\begin{align*}
\prod_{t=1}^{n}\mathbf{1}\{\omega_t\in\zeta_t)\cdot\frac{1}{\gamma\cdot m}\cdot\frac{1}{\binom{m}{\gamma\cdot m}}\geq \eta (\frac{1}{\binom{m}{\gamma\cdot m}}\cdot\frac{1}{m})^n 
\Leftrightarrow& \prod_{t=1}^{n}\mathbf{1}(\omega_t\in\zeta_t)\geq \eta\gamma^n.
\end{align*}
Therefore, the optimal rejection rule should be $\{(\omega_{1:n},\zeta_{1:n})|\sum_{i=1}^{n}Y_t=n\}$.  We note that choosing any threshold $\gamma_n\leq n$ leads to the same type II error, so$\inf_{\psi}f(\psi)=f(\psi_{1}^{'})$. The proves (a).

Together it completes the proof of Theorem~\ref{optimality-complete:rg}. 
\eop

\subsection{Proof of Theorem~\ref{thm:rg:partial:distribution}}

It is straightforward to derive the distribution of $Y_t$ under $\mathcal{H}_0$. Next, we consider $\mathcal{H}_1$.

Note that $TV(\omega_t|\zeta_t,\mathcal{S}(\mathbf{P}_t,\zeta_t))\leq 1-\theta$, with $\theta\geq 1/2$. So we can use the TV distance to derive $\mathbb{P}_{\mathcal{H}_1}(Y_t=1)$. That is, 
\begin{align*}
& 1-\theta \geq TV(\omega_t|\zeta_t,\mathcal{S}(\mathbf{P}_t,\zeta_t))
=\frac{1}{2}\sum_{i=1}^{N}|p_i\frac{\mathbf{1}\{i\in D\}}{\sum_{i\in D}p_i}-q_i| 
=\frac{1}{2}\sum_{i\in D}|\frac{p_i}{\sum_{i\in D}p_i}-q_i|+\frac{1}{2}\sum_{i\notin D}q_i\\
= \; & \frac{1}{2}\sum_{i\in D}|\frac{p_i}{\sum_{i\in D}p_i}-q_i|+\frac{1}{2}(1-\sum_{i\in D}q_i) 
=\frac{1}{2}+\frac{1}{2}\sum_{i\in D}(|\frac{p_i}{\sum_{i\in D}p_i}-q_i|-q_i)\\
\geq \; & \frac{1}{2}+\frac{1}{2}\sum_{i\in D}(\frac{p_i}{\sum_{i\in D}p_i}-q_i-q_i) 
=1-\sum_{i\in D}q_i.
\end{align*}
Therefore, for each $i$, $q_{ij}$ satisfies that $\sum_{j\in D_i}q_{ij}\geq \theta$. Consequently, 
\begin{align*}
        \mathbb{P}(Y_t=1)=\sum_{i=1}^{\gamma\cdot m}\frac{1}{\binom{m}{\gamma\cdot m}}\sum_{j\in D_i}q_{ij}\geq \theta.
\end{align*}
Accordingly, we have,
\begin{align*}
        \mathbb{P}(Y_t=0)=1-\mathbb{P}(Y_t=1)\leq 1-\theta.
\end{align*}

This completes the proof of Theorem~\ref{thm:rg:partial:distribution}. 
\eop

\subsection{Proof of Theorem~\ref{rg:partial}}

Denote the rejection region as $R=\{(\omega_{1:n},\zeta_{1:n}):\sum_{t=1}^{n}Y_t\geq x\}$. Then type I error $=\mathbb{P}_{\mathcal{H}_0}(R)$, and type II error $=\mathbb{P}_{\mathcal{H}_1}(R^c)$. The distribution of $Y_t$ under $\mathcal{H}_0$ and $\mathcal{H}_1$ is:
\begin{align*}
& \mathbb{P}_{\mathcal{H}_0}(Y_t=1)=\gamma, \quad \mathbb{P}_{\mathcal{H}_0}(Y_t=0)=1-\gamma,\\
& \mathbb{P}_{\mathcal{H}_1}(Y_t=1)=\theta, \quad \mathbb{P}_{\mathcal{H}_0}(Y_t=0)=1-\theta.
\end{align*}
Then the type I and type II errors can be calculated as,
\begin{align*}
\mathbb{P}_{\mathcal{H}_0}\left( \sum_{t=1}^{n}Y_t\geq x \right)&=\sum_{i=x}^{n}\binom{n}{i}\gamma^i\cdot(1-\gamma)^{n-i}\\
\mathbb{P}_{\mathcal{H}_1}\left( \sum_{t=1}^{n}Y_t\leq x-1 \right)&=\sum_{i=0}^{x-1}\binom{n}{i}\theta^i\cdot(1-\theta)^{n-i}.
\end{align*}
So the optimal threshold can be represented by
\begin{align*}
        x=\argmin_{x\in[0,n]}\sum_{i=x}^{n}\binom{n}{i}\gamma^i\cdot(1-\gamma)^{n-i}+\sum_{i=0}^{x-1}\binom{n}{i}\theta^i\cdot(1-\theta)^{n-i}.
\end{align*}
Denote $F(x) := \sum_{i=x}^{n}\binom{n}{i}\gamma^i\cdot(1-\gamma)^{n-i}+\sum_{i=0}^{x-1}\binom{n}{i}\theta^i\cdot(1-\theta)^{n-i}$. Then, 
\begin{align*}
F(x+1)-F(x)=\binom{n}{x}\theta^x(1-\theta)^{n-x}-\binom{n}{x}\gamma^x(1-\gamma)^{n-x}.
\end{align*}
Since
\begin{align*}
        F(x+1)-F(x)\geq 0\Leftrightarrow x \geq \left\lceil n\cdot\frac{\log(1-\gamma)-\log(1-\theta)}{\log\theta+\log(1-\gamma)-\log\gamma-\log(1-\theta)} \right\rceil,
\end{align*}
the optimal threshold is
\begin{align*}
        x=\left\lceil n\cdot\frac{\log(1-\gamma)-\log(1-\theta)}{\log\theta+\log(1-\gamma)-\log\gamma-\log(1-\theta)} \right\rceil.
\end{align*}

This completes the proof of Theorem~\ref{rg:partial}. 
\eop

\subsection{Proof of Theorem~\ref{optimality-partial:rg}}

For (b), consider the case that for each $t$, the NTP distribution is $\mathbf{P}:=(1/m,1/m,\cdots,1/m)$. Then $\mathcal{S}(\mathbf{P},\zeta) = \left( \mathbf{1}\{1\in\zeta\}\cdot\frac{1}{\gamma\cdot m},\mathbf{1}\{2\in\zeta\}\cdot\frac{1}{\gamma\cdot m},\cdots,\mathbf{1}\{m\in\zeta\}\cdot\frac{1}{\gamma\cdot m} \right)$. Consider the case that $\mathbf{Q}=(q_1,\cdots,q_m), q_i=\mathbf{1}\{i\in \zeta\}\frac{1+\theta}{2\gamma\cdot m}+\mathbf{1}\{i\notin\zeta\}\frac{1-\theta}{2(1-\gamma)\cdot m}$. Denote the asymptotic sum of type I error and type II error under $\mathbf{P}$ and $\mathbf{Q}$ as $f_{\mathbf{P},\mathbf{Q}}(\psi)$. Then we have
\begin{align*}
        \inf_{\psi}f(\psi)\geq \inf_{\psi}f_{\mathbf{P},\mathbf{Q}}(\psi).
\end{align*}

We next compute the likelihood of $(\omega_t,\zeta_t)_{1:n}$ under $\mathcal{H}_0$ and $\mathcal{H}_1$ as, 
\begin{align*}
\mathbb{P}_{\mathcal{H}_0}(\omega_{1:n},\zeta_{1:n}) & = \left( \frac{1}{\binom{m}{\gamma\cdot m}}\cdot\frac{1}{m} \right)^n\\
\mathbb{P}_{\mathcal{H}_1}(\omega_{1:n},\zeta_{1:n}) & = \prod_{t=1}^{n}\left( \mathbf{1}\{\omega_t\in\zeta_t\}\cdot\frac{1+\theta}{2\gamma\cdot m}+\mathbf{1}\{\omega_t\notin\zeta_t\}\cdot\frac{1-\theta}{2(1-\gamma)\cdot m} \right) \cdot\frac{1}{\binom{m}{\gamma\cdot m}}.
\end{align*}
By Lemma \ref{tv}, the optimal test is to reject the null hypothesis when $\mathbb{P}_{\mathcal{H}_0}(\omega_{1:n},\zeta_{1:n})\leq \mathbb{P}_{\mathcal{H}_1}(\omega_{1:n},\zeta_{1:n})$. So the rejection region is, 
\begin{align*}
R=\big\{ (\omega_{1:n},\zeta_{1:n})|\prod_{t=1}^{n}(\mathbf{1}\{\omega_t\in\zeta_t\}\cdot\frac{1+\theta}{2\gamma}+\mathbf{1}\{\omega_t\notin\zeta_t\}\cdot\frac{1-\theta}{2(1-\gamma)})\geq 1 \big\}.
\end{align*}
Henceforth, we have proved that $\inf_\psi f(\psi)\geq f(\psi_0)$. By definition, we have $f(\psi_0)\geq \inf_\psi f(\psi)$. This proves (b). 

For (a), consider the same $\mathbf{P}$ and $\mathbf{Q}$, under which, the optimal rejection region is,
\begin{align*}
R=\left\{ \prod_{t=1}^{n}(\mathbf{1}(\omega_t\in\zeta_t)\cdot\frac{1+\theta}{2\gamma m}+\mathbf{1}(\omega_t\notin\zeta_t)\frac{1-\theta}{2(1-\gamma)m})\geq \eta\cdot(\frac{1}{m})^n \right\},
\end{align*}
where $\eta$ is a constant to keep the type I error at $\alpha$. This rejection rule is equivalent to the rejection rule that $\sum_{t=1}^{n}Y_t\geq\eta^{'}$ for some $\eta^{'}$. This proves (a).

Together, it completes the proof of Theorem~\ref{optimality-partial:rg}. 
\eop

\section{Additional Numerical Studies}
\label{append-sec:numericals}

\subsection{Derivation of the optimal threshold for a general score function}
\label{append-sec:threshold}

For a general score function $h$, we determine the optimal threshold $\gamma_n$, such that we minimize the sum of type I and type II errors, i.e., 
\begin{align*}
\min_{\gamma_n}\mathbb{P}_{\mathcal{H}_0}\left( \sum_{i=1}^{n}h(Y_t)\geq \gamma_n \right) + \mathbb{P}_{\mathcal{H}_1}\left( \sum_{i=1}^{n}h(Y_t)<\gamma_n \right).
\end{align*}
By the large deviation theory, the sum of type I and type II errors has a representation that is nearly tight when $n\to\infty$, i.e., 
\begin{align*}
&\mathbb{P}_{\mathcal{H}_0}\left( \sum_{i=1}^{n}h(Y_t)\geq \gamma_n \right) + \mathbb{P}_{\mathcal{H}_1}\left( \sum_{i=1}^{n}h(Y_t)<\gamma_n \right)\\
\geq \; &\inf_{\theta_1,\theta_2}\frac{\exp\{n\cdot\log\mathbb{E}_0\exp{\theta_1 h(Y)}\}}{\exp(\theta_1 \gamma)} + \exp(\theta_2 \gamma) \cdot\exp\{n\log\mathbb{E}_1\exp{-\theta_2 h(Y)}\}.
\end{align*}
Note that, given $\theta_1,\theta_2$, choosing 
\[
\gamma_n=\frac{1}{\theta_1+\theta_2}\log\frac{\theta_1\exp\{n\cdot\log\mathbb{E}_0\exp{\theta_1 h(Y)}\}}{\theta_2\exp[n\cdot\log\mathbb{E}_1\exp\{-\theta_2 h(Y)\}]}
\]
can minimize the sum of type I and type II errors. Therefore, 
\begin{align*}
&\inf_{\gamma_n}\mathbb{P}_{\mathcal{H}_0}\left( \sum_{i=1}^{n}h(Y_t)\geq \gamma_n \right) + \mathbb{P}_{\mathcal{H}_1}\left( \sum_{i=1}^{n}h(Y_t)<\gamma_n \right)\\
\geq \; &\inf_{\theta_1,\theta_2}\exp\left(\frac{\theta_2}{\theta_1+\theta_2}\log\mathbb{E}_0\exp{\theta_1 h(Y)} \right) \cdot \exp\left( \frac{\theta_1}{\theta_1+\theta_2}\log\mathbb{E}_1\exp\{-\theta_2 h(Y)\} \right)\\
= \; &\exp\left(\inf_{\theta_1,\theta_2}\frac{\theta_2}{\theta_1+\theta_2}\log\mathbb{E}_0\exp\{\theta_1 h(Y)\}+\frac{\theta_1}{\theta_1+\theta_2}\log\mathbb{E}_1\exp\{-\theta_2h(Y)\}\right).
\end{align*}
We then calculate the value of $\gamma_n$ using 
\begin{align*}
\gamma_n = \frac{1}{\theta_1^*+\theta_2^*}\left( \log\frac{\theta_1^*}{\theta_2^*} + n \log\mathbb{E}_0\exp\{\theta_1^*h(Y)\} - n \log\mathbb{E}_1\exp\{-\theta_2^*h(Y)\} \right).
\end{align*}
Since our deviation is tight only when $n$ is large enough, in our simulation, for convenience, we set 
\[
\hat{\gamma}_n=\frac{n}{\theta_1^*+\theta_2^*}\left( \log\mathbb{E}_0\exp\{\theta_1^*h(Y)\} - \log\mathbb{E}_1\exp\{-\theta_2^*h(Y)\} \right).
\]

\subsection{Additional numerical results for Gumbel-max watermark}
\label{append-sec:gm-sim}

\begin{figure}[H]
\centering
\begin{tabular}{cc}
\includegraphics[width=0.4\textwidth,height=1.7in]{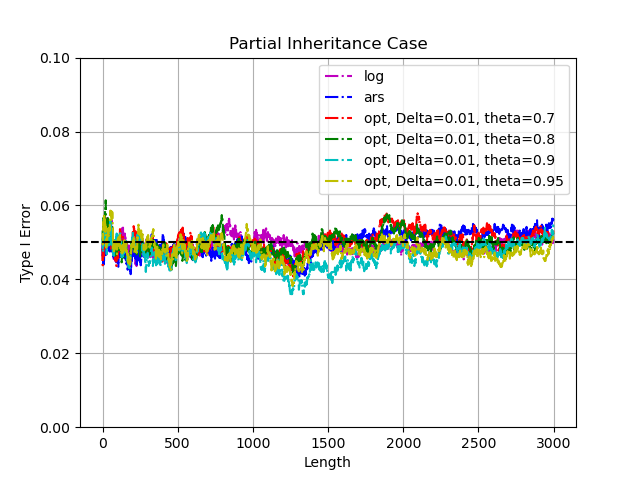} & 
\includegraphics[width=0.4\textwidth,height=1.7in]{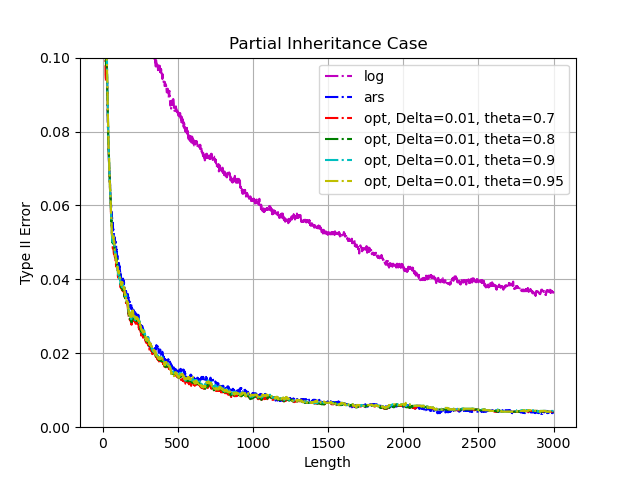} \\
\end{tabular}
\caption{Average type I and type II errors versus text length for the Gumbel-max watermark under the setting of fixed type I error, where $\Delta=0.01$ and $\theta = \{0.7, 0.8, 0.9, 0.95\}$.}
\label{fig:gm-fix-d10}
\end{figure}

\begin{figure}[H]
\centering
\begin{tabular}{cc}
\includegraphics[width=0.4\textwidth,height=1.7in]{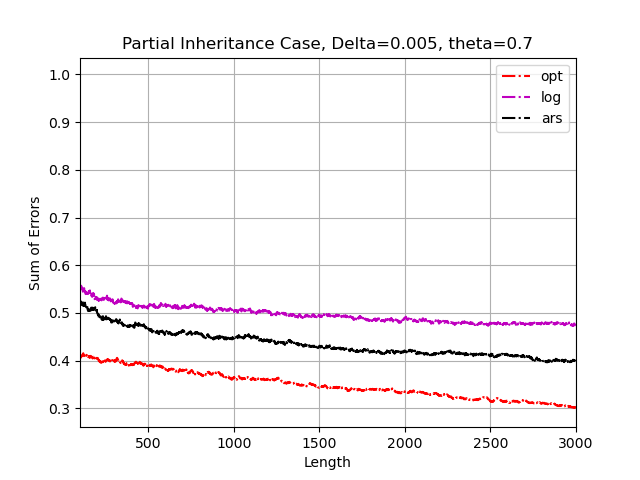} & 
\includegraphics[width=0.4\textwidth,height=1.7in]{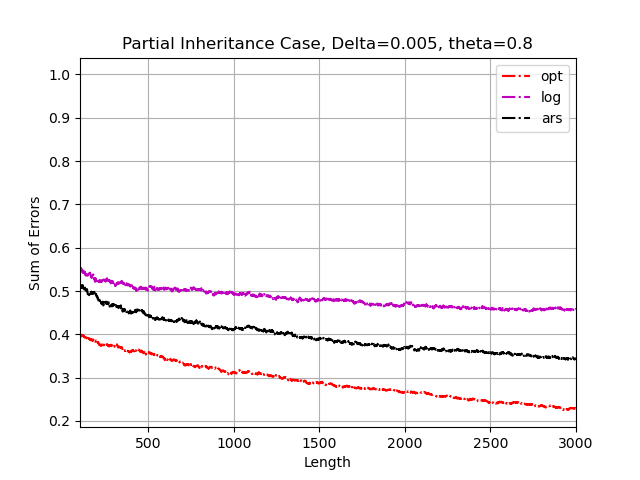} \\
\includegraphics[width=0.4\textwidth,height=1.7in]{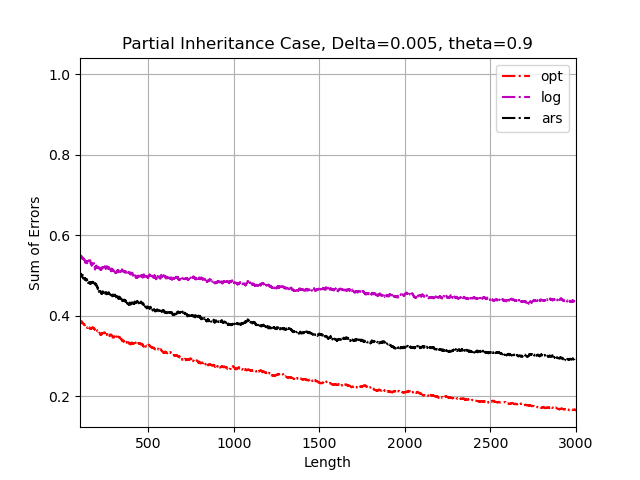} &
\includegraphics[width=0.4\textwidth,height=1.7in]{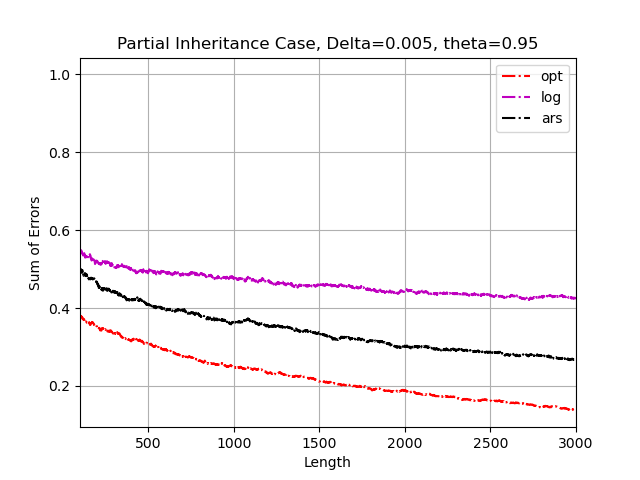} \\
\end{tabular}
\caption{Average sum of type I and type II errors versus text length for the Gumbel-max watermark under the setting of minimizing the sum of type I and type II errors, where $\Delta=0.005$ and $\theta=\{0.7, 0.8, 0.9, 0.95\}$.}
\label{fig:gm-sum-d05}
\end{figure}

We report additional simulation results for the Gumbel-max watermark, adopting the same setup as in Section \ref{subsec:gm-sim}, but considering different working values for $\Delta$ and $\theta$. Figure \ref{fig:gm-fix-d10} reports the average type I and type II errors  versus the text length, based on 5000 replications, under the setting of fixed type I error, where $\Delta = 0.01$ and $\theta = \{0.7, 0.8, 0.9, 0.95\}$. Figures \ref{fig:gm-sum-d05} and {fig:gm-sum-d10} report the average sum of type I and type II errors versus text length, based on 5,000 repetitions, under the setting of minimizing the sum of two errors, where $\Delta=0.005$ and $\Delta=0.01$, respectively, while $\theta=\{0.7, 0.8, 0.9, 0.95\}$. We have observed that our proposed tests remain effective and robust as long as $\Delta$ and $\theta$ are chosen within a reasonable range. 

\begin{figure}[H]
\centering
\begin{tabular}{cc}
\includegraphics[width=0.4\textwidth,height=1.7in]{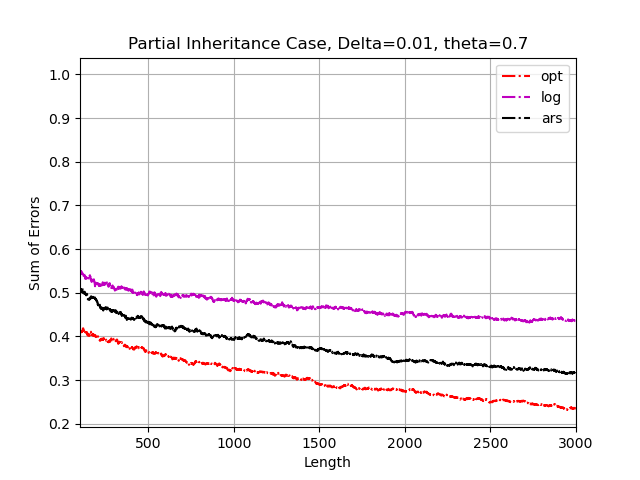} & 
\includegraphics[width=0.4\textwidth,height=1.7in]{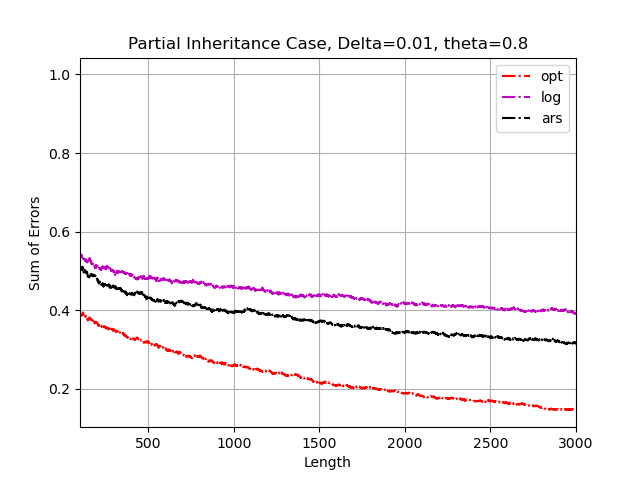} \\
\includegraphics[width=0.4\textwidth,height=1.7in]{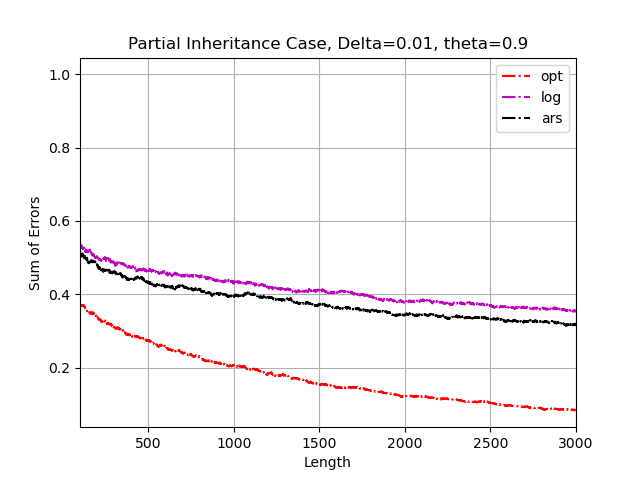} &
\includegraphics[width=0.4\textwidth,height=1.7in]{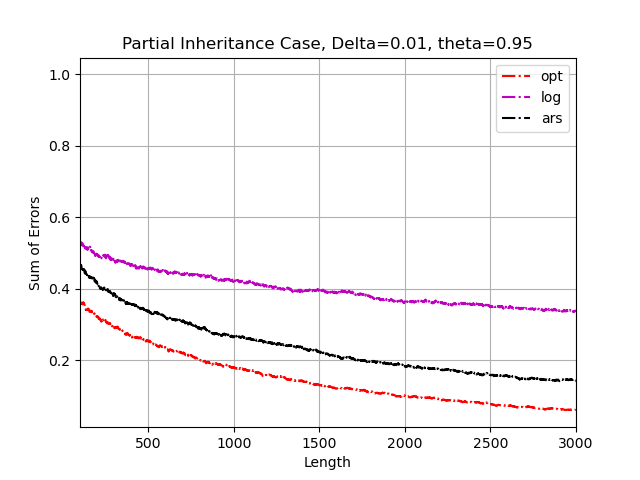} \\
\end{tabular}
\caption{Average sum of type I and type II errors versus text length for the Gumbel-max watermark under the setting of minimizing the sum of type I and type II errors, where $\Delta=0.01$ and $\theta=\{0.7, 0.8, 0.9, 0.95\}$.}
\label{fig:gm-sum-d10}
\end{figure}

\subsection{Additional numerical results for red-green-list watermark}
\label{append-sec:rg-sim}

\begin{figure}[H]
\centering
\begin{tabular}{cc}
\includegraphics[width=0.4\textwidth,height=1.7in]{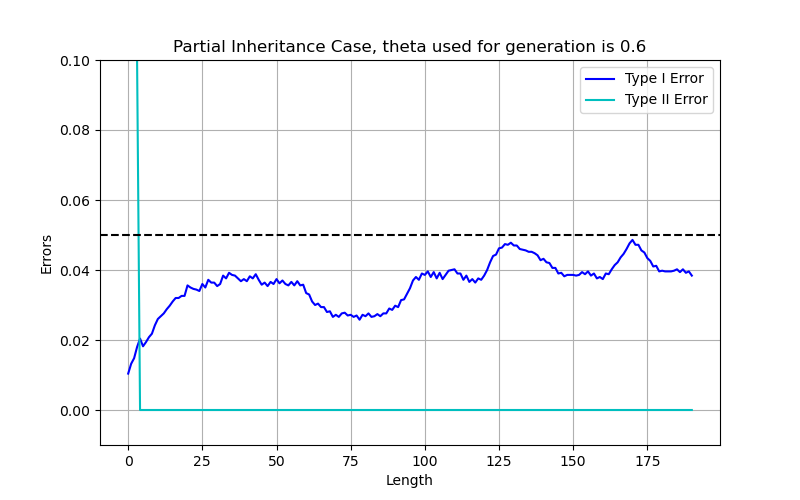} & 
\includegraphics[width=0.4\textwidth,height=1.7in]{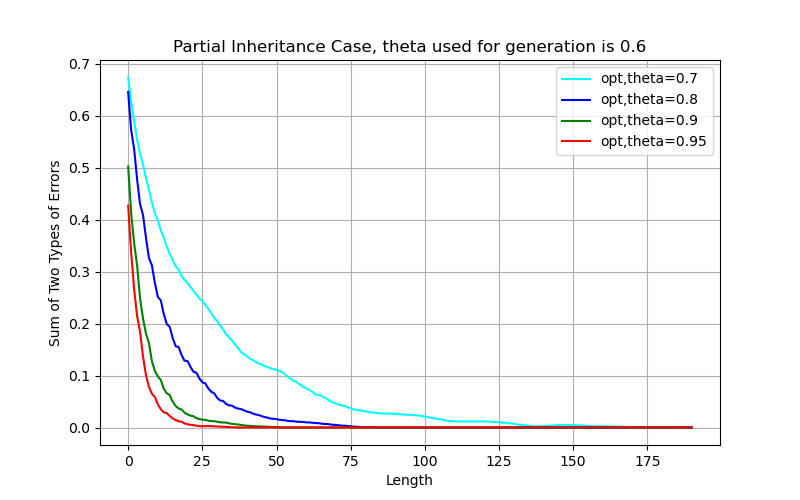} \\
\includegraphics[width=0.4\textwidth,height=1.7in]{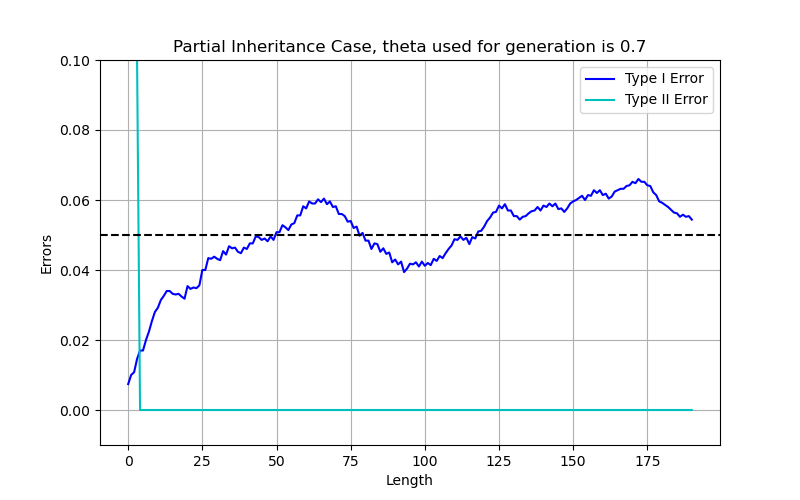} & 
\includegraphics[width=0.4\textwidth,height=1.7in]{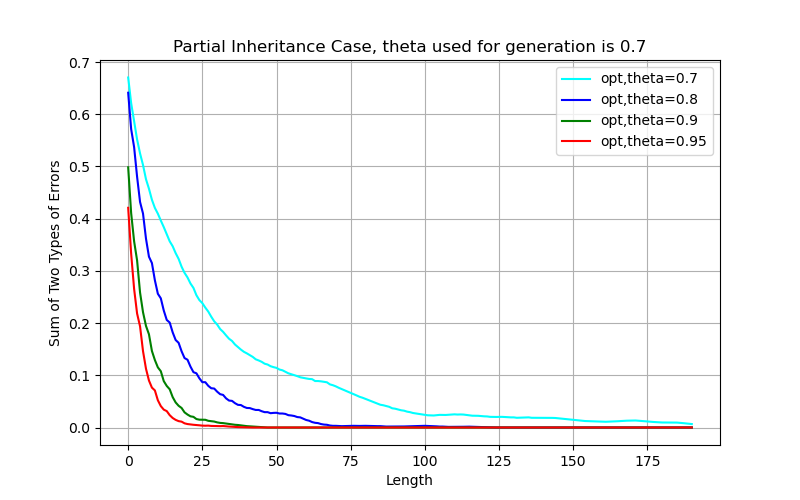} \\
\end{tabular}
\label{fig:rg-fixsum-truetheta}
\caption{Left panel: average type I and type II errors versus text length for the red-green-list watermark under the setting of fixed type I error; right panel: average sum of type I and type II errors versus text length for the red-green-list watermark under the setting of minimizing the sum of type I and type II errors, where $\theta^* = \{0.6, 0.7\}$ and $\theta = \{0.7, 0.8, 0.9, 0.95\}$.}
\end{figure}

We report additional simulation results for the red-green-list watermark, adopting the same setup as in Section \ref{subsec:rg-sim}, but considering different true value for $\theta^* = \{0.6, 0.7\}$ with the working value of $\theta = \{0.7, 0.8, 0.9, 0.95\}$. Figure \ref{fig:rg-fixsum-truetheta}, left panel, reports the average type I and type II errors versus the text length, based on 5000 replications, under the setting of fixed type I error. Figure \ref{fig:rg-fixsum-truetheta}, right panel, reports the average sum of type I and type II errors versus the text length, based on 5,000 replications, under the setting of minimizing the sum of type I and type II errors. Again, we have observed that our proposed tests remain effective and robust as long as $\theta$ is chosen within a reasonable range.

\end{document}